\documentclass{article}

\usepackage{iclr2027_conference,times}

\usepackage{amsmath,amssymb}
\usepackage{booktabs,array,tabularx}
\usepackage{graphicx}
\usepackage{microtype}
\usepackage{url}
\usepackage{placeins}
\usepackage{float}
\usepackage{flafter}
\usepackage[hidelinks]{hyperref}

\hypersetup{
  pdftitle={Is H\&E Image-to-Spatial Transcriptomics Simpler Than It Looks?},
  pdfauthor={Duc T. Nguyen, Thanh Ha Do, Phuong M. Cao, Hieu Pham}
}
\title{
Is H\&E Image-to-Spatial Transcriptomics\\
Simpler Than It Looks?
}

\author{
Duc T. Nguyen\textsuperscript{1,4}
\quad
Thanh Ha Do\textsuperscript{2}
\quad
Phuong M. Cao\textsuperscript{3}
\quad
Hieu Pham\textsuperscript{1,4}
\\[0.7em]
\begin{tabular}{l}
\small \textsuperscript{1}VinUniversity, Hanoi, Vietnam\\
\small \textsuperscript{2}Posts and Telecommunications Institute of Technology, Hanoi, Vietnam\\
\small \textsuperscript{3}National Center for Supercomputing Applications, University of Illinois Urbana-Champaign, Urbana, IL, USA\\
\small \textsuperscript{4}VinUni-Illinois Smart Health Center (VISHC), VinUniversity, Hanoi, Vietnam
\end{tabular}
}

\iclrfinalcopy

\begin{document}

\maketitle
\lhead{Preprint}
\begin{abstract}
Predicting spatial gene expression from routine H\&E histology offers a scalable route toward spatial molecular profiling. Recent work has pursued increasingly sophisticated architectures to capture spatial context and richer expression structure. At the same time, simple estimators have shown strong performance in several studies, but what they already solve and where additional complexity is needed remain unclear. We study this behavior through the structure of prediction error under the mean-squared error (MSE) objective. Differences in average expression across genes can account for a substantial part of aggregate prediction performance, while a key unresolved error lies in recovering variation within each slide. Decomposing MSE into slide-level and within-slide components, we find that the within-slide component has lower residual-normalized parameter sensitivity in controlled neural experiments. This motivates Component-Guided Loss (CGL), which increases supervision of the within-slide component. CGL-Linear is a closed-form affine instantiation that achieves overall state-of-the-art performance across HEST-1k cohorts and gene-panel sizes. The same within-slide supervision improves existing neural models. These results suggest that substantial gains can come from aligning the training objective with prediction-error structure rather than increasing model complexity.
\end{abstract}

\section{Introduction}

Spatial transcriptomics measures gene expression at known locations in a tissue section, linking molecular variation to tissue morphology~\citep{stahl2016visualization}. Its cost and experimental requirements motivate predicting these measurements from routine hematoxylin-and-eosin (H\&E) images~\citep{he2020integrating}. Accurate prediction would make spatial molecular profiling more accessible and enable the study of tissue samples for which expression measurements are unavailable.

Recent approaches use spatial attention, retrieval, and generative modeling to capture richer relationships between histology and expression~\citep{xie2023bleep,huang2025stflow,wu2026hist,hu2026histoprism,byeon2026hexst}. Yet lightweight predictors remain competitive in representation studies and broader benchmarks~\citep{jaume2024hest,pan2025peka,wang2025benchmark}. Their performance raises a basic question. What do simple predictors already capture, and how much of their remaining error can be reduced without increasing their capacity?

\begin{figure*}[h]
    \centering
    \includegraphics[width=\textwidth]{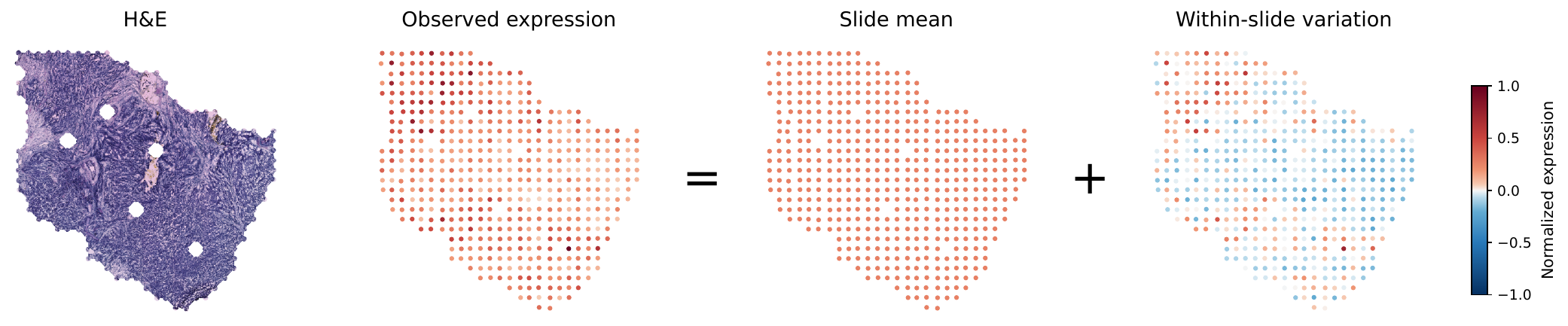}
    \caption{\textbf{Slide-wise expression structure.}
    For one gene on one slide, observed expression can be viewed as a slide-wise mean together with variation around that mean. Each dot denotes a measured spatial location, and the H\&E image is shown for spatial context.}
    \label{fig:target_decomposition}
\end{figure*}

We approach this question by distinguishing expression shared at the slide level from variation within a slide. Figure~\ref{fig:target_decomposition} illustrates this structure for a single gene. A predictor that assigns the training mean of each gene to every location can obtain a strong aggregate score when all spot--gene measurements are evaluated together. Such a prediction preserves differences in average expression across genes but contains no spatial variation. This mean-only reference shows why evaluating a predictor's contribution requires examining what it recovers beyond gene-specific baselines.

The same distinction carries into learning under mean-squared error (MSE), a widely used objective for H\&E-to-spatial-transcriptomics regression~\citep{he2020integrating,wu2026hist,hu2026histoprism}. Locations are grouped within slides, so prediction error can arise from an incorrect slide-wise mean or from inaccurate variation around that mean. We separate MSE exactly into these two components. This gives a common basis for examining both what a predictor recovers and which errors remain difficult to reduce.

Although the two components enter MSE with equal coefficients, they are optimized through shared parameters. Equal weighting therefore need not give equal progress. We characterize their response through residual-normalized parameter sensitivity, which accounts for differences in error magnitude. In controlled neural experiments, the within-slide component has lower sensitivity under standard MSE training. The same ordering appears on held-out slides. This finding motivates Component-Guided Loss (CGL), which increases the weight of within-slide error. Reweighting changes the fitting trade-off and improves held-out spatial recovery.

For an affine readout on frozen image representations, CGL remains a regularized least-squares problem. Its closed-form solution, CGL-Linear, uses the same predictor class as ordinary ridge regression. CGL-Linear improves on this baseline and achieves overall state-of-the-art performance across HEST-1k cohorts and gene-panel sizes. Since both affine estimators are solved directly, their comparison shows that additional within-slide signal can be recovered by changing the fitted objective at fixed readout capacity. The same supervision term also improves spatial recovery in existing neural methods. Together, these results identify a source of improvement available within established predictors and provide a practical reference for assessing the value of additional model complexity.

\section{Related Work}

\paragraph{Histology-to-expression prediction.}
Patch-level regression~\citep{he2020integrating} has developed alongside models that use broader tissue context. HiST builds a sparse spatial hierarchy~\citep{wu2026hist}, HistoPrism combines cancer-type conditioning with Transformer-based feature aggregation~\citep{hu2026histoprism}, and STFlow models whole-slide expression through conditional flow matching~\citep{huang2025stflow}. Representation quality provides another route to improved prediction. HEST-1k evaluates pretrained encoders with PCA and ridge regression~\citep{jaume2024hest}, while PEKA uses a lightweight readout after transcriptomic adaptation~\citep{pan2025peka}. Benchmark studies examine the limits of both approaches. \citet{wang2025benchmark} report that more complex architectures are not consistently superior, and HESCAPE finds that stronger cross-modal alignment does not necessarily improve direct expression prediction~\citep{gindra2025hescape}.

\paragraph{Spatial variation.}
Prior work has explicitly considered the gap between predicting expression levels and preserving variation. BLEEP learns a joint image--expression space for retrieval and reports that regression baselines can reproduce gene means while underestimating their variance~\citep{xie2023bleep}. SEPAL predicts deviations from training-set gene means and adds graph-based spatial correction~\citep{mejia2023sepal}. HEXST uses geometry-aware attention and an auxiliary head that matches standardized expression deviations~\citep{byeon2026hexst}. Our analysis focuses on the slide-wise components of the final prediction error under MSE. We study how these components interact with shared parameters and evaluate a penalty applied directly to within-slide prediction error. The resulting objective supports both a closed-form affine readout and augmentation of existing neural objectives.

\section{Method}

\subsection{Slide-wise error decomposition}
\label{sec:problem}

Consider \(M\) locations from \(S\) tissue slides. Location \(i\) has an image representation \(\mathbf q_i\in\mathbb R^d\), an expression vector \(\mathbf y_i\in\mathbb R^G\), and slide membership \(s_i\). We stack the representations and targets as \(\mathbf Q\in\mathbb R^{M\times d}\) and \(\mathbf Y\in\mathbb R^{M\times G}\). For predictions \(\widehat{\mathbf Y}_\theta\), let \(\mathbf E=\widehat{\mathbf Y}_\theta-\mathbf Y\). Standard MSE is \(\mathcal L_{\mathrm{MSE}}=\|\mathbf E\|_F^2/(MG)\).

The grouping of locations distinguishes errors in a slide's mean expression from errors in its internal variation. Ordering locations by slide, define
\begin{equation}
\mathbf P_s=\frac{1}{n_s}\mathbf 1_{n_s}\mathbf 1_{n_s}^{\top},
\qquad
\mathbf P=\operatorname{blockdiag}(\mathbf P_1,\ldots,\mathbf P_S),
\qquad
\mathbf C=\mathbf I-\mathbf P,
\label{eq:method-projectors}
\end{equation}
where \(n_s\) is the number of locations in slide \(s\). For any matrix \(\mathbf Z\), \(\mathbf P\mathbf Z\) repeats the mean of each slide at its locations and \(\mathbf C\mathbf Z\) retains deviations from that mean. These complementary orthogonal projections give \(\mathbf E=\mathbf P\mathbf E+\mathbf C\mathbf E\) and
\begin{equation}
\mathcal L_{\mathrm{MSE}}
=
\underbrace{\frac{\|\mathbf P\mathbf E\|_F^2}{MG}}_{\mathcal L_P}
+
\underbrace{\frac{\|\mathbf C\mathbf E\|_F^2}{MG}}_{\mathcal L_C}.
\label{eq:method-error-decomposition}
\end{equation}
Here, \(\mathcal L_P\) measures error in the mean expression of each gene within each slide. The term \(\mathcal L_C\) measures error in deviations around those means. A prediction can therefore have small slide-mean error while still missing within-slide variation. The within-slide component is invariant to additive slide-constant shifts, since \(\mathbf C(\mathbf Y+\mathbf D)=\mathbf C\mathbf Y\) whenever \(\mathbf P\mathbf D=\mathbf D\). Appendix~\ref{app:projection-properties} gives the derivation.

\subsection{Component-guided loss}
\label{sec:optimization-sensitivity}
\label{sec:local-loss}
\label{sec:cgl}

Orthogonality in prediction space does not make the two errors equally responsive to shared parameters. Let \(\mathbf e_P=\operatorname{vec}(\mathbf P\mathbf E)\) and \(\mathbf e_C=\operatorname{vec}(\mathbf C\mathbf E)\). Denote the prediction Jacobian by \(\mathbf J_\theta=\partial\operatorname{vec}(\widehat{\mathbf Y}_\theta)/\partial\theta\). For \(j\in\{P,C\}\), the component gradient is \(\mathbf g_j=\nabla_\theta\mathcal L_j=2\mathbf J_\theta^{\top}\mathbf e_j/(MG)\). Its magnitude depends on both residual size and the model's parameter response. For \(\mathcal L_j>0\), we separate these effects through the residual-normalized parameter sensitivity
\begin{equation}
\kappa_j
=
\frac{\|\mathbf J_\theta^{\top}\mathbf e_j\|_2^2}{\|\mathbf e_j\|_2^2}
=
\frac{MG}{4}\frac{\|\mathbf g_j\|_2^2}{\mathcal L_j}.
\label{eq:method-sensitivity}
\end{equation}
The identity \(\|\mathbf g_j\|_2^2=4\mathcal L_j\kappa_j/(MG)\) shows that a component with lower sensitivity produces a smaller gradient for equal residual energy. The relative sensitivities depend on the current residuals and the model Jacobian.

Under MSE gradient flow, \(\dot\theta=-(\mathbf g_P+\mathbf g_C)\), the component losses satisfy
\begin{equation}
\dot{\mathcal L}_P=-\|\mathbf g_P\|_2^2-\langle\mathbf g_P,\mathbf g_C\rangle,
\qquad
\dot{\mathcal L}_C=-\|\mathbf g_C\|_2^2-\langle\mathbf g_P,\mathbf g_C\rangle.
\label{eq:method-component-flow}
\end{equation}
The interaction term contributes equally to the absolute decrease rates. Their difference is therefore \(\|\mathbf g_P\|_2^2-\|\mathbf g_C\|_2^2\). Equal coefficients in MSE need not yield equal component-wise progress, even though the residuals are orthogonal.

Slide-wise centering gives a useful interpretation of this asymmetry. A parameter change that shifts predictions together within a slide can act on its mean error. The same shared response cancels from the within-slide gradient because centered residuals sum to zero. Reducing \(\mathcal L_C\) therefore depends on how parameter responses differ among locations. When these differences are weak, within-slide error may produce little gradient despite substantial residual energy. Appendix~\ref{app:optimization-sensitivity} derives this cancellation and the component dynamics.

This potential disadvantage motivates placing additional weight on within-slide recovery. We define \emph{Component-Guided Loss} (CGL) as
\begin{equation}
\boxed{
\mathcal L_{\mathrm{CGL}}
=
\mathcal L_P+(1+\lambda)\mathcal L_C
=
\mathcal L_{\mathrm{MSE}}+\lambda\mathcal L_C,
\qquad \lambda\geq0.
}
\label{eq:method-cgl}
\end{equation}
Its gradient is \(\mathbf g_P+(1+\lambda)\mathbf g_C\). At the same parameter state, CGL adds \(\lambda\|\mathbf g_C\|_2^2\) to the gradient-flow decrease rate of \(\mathcal L_C\). This provides a direct within-slide descent contribution while retaining supervision of the slide mean. Appendix~\ref{app:cgl-properties} gives the proof. The weight is fixed during fitting and selected on validation data.

\subsection{CGL-Linear}
\label{sec:shared-linear}

CGL admits a closed-form solution for a shared affine readout on frozen features. We call this estimator \emph{CGL-Linear}. For training data, let \(\boldsymbol\mu_Q\) and \(\boldsymbol\mu_Y\) be the pooled feature and expression means. Define \(\mathbf X=\mathbf Q-\mathbf1\boldsymbol\mu_Q^{\top}\) and \(\mathbf Y_c=\mathbf Y-\mathbf1\boldsymbol\mu_Y^{\top}\). A single coefficient matrix \(\mathbf A\in\mathbb R^{d\times G}\) gives predictions \(\widehat{\mathbf Y}=\mathbf1\boldsymbol\mu_Y^{\top}+\mathbf X\mathbf A\).

For residual \(\mathbf R=\mathbf X\mathbf A-\mathbf Y_c\), the unnormalized data-fitting objective is
\begin{equation}
\mathcal E_\lambda(\mathbf A)
=
\|\mathbf P\mathbf R\|_F^2+(1+\lambda)\|\mathbf C\mathbf R\|_F^2
=
\langle\mathbf R,(\mathbf I+\lambda\mathbf C)\mathbf R\rangle_F.
\label{eq:method-shared-fit}
\end{equation}
Both components are fitted through the same \(\mathbf A\). Adding \(L_2\) regularization with coefficient \(\alpha_{\mathrm{raw}}>0\) yields
\begin{equation}
\mathbf A_{\lambda,\alpha}
=
\left[\mathbf X^{\top}(\mathbf I+\lambda\mathbf C)\mathbf X+\alpha_{\mathrm{raw}}\mathbf I\right]^{-1}
\mathbf X^{\top}(\mathbf I+\lambda\mathbf C)\mathbf Y_c.
\label{eq:method-cgl-linear}
\end{equation}
The coefficients are obtained by a single linear solve. Appendix~\ref{app:weighted-linear} derives the solution with an unregularized intercept and specifies the regularization scale.

For new features \(\mathbf Q_*\), prediction is
\begin{equation}
\widehat{\mathbf Y}_*
=
\mathbf1\boldsymbol\mu_Y^{\top}
+(\mathbf Q_*-\mathbf1\boldsymbol\mu_Q^{\top})\mathbf A_{\lambda,\alpha}.
\label{eq:method-shared-inference}
\end{equation}
Slide membership is needed only for fitting. Inference is pointwise and uses the training means. Setting \(\lambda=0\) recovers ordinary affine ridge regression, so the two estimators have the same predictor class and parameter count. The component weight changes which affine function is fitted within that class. The selection protocol is given in Appendix~\ref{app:selection}.

\section{Experiments}

\subsection{Experimental setup}
\label{sec:experimental_setting}

\paragraph{Data.}
We evaluate ten organ-level cohorts from HEST-1k~\citep{jaume2024hest}, covering brain, breast, colon, heart, kidney, liver, lung, prostate, skin, and uterus. HEST-1k aggregates spatial transcriptomics profiles from public and internal cohorts collected across multiple sources, and the cohorts used here retain this diversity in data origin. Models are fitted separately for each cohort using approximately \(70/15/15\) training, validation, and test splits, constructed to avoid patient-level leakage across splits. Training data determine nested panels of \(G\in\{500,1000,1500,2000\}\) highly variable genes. Appendix~\ref{app:protocol} describes the cohort construction, splits, and preprocessing.

\paragraph{Compared methods.}
We compare CGL-Linear and ordinary affine ridge regression, denoted Linear, with HiST~\citep{wu2026hist}, HistoPrism~\citep{hu2026histoprism}, HEXST~\citep{byeon2026hexst}, STFlow~\citep{huang2025stflow}, BLEEP~\citep{xie2023bleep}, and SEPAL~\citep{mejia2023sepal}. All methods use the same data splits and targets. We repeat the benchmark with frozen UNI, a pathology foundation model~\citep{chen2024uni}; DINOv2-Large, a general-purpose self-supervised vision model~\citep{oquab2024dinov2}; and ResNet18, a standard convolutional vision model~\citep{he2016resnet}. Neural methods are averaged over three seeds, while Linear and CGL-Linear are deterministic. Hyperparameters, including component weights, are selected using validation data following Appendix~\ref{app:selection}.

\paragraph{Evaluation.}
Following prior H\&E-to-spatial-transcriptomics work, we report the Pearson correlation coefficient (PCC) to measure association between predicted and measured expression~\citep{wu2026hist,hu2026histoprism}. Flat PCC correlates the complete flattened spot--gene matrices. Gene PCC computes correlation separately for each gene after pooling locations across the evaluated slides,
\[
\begin{aligned}
\mathrm{PCC}_{\mathrm{flat}}
&=
\operatorname{corr}\!\left(
\operatorname{vec}(\mathbf Y),
\operatorname{vec}(\widehat{\mathbf Y})
\right),
&
\mathrm{PCC}_{\mathrm{gene}}
&=
\frac{1}{|\mathcal V|}
\sum_{g\in\mathcal V}
\operatorname{corr}\!\left(
\mathbf Y_{:g},
\widehat{\mathbf Y}_{:g}
\right),
\end{aligned}
\]
where \(\mathcal V\) contains genes for which the correlation is defined. Gene PCC measures gene-wise association across pooled locations and can reflect both between-slide and within-slide variation.

Because correlation does not measure the magnitude of prediction error, we additionally report \(R^2\). Flat \(R^2\) measures overall reconstruction, while Local \(R^2\) measures reconstruction after removing the slide-wise mean,
\[
\begin{aligned}
R^2_{\mathrm{flat}}
&=
1-
\frac{
\|\widehat{\mathbf Y}-\mathbf Y\|_F^2
}{
\|\mathbf Y-\bar y\mathbf1_M\mathbf1_G^\top\|_F^2
},
&
R^2_{\mathrm{local}}
&=
1-
\frac{
\|\mathbf C(\widehat{\mathbf Y}-\mathbf Y)\|_F^2
}{
\|\mathbf C\mathbf Y\|_F^2
},
\end{aligned}
\]
with \(\bar y=(MG)^{-1}\sum_{i,g}Y_{ig}\). Each metric is computed separately for each cohort and then macro-averaged across the ten cohorts.

\subsection{Aggregate prediction}
\label{sec:aggregate}

We first examine what aggregate scores establish about the prediction task. MeanOnly assigns every location the training-set gene-mean vector,
\(\widehat{\mathbf Y}_{\mathrm{MeanOnly}}=\mathbf1\boldsymbol\mu_Y^\top\).
It uses no image information and contains no within-slide variation, so \(R^2_{\mathrm{local}}=0\). Gene PCC is undefined because each gene has a constant prediction.

\begin{table}[htbp]
\centering
\small
\caption{Mean-only reference and learned predictors with UNI and \(G=500\). Scores are equal-cohort test averages over ten cohorts. Gene PCC is undefined for MeanOnly because its prediction is constant for every gene. Bold marks the highest mean in each column.}
\label{tab:meanonly}
\setlength{\tabcolsep}{5.5pt}
\renewcommand{\arraystretch}{1.05}
\begin{tabular}{lcrrrr}
\toprule
Method & Venue & $\mathrm{PCC}_{\mathrm{flat}}$ & $\mathrm{PCC}_{\mathrm{gene}}$ & $R^2_{\mathrm{flat}}$ & $R^2_{\mathrm{local}}$ \\
\midrule
MeanOnly   & --          & $0.6921$ & \textnormal{N.A.} & $0.4280$ & $0.0000$ \\
\midrule
Linear     & --          & $0.7226$ & $0.1354$ & $0.4953$ & $-0.0423$ \\
HiST       & ICML'26     & $0.7363$ & $0.1343$ & $0.5067$ & $0.0805$ \\
HistoPrism & ICLR'26     & $0.7372$ & $0.1337$ & $0.5032$ & $0.0921$ \\
HEXST      & ICML'26     & $0.7126$ & $0.1560$ & $0.4547$ & $0.0497$ \\
STFlow     & ICML'25     & $0.7053$ & $0.1366$ & $0.4236$ & $-0.0565$ \\
BLEEP      & NeurIPS'23  & $0.6766$ & $0.1224$ & $0.4187$ & $-0.1171$ \\
SEPAL      & ICCVW'23    & $0.7450$ & $0.1385$ & $0.5070$ & $0.0717$ \\
CGL-Linear (Ours) & --          & $\mathbf{0.7524}$ & $\mathbf{0.1656}$ & $\mathbf{0.5247}$ & $\mathbf{0.1137}$ \\
\bottomrule
\end{tabular}
\end{table}

Table~\ref{tab:meanonly} shows that MeanOnly nevertheless obtains substantial Flat PCC and Flat \(R^2\). These scores arise from gene-specific differences in average expression even though the prediction is identical at every spatial location. Linear improves both aggregate measures and produces nonzero gene-wise association, yet its macro Local \(R^2\) remains negative. Strong aggregate performance can therefore coexist with poor reconstruction of variation within tissue slides.

The comparison also clarifies what a simple image-based readout already captures. Frozen image representations combined with a linear predictor recover substantial expression structure beyond the mean-only reference, while much less of the centered within-slide component is recovered. We next examine how these two error components behave during MSE training.

\subsection{Optimization sensitivity}
\label{sec:cgl_effect}

We examine whether the sensitivity difference characterized in Section~\ref{sec:optimization-sensitivity} appears during learning. A one-hidden-layer MLP with 256 hidden units and GELU activation is trained on frozen DINOv2~\citep{oquab2024dinov2} and ResNet18~\citep{he2016resnet} representations with \(G=500\). Standard MSE and CGL use the same architecture, initialization, optimizer, and training budget. The control protocol is given in Appendix~\ref{app:optimization-control}.

\begin{figure}[htbp]
\centering
\includegraphics[width=\linewidth]{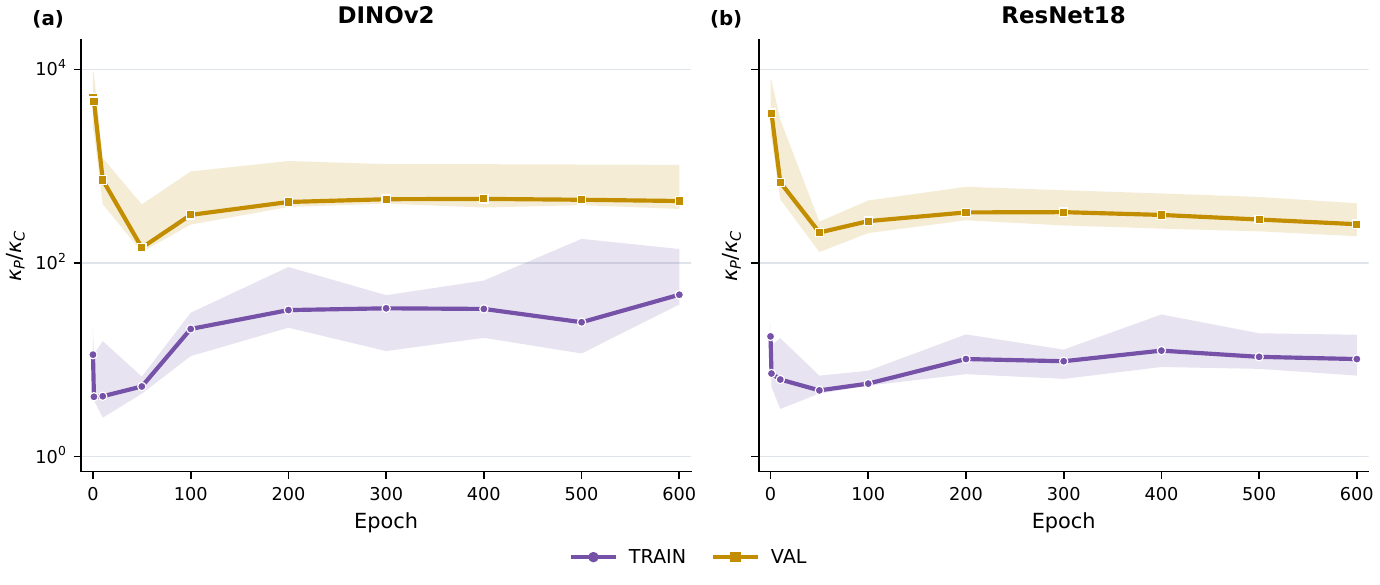}
\caption{\textbf{Component sensitivity under standard MSE.}
Curves show the cohort median of \(\kappa_P/\kappa_C\) for DINOv2 and ResNet18 on training and validation data. Shaded bands show cohort interquartile ranges. Values above one indicate greater residual-normalized sensitivity for the slide-level component.}
\label{fig:kappa_imbalance}
\end{figure}

Figure~\ref{fig:kappa_imbalance} shows a persistent sensitivity imbalance under standard MSE. The median \(\kappa_P/\kappa_C\) remains above one throughout training for both representations, and the same ordering appears on validation slides. Within-slide residuals therefore induce weaker parameter-space responses per unit residual energy in these controls. This empirical ordering identifies the within-slide component as the less responsive component in the setting studied here.

\begin{figure*}[htbp]
\centering
\includegraphics[width=\textwidth]{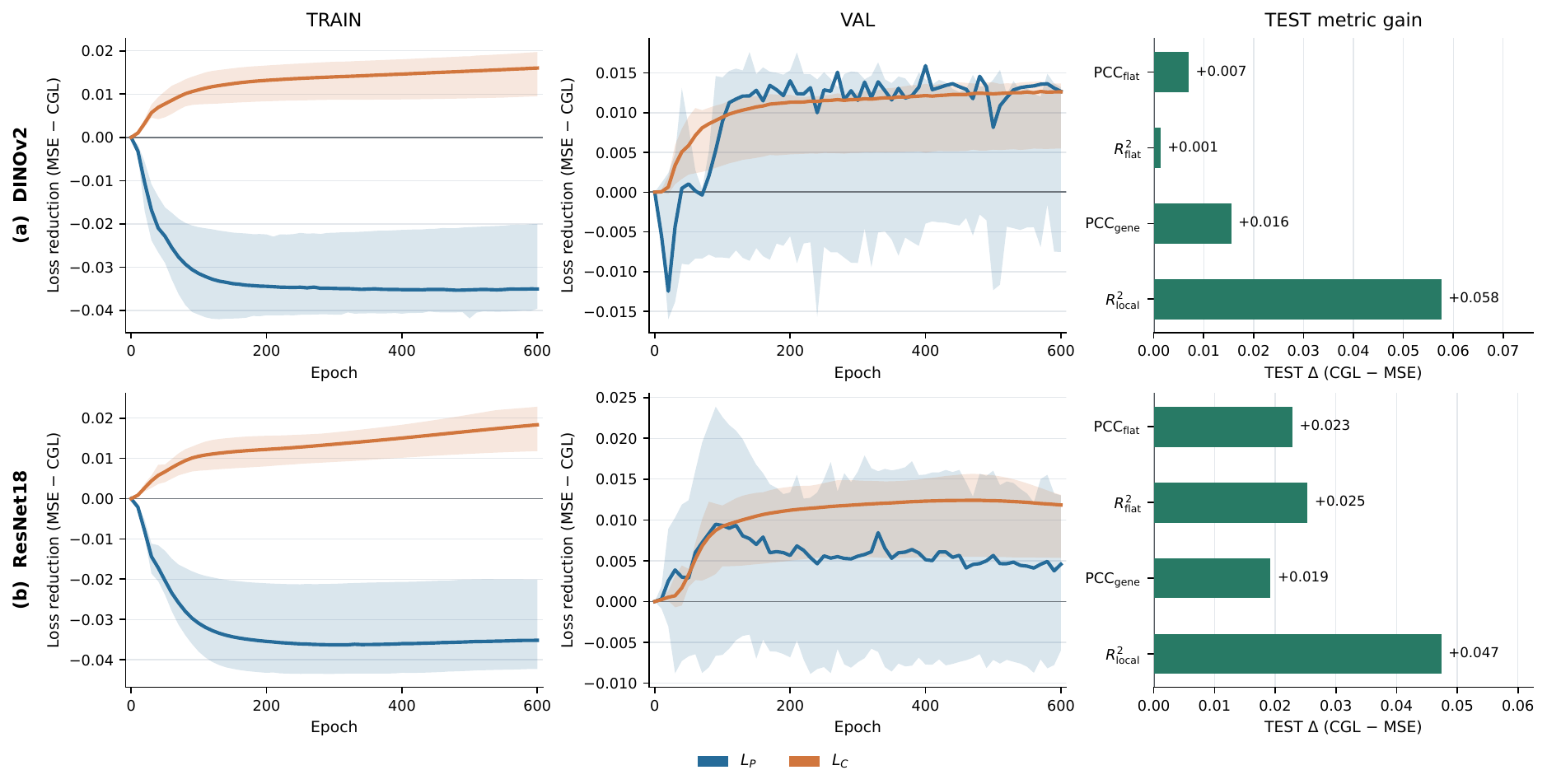}
\caption{\textbf{Effect of CGL on the controlled MLP.}
For DINOv2 (a) and ResNet18 (b), the left and middle columns show
\(\mathcal L_j^{\mathrm{MSE}}-\mathcal L_j^{\mathrm{CGL}}\)
on training and validation data. Positive values indicate lower error under CGL. Lines show equal-cohort means and shaded regions show cohort interquartile ranges. The right column reports test metric changes from MSE to CGL.}
\label{fig:cgl_dynamics}
\end{figure*}

We then train the same MLP with CGL. Figure~\ref{fig:cgl_dynamics} shows that the additional within-slide weight changes the allocation of fitting on the training data. CGL reduces \(\mathcal L_C\) while allowing a larger \(\mathcal L_P\). On validation slides, the mean difference favors CGL for the within-slide component and, over most of training, also for the slide-level component.

The held-out metrics follow the same direction. CGL improves Local \(R^2\) for both representations, directly reflecting better reconstruction of the component emphasized during training. Flat PCC, Gene PCC, and Flat \(R^2\) also improve in these controls. The experiment therefore links the measured sensitivity imbalance to a simple change in supervision that improves held-out prediction.

\subsection{Benchmark results}
\label{sec:main_benchmark}

\begin{figure*}[htbp]
\centering
\includegraphics[width=\textwidth]{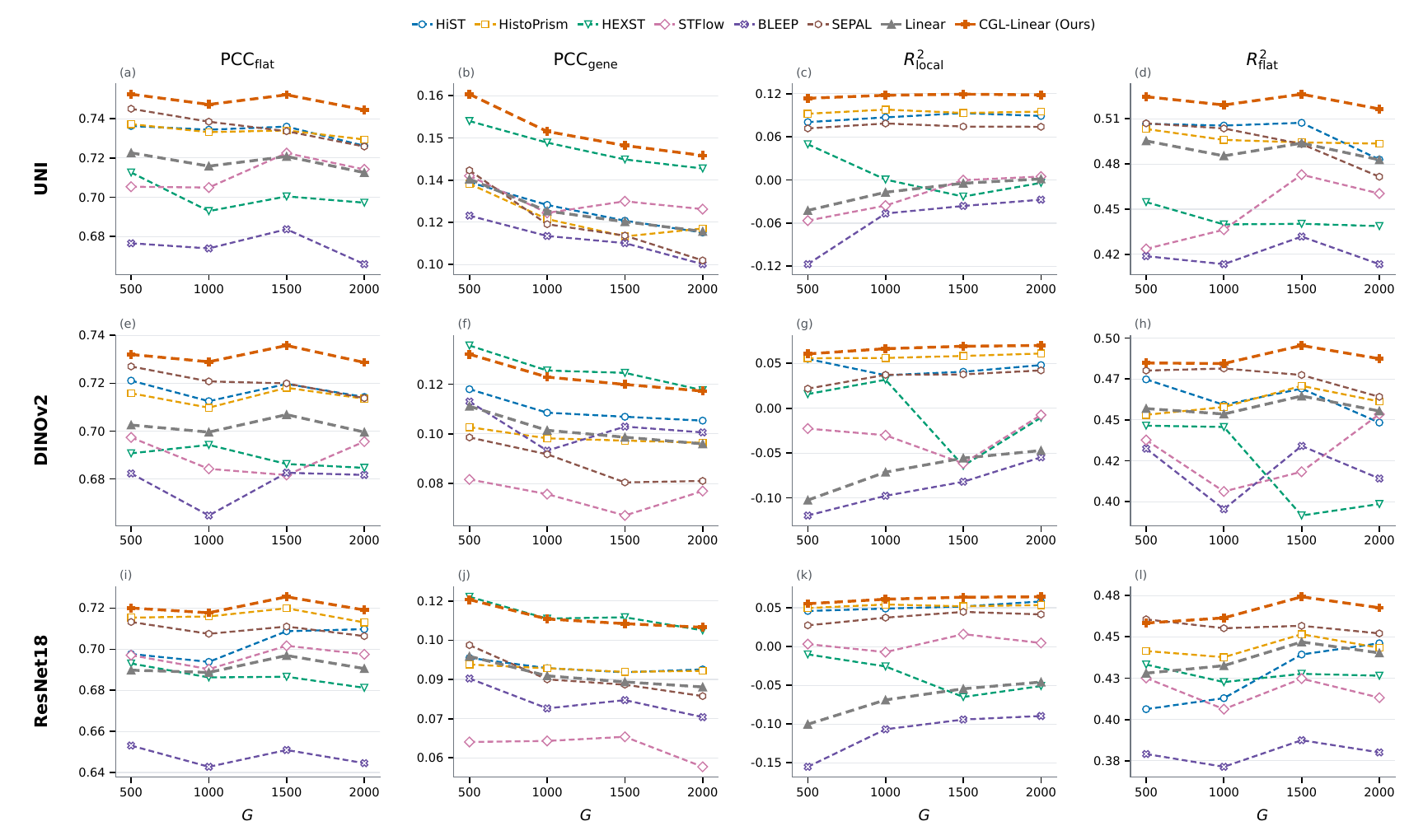}
\caption{\textbf{Benchmark results across image representations and gene panels.}
Each point is an equal-cohort macro test score over ten cohorts. Neural methods are averaged over three seeds, while Linear and CGL-Linear are deterministic. All methods use matched image representations, data splits, and targets. Higher is better for every metric.}
\label{fig:main_benchmark}
\end{figure*}

We evaluate CGL-Linear across the full benchmark in Figure~\ref{fig:main_benchmark}. It achieves the highest macro \(\mathrm{PCC}_{\mathrm{flat}}\) and \(R^2_{\mathrm{local}}\) throughout the encoder--panel grid. It also leads \(R^2_{\mathrm{flat}}\) in all but one configuration, where SEPAL~\citep{mejia2023sepal} is slightly higher. Gene PCC remains among the strongest results throughout the benchmark.

The comparison with Linear makes the effect of CGL explicit. Linear uses the same affine predictor with the original MSE objective. Across the benchmark, CGL-Linear increases macro Local \(R^2\) by \(0.110\)--\(0.163\) and makes it positive in every configuration. Flat PCC, Gene PCC, and Flat \(R^2\) also improve throughout the grid. The strongest improvement is concentrated in within-slide reconstruction while overall predictive quality is maintained or improved.

The comparison across methods further illustrates the distinction between the evaluation views. HEXST~\citep{byeon2026hexst} often obtains strong Gene PCC while showing lower Local \(R^2\). SEPAL~\citep{mejia2023sepal} also performs strongly in several settings after explicitly modeling expression deviations. CGL-Linear remains competitive in gene-wise association while providing consistently stronger within-slide reconstruction. Cohort-level results in Appendix~\ref{app:cohorts} report the corresponding tissue-specific performance and seed variation.

\subsection{Computational efficiency}
\label{sec:efficiency}

We measure fitting and inference time from \(10{,}000\) to \(50{,}000\) spatial locations using frozen ResNet18 features with \(G=500\) on an NVIDIA A5000 GPU. CGL-Linear uses \(256{,}000\) fitted coefficients and requires only \(0.446\)--\(1.255\) seconds for the complete closed-form fit and \(0.039\)--\(0.078\) seconds for inference across this range.

\begin{figure*}[htbp]
\centering
\includegraphics[width=\textwidth]{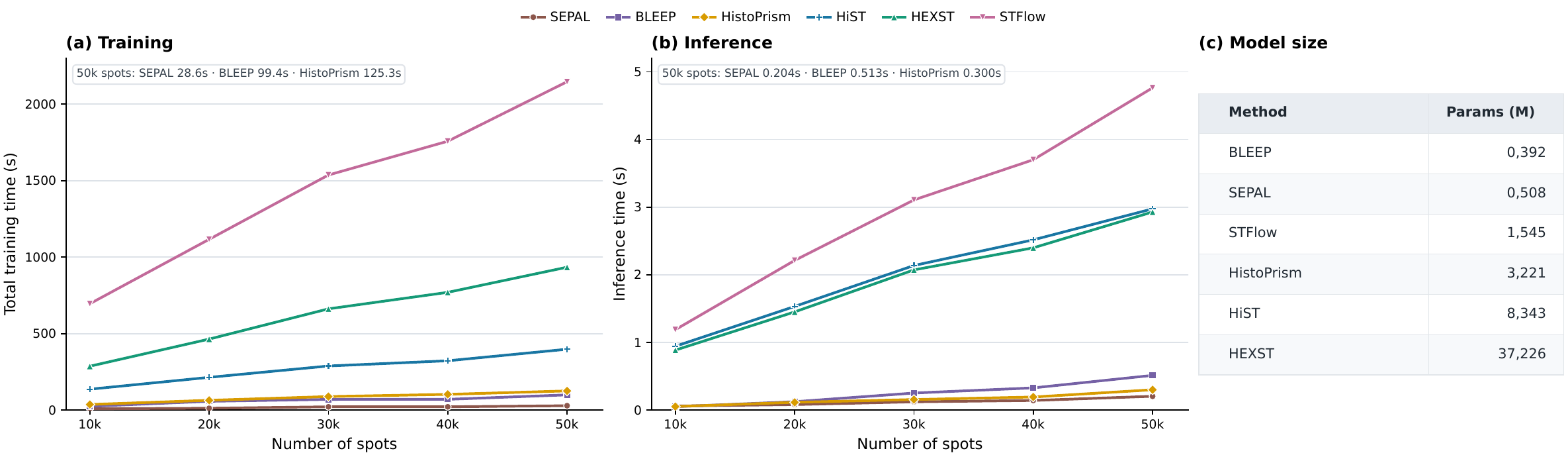}
\caption{\textbf{Computational scaling.} Total training time, inference time, and model size are compared as the number of spatial locations increases.}
\label{fig:runtime_scaling}
\end{figure*}

The computational gap increases with dataset size. At \(50{,}000\) spots, CGL-Linear fits in \(1.255\) seconds, compared with total training times of \(28.6\) seconds for SEPAL~\citep{mejia2023sepal}, \(99.4\) seconds for BLEEP~\citep{xie2023bleep}, and \(125.3\) seconds for HistoPrism~\citep{hu2026histoprism}. HiST~\citep{wu2026hist}, HEXST~\citep{byeon2026hexst}, and STFlow~\citep{huang2025stflow} require substantially longer training. Inference for CGL-Linear remains below \(0.1\) seconds throughout the evaluated range. Together with its strong predictive performance, this efficiency suggests that CGL-Linear could serve as a fast and lightweight reference baseline for H\&E-to-spatial-transcriptomics research.

\subsection{Transfer to existing models}
\label{sec:cgl_transfer}

We next examine whether within-slide supervision transfers to existing H\&E-to-spatial-transcriptomics models whose training objectives include a squared-error regression component. We augment HiST~\citep{wu2026hist}, HistoPrism~\citep{hu2026histoprism}, HEXST~\citep{byeon2026hexst}, STFlow~\citep{huang2025stflow}, and SEPAL~\citep{mejia2023sepal} using
\[
\mathcal L_{\mathrm{aug}}
=
\mathcal L_{\mathrm{base}}
+
\lambda\mathcal L_C.
\]
Each method retains its original architecture and full training objective, with \(\mathcal L_C\) added as an auxiliary within-slide term. The component weight is selected using validation data for each method. Additional implementation details are given in Appendix~\ref{app:cgl-transfer}.

\begin{figure*}[htbp]
\centering
\includegraphics[width=\textwidth]{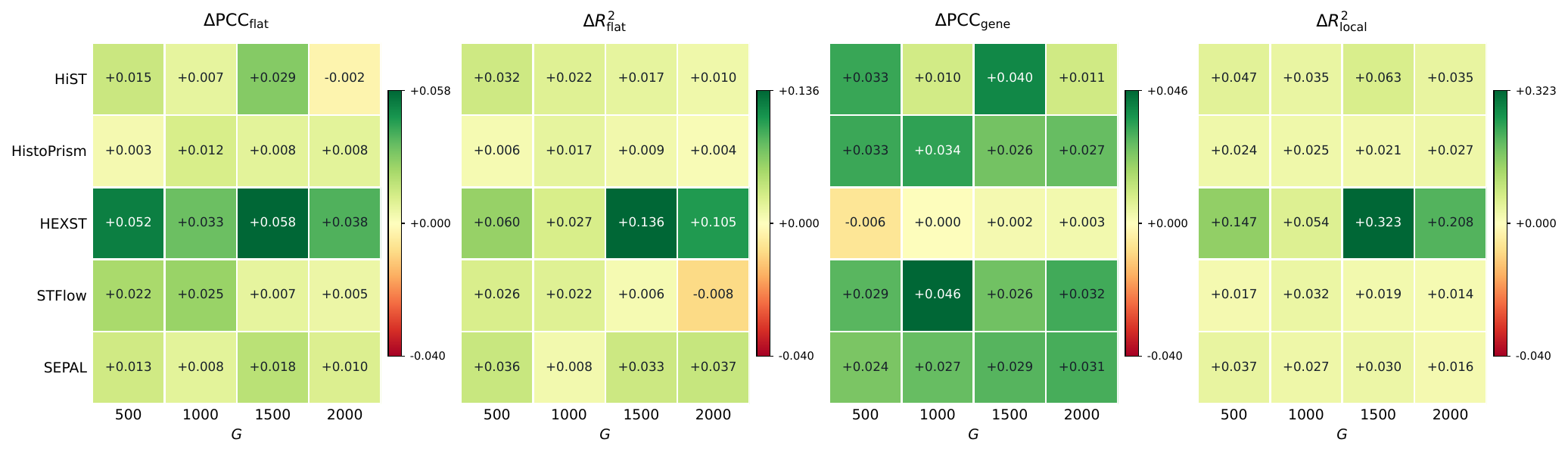}
\caption{\textbf{Adding within-slide supervision to existing models.}
Cells show equal-cohort macro test gains over the corresponding base method across ten cohorts using DINOv2 representations. Positive values favor the augmented objective.}
\label{fig:cgl_transfer}
\end{figure*}

Figure~\ref{fig:cgl_transfer} shows positive Local \(R^2\) gains for every method and gene panel evaluated. Most settings also improve the aggregate and gene-wise metrics, with only small regressions in a few comparisons. The effect therefore extends beyond the affine readout and the controlled MLP to substantially different model families.

The result is also informative for methods that already model expression variation. HEXST~\citep{byeon2026hexst} shows large improvements in Local \(R^2\) while Gene PCC changes little. SEPAL~\citep{mejia2023sepal} also improves despite explicitly predicting deviations from training-set gene means. Its deviation target removes a cohort-level gene baseline, whereas \(\mathcal L_C\) measures prediction error around each slide mean. The two forms of supervision are therefore distinct.

Across hierarchical Transformers, graph-based predictors, and flow-based models, the same within-slide term can be incorporated without changing the underlying architecture or replacing the original objective. These results demonstrate the practical flexibility of component-guided supervision across different H\&E-to-spatial-transcriptomics model designs that already rely on squared-error learning.
\subsection{Spatial predictions}
\label{sec:qualitative}

We inspect two tissue-relevant genes to relate the quantitative results to spatial expression patterns. MUC2 encodes a gel-forming mucin secreted by intestinal goblet cells and is a major component of the colonic mucus layer~\citep{johansson2011muc2}. MYH6 encodes the cardiac \(\alpha\)-myosin heavy chain, a major contractile myosin isoform with predominant expression in human atrial myocardium~\citep{reiser2001myosin,barrick2021cardiac}. We examine MUC2 in held-out colon slide MISC71 and MYH6 in held-out heart slide MISC138.

\begin{figure*}[htbp]
\centering
\includegraphics[width=\textwidth]{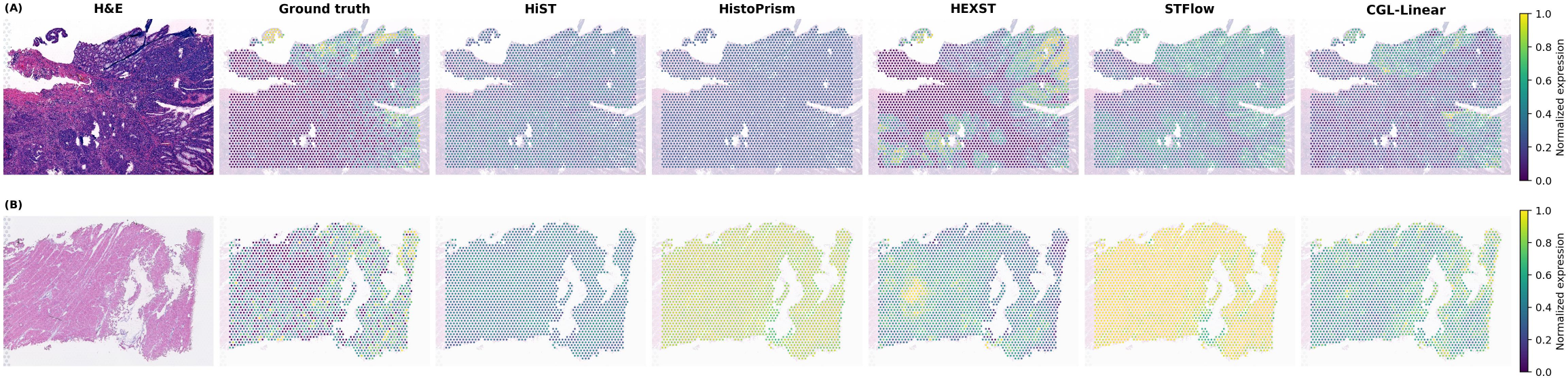}
\caption{\textbf{Spatial expression predictions on held-out slides.}
Row (A) shows MUC2 in colon slide MISC71, and row (B) shows MYH6 in heart slide MISC138. Each row contains the H\&E image, measured expression, and predictions from the displayed methods.}
\label{fig:qualitative}
\end{figure*}

For MUC2, the measured expression contains localized high-expression regions along the upper and right portions of the tissue. Several predictions capture the broad pattern but attenuate its contrast or place elevated expression in different regions. CGL-Linear follows the measured spatial arrangement more closely in this example and preserves the separation between these regions and the lower-expression interior.

MYH6 exhibits a broader expression pattern with marked regional variation across the heart section. Some methods produce relatively uniform predictions or shift the regions of high expression. CGL-Linear retains more of the measured contrast between lower-expression regions on the left and higher-expression regions toward the upper and right parts of the section. These examples complement the quantitative results by showing how differences in within-slide reconstruction appear spatially.

\FloatBarrier
\section{Conclusion}

We studied H\&E-to-spatial-transcriptomics prediction by separating mean-squared error into slide-level and within-slide components. This view reveals that strong aggregate prediction can coexist with weak within-slide recovery and, in controlled neural experiments, that the within-slide component is less responsive to shared parameters under standard MSE training. Component-Guided Loss directly increases supervision of this component. Its closed-form affine instantiation, CGL-Linear, substantially improves over ordinary linear regression, achieves overall state-of-the-art performance across HEST-1k cohorts, and could serve as a fast and lightweight reference baseline. The same within-slide supervision also improves several existing model families. We hope this perspective provides a basis for assessing how much additional predictive signal can be recovered by better exploiting the structure of the prediction problem before resorting to additional model complexity.

The current analysis is specific to squared-error prediction objectives. The exact decomposition and sensitivity characterization are derived for MSE, and the empirical sensitivity imbalance is established in controlled MSE-trained models. Although adding the within-slide term also improves models with more complex objectives, we do not establish that the same optimization behavior holds for arbitrary training losses. Extending the component-level analysis beyond squared-error objectives is therefore an important direction for future work.

\FloatBarrier
\label{main:end}

\clearpage

\section*{Reproducibility Statement}

Appendix~\ref{app:math} derives the error decomposition, sensitivity identities, and CGL-Linear estimator. Appendix~\ref{app:protocol} describes experimental configurations. Appendix~\ref{app:cohorts} reports detailed cohort results.

\section*{AI Use Statement}

Generative AI tools were used to assist with literature retrieval and discovery, and to improve the clarity and readability of the manuscript. All AI-assisted writing and retrieved information were reviewed and verified by the authors. The authors take responsibility for the final content of this work.

\clearpage

\bibliography{references}
\bibliographystyle{iclr2027_conference}

\clearpage
\appendix

\section{Mathematical details}
\label{app:math}

We use the notation of Section~\ref{sec:problem} and write \(N=MG\). Observations and slide membership remain fixed during differentiation.

\subsection{Slide-wise projections}
\label{app:projection-properties}

Each block \(\mathbf P_s=n_s^{-1}\mathbf1_{n_s}\mathbf1_{n_s}^{\top}\) is symmetric, and
\[
\mathbf P_s^2
=\frac{\mathbf1_{n_s}(\mathbf1_{n_s}^{\top}\mathbf1_{n_s})\mathbf1_{n_s}^{\top}}{n_s^2}
=\mathbf P_s.
\]
Consequently, \(\mathbf P\) is an orthogonal projector. Its complement \(\mathbf C=\mathbf I-\mathbf P\) satisfies \(\mathbf C^{\top}=\mathbf C\), \(\mathbf C^2=\mathbf C\), and \(\mathbf P\mathbf C=\mathbf C\mathbf P=\mathbf0\). For any matrix \(\mathbf Z\),
\(\langle\mathbf P\mathbf Z,\mathbf C\mathbf Z\rangle_F=\operatorname{tr}(\mathbf Z^{\top}\mathbf P\mathbf C\mathbf Z)=0\). Thus
\begin{equation}
\|\mathbf Z\|_F^2=\|\mathbf P\mathbf Z\|_F^2+\|\mathbf C\mathbf Z\|_F^2.
\label{eq:app-projection}
\end{equation}
Taking \(\mathbf Z=\mathbf E\) proves Eq.~\eqref{eq:method-error-decomposition}.

Let \(\mathbf e_{si}=\widehat{\mathbf y}_{si}-\mathbf y_{si}\) and \(\bar{\mathbf e}_s=n_s^{-1}\sum_i\mathbf e_{si}\). Then
\begin{equation}
\mathcal L_P=\frac1N\sum_s n_s\|\bar{\mathbf e}_s\|_2^2,
\qquad
\mathcal L_C=\frac1N\sum_s\sum_i\|\mathbf e_{si}-\bar{\mathbf e}_s\|_2^2.
\label{eq:app-component-means}
\end{equation}
These expressions separate error in slide-wise means from error in within-slide deviations. If \(\mathbf P\mathbf D=\mathbf D\), then \(\mathbf C\mathbf D=\mathbf0\), so \(\mathbf C(\mathbf Y+\mathbf D)=\mathbf C\mathbf Y\).

For MeanOnly, \(\mathbf C\widehat{\mathbf Y}=\mathbf0\). Therefore \(\mathbf C(\widehat{\mathbf Y}-\mathbf Y)=-\mathbf C\mathbf Y\), giving \(R^2_{\mathrm{local}}=0\) whenever \(\|\mathbf C\mathbf Y\|_F^2>0\). This identity is independent of how close the training gene means are to those in the evaluated split.

\subsection{Parameter sensitivity}
\label{app:optimization-sensitivity}

Use column-wise vectorization and let \(\boldsymbol\Pi_P=\mathbf I_G\otimes\mathbf P\), \(\boldsymbol\Pi_C=\mathbf I_G\otimes\mathbf C\), and \(\mathbf e_j=\boldsymbol\Pi_j\operatorname{vec}(\mathbf E)\). With \(\mathbf J_\theta=\partial\operatorname{vec}(\widehat{\mathbf Y}_\theta)/\partial\theta\), differentiation gives
\begin{equation}
\mathbf g_j=\nabla_\theta\mathcal L_j=\frac2N\mathbf J_\theta^{\top}\mathbf e_j,
\qquad j\in\{P,C\}.
\label{eq:app-gradient}
\end{equation}
For \(\mathcal L_j>0\), the definition of \(\kappa_j\) implies
\begin{equation}
\|\mathbf g_j\|_2^2=\frac4N\mathcal L_j\kappa_j,
\qquad
\frac{\|\mathbf g_P\|_2^2}{\|\mathbf g_C\|_2^2}
=\frac{\mathcal L_P}{\mathcal L_C}\frac{\kappa_P}{\kappa_C},
\label{eq:app-sensitivity}
\end{equation}
where the ratio requires both component errors to be nonzero and \(\|\mathbf g_C\|_2>0\). If \(\mathcal L_j=0\), then \(\mathbf g_j=\mathbf0\) and the normalized sensitivity is undefined. Orthogonal residuals need not yield orthogonal gradients because
\(\langle\mathbf g_P,\mathbf g_C\rangle=4\mathbf e_P^{\top}\mathbf J_\theta\mathbf J_\theta^{\top}\mathbf e_C/N^2\).

The grouping gives a useful interpretation. Let \(\mathbf J_{si}=\partial\widehat{\mathbf y}_{si}/\partial\theta\) and \(\bar{\mathbf J}_s=n_s^{-1}\sum_i\mathbf J_{si}\). Since \(\sum_i(\mathbf e_{si}-\bar{\mathbf e}_s)=\mathbf0\), Eq.~\eqref{eq:app-gradient} becomes
\begin{align}
\mathbf g_P&=\frac2N\sum_s n_s\bar{\mathbf J}_s^{\top}\bar{\mathbf e}_s,\label{eq:app-mean-gradient}\\
\mathbf g_C&=\frac2N\sum_s\sum_i(\mathbf J_{si}-\bar{\mathbf J}_s)^{\top}(\mathbf e_{si}-\bar{\mathbf e}_s).\label{eq:app-centered-gradient}
\end{align}
Parameter responses common to a slide cancel from its contribution to \(\mathbf g_C\). If responses are identical within every slide, \(\mathbf g_C=\mathbf0\) even when within-slide error is nonzero. In general, the relative sensitivities depend on the response geometry and its alignment with the residuals. Slide membership alone does not determine their ordering.

Under standard MSE gradient flow, \(\dot\theta=-(\mathbf g_P+\mathbf g_C)\). Applying the chain rule yields
\begin{align}
\dot{\mathcal L}_P&=-\|\mathbf g_P\|_2^2-\langle\mathbf g_P,\mathbf g_C\rangle,\notag\\
\dot{\mathcal L}_C&=-\|\mathbf g_C\|_2^2-\langle\mathbf g_P,\mathbf g_C\rangle.
\label{eq:app-flow}
\end{align}
Hence \((-\dot{\mathcal L}_P)-(-\dot{\mathcal L}_C)=\|\mathbf g_P\|_2^2-\|\mathbf g_C\|_2^2\). The sensitivity \(\kappa_j\) determines the self-descent contribution per unit component error. The full relative decrease also depends on \(\langle\mathbf g_P,\mathbf g_C\rangle/\mathcal L_j\).

\subsection{Component reweighting}
\label{app:cgl-properties}

At a fixed parameter state, let \(\mathbf v_\lambda=-\mathbf g_P-(1+\lambda)\mathbf g_C\). Define \(D_j(\lambda)=-\langle\mathbf g_j,\mathbf v_\lambda\rangle\), the instantaneous decrease rate of the unweighted component \(\mathcal L_j\) along this direction. Direct subtraction gives
\begin{align}
D_C(\lambda)-D_C(0)&=\lambda\|\mathbf g_C\|_2^2,\label{eq:app-local-rate}\\
D_P(\lambda)-D_P(0)&=\lambda\langle\mathbf g_P,\mathbf g_C\rangle.\label{eq:app-slide-rate}
\end{align}
For \(\lambda\geq0\), CGL adds a nonnegative within-slide descent contribution. The change in slide-level progress depends on gradient alignment. This comparison holds at the same parameter state under gradient flow. It does not order complete training trajectories or held-out errors.

\subsection{CGL-Linear}
\label{app:weighted-linear}

All fitting quantities refer to the training set. Write \(\mathbf W_\lambda=\mathbf I+\lambda\mathbf C\) and \(\mathbf E_{\mathbf A,\mathbf b}=\mathbf Q\mathbf A+\mathbf1\mathbf b^{\top}-\mathbf Y\). The regularized affine problem is
\begin{equation}
\min_{\mathbf A,\mathbf b}
\langle\mathbf E_{\mathbf A,\mathbf b},\mathbf W_\lambda\mathbf E_{\mathbf A,\mathbf b}\rangle_F
+\alpha_{\mathrm{raw}}\|\mathbf A\|_F^2.
\label{eq:app-affine-fit}
\end{equation}
The intercept is unregularized. Since \(\mathbf W_\lambda\mathbf1=\mathbf1\), its first-order condition is \(\mathbf E_{\mathbf A,\mathbf b}^{\top}\mathbf1=\mathbf0\). This gives \(\mathbf b=\boldsymbol\mu_Y-\mathbf A^{\top}\boldsymbol\mu_Q\), with \(\boldsymbol\mu_Q=\mathbf Q^{\top}\mathbf1/M\) and \(\boldsymbol\mu_Y=\mathbf Y^{\top}\mathbf1/M\).

Substituting \(\mathbf X=\mathbf Q-\mathbf1\boldsymbol\mu_Q^{\top}\) and \(\mathbf Y_c=\mathbf Y-\mathbf1\boldsymbol\mu_Y^{\top}\) reduces Eq.~\eqref{eq:app-affine-fit} to
\begin{equation}
\min_{\mathbf A}
\|\mathbf P(\mathbf X\mathbf A-\mathbf Y_c)\|_F^2
+(1+\lambda)\|\mathbf C(\mathbf X\mathbf A-\mathbf Y_c)\|_F^2
+\alpha_{\mathrm{raw}}\|\mathbf A\|_F^2.
\label{eq:app-centered-fit}
\end{equation}
Its stationarity condition is
\((\mathbf X^{\top}\mathbf W_\lambda\mathbf X+\alpha_{\mathrm{raw}}\mathbf I)\mathbf A=\mathbf X^{\top}\mathbf W_\lambda\mathbf Y_c\), yielding
\begin{equation}
\mathbf A_{\lambda,\alpha}
=\left[\mathbf X^{\top}(\mathbf I+\lambda\mathbf C)\mathbf X+\alpha_{\mathrm{raw}}\mathbf I\right]^{-1}
\mathbf X^{\top}(\mathbf I+\lambda\mathbf C)\mathbf Y_c.
\label{eq:app-linear-solution}
\end{equation}
For \(\lambda\geq0\) and \(\alpha_{\mathrm{raw}}>0\), the coefficient matrix is positive definite, so the solution is unique. The inverse notation represents a linear solve. Dividing Eq.~\eqref{eq:app-centered-fit} by \(N\) gives the normalized CGL objective plus \((\alpha_{\mathrm{raw}}/N)\|\mathbf A\|_F^2\).

The implemented penalty convention is \(\alpha_{\mathrm{raw}}=\alpha\operatorname{tr}(\mathbf X^{\top}\mathbf X)/d\), computed from unweighted training features independently of \(\lambda\). At \(\lambda=0\), the estimator is ordinary affine ridge regression. For a new feature vector \(\mathbf q_*\), prediction is \(\widehat{\mathbf y}_*=\boldsymbol\mu_Y+\mathbf A_{\lambda,\alpha}^{\top}(\mathbf q_*-\boldsymbol\mu_Q)\). All fitted quantities come from training data; neither slide projection is required at inference.

\clearpage
\section{Experimental protocol}
\label{app:protocol}

\paragraph{Data preparation}

The benchmark comprises ten organ-level cohorts derived from HEST-1k~\citep{jaume2024hest}. A separate model is fitted for each cohort. Fixed biological-unit-disjoint splits separate training, validation, and test data. Table~\ref{tab:data} lists the aligned slides and locations.

\begin{table}[H]
\caption{Cohort sizes in the fixed canonical split. Slide counts are recovered from the $500$-gene evaluation support; spot counts are the aligned H\&E--expression pairs used by the experiments.}
\label{tab:data}
\centering
\small
\setlength{\tabcolsep}{4.1pt}
\begin{tabular}{lrrrrrr}
\toprule
Cohort & \multicolumn{3}{c}{Slides (train / val / test)} & \multicolumn{3}{c}{Spots (train / val / test)} \\
\midrule
Brain & 9 & 2 & 3 & 23,780 & 4,891 & 4,219 \\
Colon & 33 & 4 & 4 & 94,871 & 6,359 & 15,138 \\
Heart & 26 & 7 & 8 & 70,802 & 14,100 & 14,195 \\
Kidney & 24 & 5 & 5 & 27,739 & 5,649 & 5,691 \\
Liver & 10 & 2 & 2 & 26,518 & 5,695 & 4,955 \\
Lung & 12 & 4 & 4 & 20,810 & 8,763 & 8,933 \\
Prostate & 23 & 2 & 2 & 62,481 & 7,898 & 7,784 \\
Skin & 42 & 7 & 10 & 36,585 & 7,984 & 7,488 \\
Uterus & 10 & 3 & 2 & 12,503 & 2,090 & 2,351 \\
Breast & 65 & 15 & 15 & 26,752 & 5,988 & 5,920 \\
\midrule
Total & 254 & 51 & 55 & 402,841 & 69,417 & 76,674 \\
\bottomrule
\end{tabular}
\end{table}

 Highly variable genes are selected from training data using \texttt{seurat\_v3}, with slide identity as the batch key. The panels of 500, 1,000, 1,500, and 2,000 genes are nested. All compared methods use these same panels and splits. Scores therefore refer to this common benchmark, not the original HEST-Benchmark configuration.

\paragraph{Image representations}

Frozen UNI, DINOv2-Large, and ResNet18 representations have dimensions 1,024, 1,024, and 512, respectively. Features are extracted once per encoder and shared across the compared methods. Each method retains its downstream processing. 

\paragraph{Model selection}
\label{app:selection}

Model configurations are selected with DINOv2-Large using equal-cohort macro validation MSE, then held fixed for UNI and ResNet18. Checkpoints are selected using validation MSE. The main neural benchmark uses seeds \(0,1,2\). Linear and CGL-Linear are deterministic once their hyperparameters have been selected. Test scores do not enter model selection.

Both affine estimators use the penalty convention in Appendix~\ref{app:weighted-linear}. Linear has \(\lambda=0\); CGL-Linear uses a validation-selected \(\lambda\geq0\). Regularization is also selected on validation data. Table~\ref{tab:best-configs} summarizes the model configurations.

\begin{table*}[h]
\centering
\footnotesize
\setlength{\tabcolsep}{4pt}
\renewcommand{\arraystretch}{1.10}
\caption{Configurations of the compared predictors. LR denotes learning rate and WD denotes weight decay.}
\label{tab:best-configs}
\begin{tabularx}{\textwidth}{@{}p{1.45cm}p{1.1cm}XX@{}}
\toprule
Method & Epochs & Readout configuration & Optimization \\
\midrule
HiST & 300 & Sparse hierarchy 111--21, window size 3, feed-forward ratio 4, three global tokens. & AdamW, LR \(10^{-4}\), WD \(10^{-5}\), cosine schedule, centered-target MSE. \\
HistoPrism & 300 & Hidden size 256, two Transformer layers, eight heads, dropout 0.1. & AdamW, LR \(5\times10^{-5}\), WD \(10^{-2}\), linear decay with 5\% warm-up, centered-target MSE. \\
HEXST & 300 & Hidden size 512, HexMSA/HexRoPE, 3,072-dimensional scFoundation alignment, dropout 0.2. & Adam, LR \(4\times10^{-5}\), WD \(10^{-4}\), warm-up and cosine schedule, composite regression/correlation/deviation/alignment objective. \\
STFlow & 300 & Hidden size 128, four layers, four heads, eight neighbors, five sampling steps. & Adam, LR \(5\times10^{-4}\), gradient clipping at 1.0, no scheduler, conditional flow-matching objective. \\
BLEEP & 300 & Dual 256-dimensional projections, dropout 0.1, temperature 1, top-50 retrieval. & AdamW, LR \(10^{-3}\), WD \(10^{-3}\), batch size 512, symmetric contrastive cross-entropy. \\
SEPAL & \(300\) & Image readout followed by a six-neighbor graph head, relative positional encoding, two GCN layers, SAGPooling. & SGD, LR 0.0099913, momentum 0.9, batch size 32, MSE in both stages, frozen first-stage predictor during graph fitting. \\
\midrule

CGL-Linear & -- & One affine readout. & Direct CGL solve, $\lambda=5$ and $\alpha=5$. \\
\bottomrule
\end{tabularx}
\end{table*}

\paragraph{Optimization controls}
\label{app:optimization-control}

The controls use a single hidden layer with 256 units, GELU activation, and an affine output layer. Frozen DINOv2-Large and ResNet18 features predict \(G=500\) genes. The MSE and CGL runs start from identical parameters and use full-gradient Adam at learning rate \(10^{-3}\) for 600 updates, with seed 0 and no parameter weight decay. The output bias is initialized with training gene means. CGL uses \(\lambda=30\), chosen on validation data.

Component losses use the normalization in Eq.~\eqref{eq:method-error-decomposition}. At each recorded state, the gradients of \(\mathcal L_P\) and \(\mathcal L_C\) are evaluated on the complete training or validation split. Validation gradients are diagnostic only and do not update the predictor. Sensitivity curves summarize cohort medians and interquartile ranges. Loss-difference curves summarize equal-cohort means and interquartile ranges. These bands describe variation across cohorts, not confidence intervals. The gradient-flow identities describe Euclidean gradient directions; the observed Adam trajectories are evaluated empirically.
\FloatBarrier
\subsection{Neural augmentation}
\label{app:cgl-transfer}

The transfer experiments use DINOv2-Large across the four gene panels. HiST, HistoPrism, HEXST, STFlow, and SEPAL retain their original architectures and training objectives, with the within-slide term added as
\[
\mathcal L_{\mathrm{aug}}
=
\mathcal L_{\mathrm{base}}
+
\lambda \mathcal L_C.
\]
For each method, we search \(\lambda\in\{5,10,15,20\}\) and select the value with the lowest equal-cohort macro validation MSE. Test performance is reported at the corresponding validation-selected checkpoint.

\begin{table}[h]
\centering
\small
\caption{Validation-selected component weights used in the neural augmentation experiments.}
\label{tab:cgl-transfer-lambda}
\begin{tabular}{lc}
\toprule
Method & \(\lambda\) \\
\midrule
HiST        & 10 \\
HistoPrism  & 15 \\
HEXST       & 20 \\
STFlow      & 10 \\
SEPAL       & 5 \\
\bottomrule
\end{tabular}
\end{table}

The reported heatmap compares each augmented model with its corresponding base model across the four gene panels, yielding 20 paired method--panel comparisons.

\clearpage
\section{Cohort-level results}
\label{app:cohorts}

The following tables report the four evaluation metrics for all three encoders and four gene panels. Neural entries report mean and standard deviation over three seeds (0, 1, 2). Linear and CGL-Linear are deterministic. Macro rows average cohort means equally. Bold and underline identify the best and second-best unrounded means within each row. Display rounding can produce apparent ties. All values are generated from the accompanying benchmark CSV.

\begin{table*}[!ht]
\centering
\caption{Cohort-level test results with UNI and \(G=500\). Neural entries are mean \(\pm\) seed standard deviation over three runs. Linear and CGL-Linear are deterministic. Bold and underlining mark the best and second-best unrounded means in each row. Macro rows average cohort means without weighting by cohort size. Higher is better.}
\label{tab:cohort-uni-500}
\scriptsize
\setlength{\tabcolsep}{2.0pt}
\renewcommand{\arraystretch}{1.08}
\resizebox{\textwidth}{!}{%
\begin{tabular}{lrrrrrrrr}
\toprule
Cohort & HiST & HistoPrism & HEXST & STFlow & BLEEP & SEPAL & Linear & CGL-Linear \\
\midrule
\multicolumn{9}{l}{\(\mathrm{PCC}_{\mathrm{flat}}\;\uparrow\)} \\
Brain & $\underline{0.775}\,\pm\,0.001$ & $0.772\,\pm\,0.004$ & $0.764\,\pm\,0.001$ & $0.754\,\pm\,0.010$ & $0.702\,\pm\,0.026$ & $0.762\,\pm\,0.008$ & $0.735$ & $\mathbf{0.776}$ \\
Colon & $\underline{0.744}\,\pm\,0.018$ & $0.731\,\pm\,0.009$ & $0.719\,\pm\,0.003$ & $0.710\,\pm\,0.002$ & $0.622\,\pm\,0.068$ & $0.713\,\pm\,0.011$ & $0.713$ & $\mathbf{0.758}$ \\
Heart & $0.789\,\pm\,0.009$ & $\mathbf{0.804}\,\pm\,0.007$ & $0.647\,\pm\,0.002$ & $\underline{0.791}\,\pm\,0.005$ & $0.771\,\pm\,0.009$ & $0.768\,\pm\,0.018$ & $0.762$ & $0.736$ \\
Kidney & $0.734\,\pm\,0.006$ & $0.714\,\pm\,0.000$ & $0.719\,\pm\,0.000$ & $0.676\,\pm\,0.004$ & $0.702\,\pm\,0.014$ & $\underline{0.745}\,\pm\,0.011$ & $0.727$ & $\mathbf{0.747}$ \\
Liver & $0.864\,\pm\,0.003$ & $0.871\,\pm\,0.002$ & $0.858\,\pm\,0.001$ & $\underline{0.872}\,\pm\,0.000$ & $0.858\,\pm\,0.001$ & $0.858\,\pm\,0.004$ & $0.862$ & $\mathbf{0.874}$ \\
Lung & $0.622\,\pm\,0.018$ & $0.593\,\pm\,0.037$ & $0.655\,\pm\,0.003$ & $0.645\,\pm\,0.043$ & $0.564\,\pm\,0.006$ & $\underline{0.663}\,\pm\,0.005$ & $\mathbf{0.663}$ & $0.658$ \\
Prostate & $0.659\,\pm\,0.003$ & $0.685\,\pm\,0.013$ & $\mathbf{0.726}\,\pm\,0.003$ & $0.545\,\pm\,0.068$ & $0.476\,\pm\,0.022$ & $0.701\,\pm\,0.023$ & $0.591$ & $\underline{0.716}$ \\
Skin & $0.680\,\pm\,0.009$ & $\mathbf{0.710}\,\pm\,0.007$ & $0.516\,\pm\,0.008$ & $0.659\,\pm\,0.006$ & $0.640\,\pm\,0.005$ & $0.667\,\pm\,0.039$ & $0.676$ & $\underline{0.707}$ \\
Uterus & $0.827\,\pm\,0.019$ & $0.817\,\pm\,0.011$ & $0.849\,\pm\,0.003$ & $0.805\,\pm\,0.028$ & $0.807\,\pm\,0.010$ & $\underline{0.866}\,\pm\,0.006$ & $0.835$ & $\mathbf{0.871}$ \\
Breast & $0.669\,\pm\,0.003$ & $0.676\,\pm\,0.006$ & $0.674\,\pm\,0.000$ & $0.596\,\pm\,0.016$ & $0.624\,\pm\,0.012$ & $\mathbf{0.707}\,\pm\,0.009$ & $0.661$ & $\underline{0.681}$ \\
\cmidrule(lr){1-9}
\textit{Macro} & $0.736$ & $0.737$ & $0.713$ & $0.705$ & $0.677$ & $\underline{0.745}$ & $0.723$ & $\mathbf{0.752}$ \\
\midrule
\multicolumn{9}{l}{\(\mathrm{PCC}_{\mathrm{gene}}\;\uparrow\)} \\
Brain & $\underline{0.045}\,\pm\,0.004$ & $0.018\,\pm\,0.005$ & $0.037\,\pm\,0.013$ & $0.040\,\pm\,0.003$ & $0.040\,\pm\,0.004$ & $0.010\,\pm\,0.000$ & $0.040$ & $\mathbf{0.051}$ \\
Colon & $0.205\,\pm\,0.004$ & $0.198\,\pm\,0.003$ & $\mathbf{0.248}\,\pm\,0.008$ & $0.185\,\pm\,0.002$ & $0.142\,\pm\,0.018$ & $0.184\,\pm\,0.022$ & $0.195$ & $\underline{0.227}$ \\
Heart & $0.053\,\pm\,0.004$ & $0.050\,\pm\,0.006$ & $0.042\,\pm\,0.007$ & $\mathbf{0.069}\,\pm\,0.001$ & $0.048\,\pm\,0.001$ & $0.036\,\pm\,0.003$ & $0.058$ & $\underline{0.061}$ \\
Kidney & $\mathbf{0.105}\,\pm\,0.010$ & $0.088\,\pm\,0.015$ & $0.099\,\pm\,0.006$ & $0.077\,\pm\,0.001$ & $0.087\,\pm\,0.009$ & $0.081\,\pm\,0.024$ & $0.092$ & $\underline{0.105}$ \\
Liver & $0.091\,\pm\,0.001$ & $0.082\,\pm\,0.004$ & $0.039\,\pm\,0.002$ & $\underline{0.104}\,\pm\,0.010$ & $0.088\,\pm\,0.003$ & $0.062\,\pm\,0.004$ & $0.088$ & $\mathbf{0.107}$ \\
Lung & $0.231\,\pm\,0.011$ & $0.235\,\pm\,0.006$ & $\underline{0.332}\,\pm\,0.024$ & $0.313\,\pm\,0.037$ & $0.220\,\pm\,0.004$ & $0.296\,\pm\,0.052$ & $0.288$ & $\mathbf{0.347}$ \\
Prostate & $0.045\,\pm\,0.002$ & $\underline{0.074}\,\pm\,0.018$ & $0.071\,\pm\,0.033$ & $0.027\,\pm\,0.021$ & $0.040\,\pm\,0.009$ & $\mathbf{0.075}\,\pm\,0.006$ & $0.002$ & $0.074$ \\
Skin & $0.390\,\pm\,0.001$ & $0.416\,\pm\,0.014$ & $\mathbf{0.453}\,\pm\,0.016$ & $0.401\,\pm\,0.012$ & $0.397\,\pm\,0.002$ & $0.385\,\pm\,0.046$ & $0.409$ & $\underline{0.444}$ \\
Uterus & $0.092\,\pm\,0.022$ & $0.078\,\pm\,0.004$ & $\underline{0.137}\,\pm\,0.003$ & $0.095\,\pm\,0.021$ & $0.105\,\pm\,0.008$ & $0.095\,\pm\,0.003$ & $0.111$ & $\mathbf{0.155}$ \\
Breast & $0.086\,\pm\,0.004$ & $0.099\,\pm\,0.007$ & $\underline{0.103}\,\pm\,0.004$ & $0.054\,\pm\,0.015$ & $0.057\,\pm\,0.011$ & $\mathbf{0.161}\,\pm\,0.027$ & $0.070$ & $0.085$ \\
\cmidrule(lr){1-9}
\textit{Macro} & $0.134$ & $0.134$ & $\underline{0.156}$ & $0.137$ & $0.122$ & $0.138$ & $0.135$ & $\mathbf{0.166}$ \\
\midrule
\multicolumn{9}{l}{\(R^2_{\mathrm{local}}\;\uparrow\)} \\
Brain & $-0.081\,\pm\,0.015$ & $-0.040\,\pm\,0.009$ & $\mathbf{0.012}\,\pm\,0.005$ & $-0.224\,\pm\,0.087$ & $-0.472\,\pm\,0.027$ & $\underline{0.002}\,\pm\,0.002$ & $-0.290$ & $-0.021$ \\
Colon & $\mathbf{0.158}\,\pm\,0.016$ & $0.120\,\pm\,0.015$ & $0.094\,\pm\,0.008$ & $0.028\,\pm\,0.002$ & $-0.215\,\pm\,0.034$ & $0.028\,\pm\,0.031$ & $-0.138$ & $\underline{0.153}$ \\
Heart & $-0.013\,\pm\,0.002$ & $\mathbf{0.017}\,\pm\,0.004$ & $\underline{0.007}\,\pm\,0.001$ & $-0.152\,\pm\,0.007$ & $-0.219\,\pm\,0.024$ & $-0.093\,\pm\,0.050$ & $-0.213$ & $-0.020$ \\
Kidney & $0.051\,\pm\,0.015$ & $\underline{0.059}\,\pm\,0.010$ & $0.052\,\pm\,0.007$ & $-0.174\,\pm\,0.052$ & $-0.122\,\pm\,0.023$ & $0.026\,\pm\,0.016$ & $-0.086$ & $\mathbf{0.076}$ \\
Liver & $0.019\,\pm\,0.031$ & $0.030\,\pm\,0.011$ & $0.011\,\pm\,0.012$ & $\mathbf{0.063}\,\pm\,0.008$ & $-0.100\,\pm\,0.027$ & $0.031\,\pm\,0.000$ & $-0.049$ & $\underline{0.055}$ \\
Lung & $0.194\,\pm\,0.031$ & $0.115\,\pm\,0.015$ & $0.119\,\pm\,0.007$ & $\underline{0.197}\,\pm\,0.006$ & $0.096\,\pm\,0.014$ & $0.194\,\pm\,0.050$ & $0.161$ & $\mathbf{0.234}$ \\
Prostate & $-0.019\,\pm\,0.012$ & $\underline{0.023}\,\pm\,0.018$ & $0.008\,\pm\,0.009$ & $-0.121\,\pm\,0.162$ & $-0.252\,\pm\,0.348$ & $0.016\,\pm\,0.007$ & $-0.127$ & $\mathbf{0.023}$ \\
Skin & $\underline{0.452}\,\pm\,0.005$ & $\mathbf{0.456}\,\pm\,0.014$ & $0.103\,\pm\,0.013$ & $0.360\,\pm\,0.007$ & $0.325\,\pm\,0.024$ & $0.375\,\pm\,0.055$ & $0.380$ & $0.430$ \\
Uterus & $0.049\,\pm\,0.010$ & $0.080\,\pm\,0.004$ & $0.070\,\pm\,0.008$ & $-0.151\,\pm\,0.002$ & $-0.092\,\pm\,0.063$ & $\underline{0.088}\,\pm\,0.022$ & $-0.006$ & $\mathbf{0.154}$ \\
Breast & $-0.005\,\pm\,0.002$ & $\mathbf{0.061}\,\pm\,0.001$ & $0.020\,\pm\,0.001$ & $-0.393\,\pm\,0.123$ & $-0.121\,\pm\,0.040$ & $0.051\,\pm\,0.005$ & $-0.055$ & $\underline{0.052}$ \\
\cmidrule(lr){1-9}
\textit{Macro} & $0.080$ & $\underline{0.092}$ & $0.050$ & $-0.057$ & $-0.117$ & $0.072$ & $-0.042$ & $\mathbf{0.114}$ \\
\midrule
\multicolumn{9}{l}{\(R^2_{\mathrm{flat}}\;\uparrow\)} \\
Brain & $0.593\,\pm\,0.004$ & $\underline{0.593}\,\pm\,0.004$ & $0.582\,\pm\,0.002$ & $0.548\,\pm\,0.036$ & $0.477\,\pm\,0.033$ & $0.580\,\pm\,0.011$ & $0.511$ & $\mathbf{0.599}$ \\
Colon & $\underline{0.549}\,\pm\,0.031$ & $0.506\,\pm\,0.026$ & $0.508\,\pm\,0.004$ & $0.460\,\pm\,0.007$ & $0.359\,\pm\,0.105$ & $0.502\,\pm\,0.023$ & $0.501$ & $\mathbf{0.568}$ \\
Heart & $0.554\,\pm\,0.005$ & $\mathbf{0.608}\,\pm\,0.008$ & $0.205\,\pm\,0.008$ & $\underline{0.583}\,\pm\,0.015$ & $0.574\,\pm\,0.016$ & $0.560\,\pm\,0.022$ & $0.505$ & $0.458$ \\
Kidney & $0.519\,\pm\,0.012$ & $0.477\,\pm\,0.001$ & $0.516\,\pm\,0.000$ & $0.413\,\pm\,0.009$ & $0.487\,\pm\,0.022$ & $\underline{0.547}\,\pm\,0.026$ & $0.522$ & $\mathbf{0.557}$ \\
Liver & $0.745\,\pm\,0.006$ & $\mathbf{0.757}\,\pm\,0.005$ & $0.727\,\pm\,0.001$ & $0.757\,\pm\,0.005$ & $0.732\,\pm\,0.005$ & $0.733\,\pm\,0.007$ & $0.740$ & $\underline{0.757}$ \\
Lung & $0.371\,\pm\,0.006$ & $0.346\,\pm\,0.050$ & $0.250\,\pm\,0.002$ & $\underline{0.396}\,\pm\,0.075$ & $0.166\,\pm\,0.050$ & $0.264\,\pm\,0.100$ & $\mathbf{0.432}$ & $0.353$ \\
Prostate & $0.180\,\pm\,0.021$ & $0.134\,\pm\,0.028$ & $\mathbf{0.346}\,\pm\,0.003$ & $0.033\,\pm\,0.031$ & $-0.010\,\pm\,0.168$ & $0.220\,\pm\,0.080$ & $0.186$ & $\underline{0.259}$ \\
Skin & $0.446\,\pm\,0.013$ & $\underline{0.500}\,\pm\,0.007$ & $0.260\,\pm\,0.008$ & $0.405\,\pm\,0.008$ & $0.393\,\pm\,0.014$ & $0.433\,\pm\,0.065$ & $0.443$ & $\mathbf{0.500}$ \\
Uterus & $0.679\,\pm\,0.027$ & $0.658\,\pm\,0.022$ & $0.701\,\pm\,0.004$ & $0.645\,\pm\,0.046$ & $0.635\,\pm\,0.002$ & $\underline{0.732}\,\pm\,0.013$ & $0.693$ & $\mathbf{0.735}$ \\
Breast & $0.431\,\pm\,0.011$ & $0.452\,\pm\,0.009$ & $0.454\,\pm\,0.001$ & $-0.004\,\pm\,0.089$ & $0.375\,\pm\,0.017$ & $\mathbf{0.499}\,\pm\,0.013$ & $0.421$ & $\underline{0.460}$ \\
\cmidrule(lr){1-9}
\textit{Macro} & $0.507$ & $0.503$ & $0.455$ & $0.424$ & $0.419$ & $\underline{0.507}$ & $0.495$ & $\mathbf{0.525}$ \\
\bottomrule
\end{tabular}%
}
\end{table*}

\clearpage
\begin{table*}[!ht]
\centering
\caption{Cohort-level test results with UNI and \(G=1000\). Neural entries are mean \(\pm\) seed standard deviation over three runs. Linear and CGL-Linear are deterministic. Bold and underlining mark the best and second-best unrounded means in each row. Macro rows average cohort means without weighting by cohort size. Higher is better.}
\label{tab:cohort-uni-1000}
\scriptsize
\setlength{\tabcolsep}{2.0pt}
\renewcommand{\arraystretch}{1.08}
\resizebox{\textwidth}{!}{%
\begin{tabular}{lrrrrrrrr}
\toprule
Cohort & HiST & HistoPrism & HEXST & STFlow & BLEEP & SEPAL & Linear & CGL-Linear \\
\midrule
\multicolumn{9}{l}{\(\mathrm{PCC}_{\mathrm{flat}}\;\uparrow\)} \\
Brain & $0.807\,\pm\,0.008$ & $\underline{0.808}\,\pm\,0.004$ & $0.800\,\pm\,0.001$ & $0.797\,\pm\,0.029$ & $0.748\,\pm\,0.003$ & $0.799\,\pm\,0.007$ & $0.779$ & $\mathbf{0.812}$ \\
Colon & $\underline{0.746}\,\pm\,0.011$ & $0.743\,\pm\,0.004$ & $0.716\,\pm\,0.011$ & $0.691\,\pm\,0.044$ & $0.659\,\pm\,0.002$ & $0.704\,\pm\,0.003$ & $0.717$ & $\mathbf{0.759}$ \\
Heart & $0.776\,\pm\,0.005$ & $\mathbf{0.787}\,\pm\,0.015$ & $0.674\,\pm\,0.001$ & $\underline{0.776}\,\pm\,0.005$ & $0.773\,\pm\,0.004$ & $0.760\,\pm\,0.005$ & $0.760$ & $0.747$ \\
Kidney & $\underline{0.720}\,\pm\,0.016$ & $0.708\,\pm\,0.001$ & $0.709\,\pm\,0.001$ & $0.680\,\pm\,0.015$ & $0.712\,\pm\,0.003$ & $0.719\,\pm\,0.020$ & $0.714$ & $\mathbf{0.733}$ \\
Liver & $0.861\,\pm\,0.000$ & $\underline{0.872}\,\pm\,0.005$ & $0.856\,\pm\,0.002$ & $0.868\,\pm\,0.002$ & $0.860\,\pm\,0.001$ & $0.859\,\pm\,0.001$ & $0.865$ & $\mathbf{0.874}$ \\
Lung & $0.605\,\pm\,0.028$ & $0.575\,\pm\,0.022$ & $0.589\,\pm\,0.054$ & $0.630\,\pm\,0.027$ & $0.557\,\pm\,0.014$ & $\mathbf{0.643}\,\pm\,0.001$ & $0.641$ & $\underline{0.642}$ \\
Prostate & $0.673\,\pm\,0.004$ & $0.667\,\pm\,0.002$ & $\mathbf{0.709}\,\pm\,0.004$ & $0.573\,\pm\,0.067$ & $0.421\,\pm\,0.022$ & $\underline{0.702}\,\pm\,0.001$ & $0.555$ & $0.694$ \\
Skin & $0.695\,\pm\,0.011$ & $\underline{0.710}\,\pm\,0.008$ & $0.494\,\pm\,0.010$ & $0.658\,\pm\,0.005$ & $0.603\,\pm\,0.032$ & $0.680\,\pm\,0.002$ & $0.684$ & $\mathbf{0.712}$ \\
Uterus & $0.824\,\pm\,0.005$ & $0.817\,\pm\,0.017$ & $0.751\,\pm\,0.012$ & $0.811\,\pm\,0.007$ & $0.809\,\pm\,0.013$ & $\underline{0.851}\,\pm\,0.002$ & $0.823$ & $\mathbf{0.858}$ \\
Breast & $0.638\,\pm\,0.010$ & $\underline{0.645}\,\pm\,0.009$ & $0.632\,\pm\,0.000$ & $0.565\,\pm\,0.005$ & $0.597\,\pm\,0.001$ & $\mathbf{0.667}\,\pm\,0.001$ & $0.620$ & $0.641$ \\
\cmidrule(lr){1-9}
\textit{Macro} & $0.734$ & $0.733$ & $0.693$ & $0.705$ & $0.674$ & $\underline{0.738}$ & $0.716$ & $\mathbf{0.747}$ \\
\midrule
\multicolumn{9}{l}{\(\mathrm{PCC}_{\mathrm{gene}}\;\uparrow\)} \\
Brain & $\underline{0.040}\,\pm\,0.007$ & $0.013\,\pm\,0.002$ & $0.034\,\pm\,0.007$ & $0.033\,\pm\,0.016$ & $0.030\,\pm\,0.004$ & $0.010\,\pm\,0.001$ & $0.037$ & $\mathbf{0.050}$ \\
Colon & $0.208\,\pm\,0.000$ & $0.194\,\pm\,0.006$ & $\mathbf{0.259}\,\pm\,0.003$ & $0.161\,\pm\,0.026$ & $0.156\,\pm\,0.012$ & $0.173\,\pm\,0.018$ & $0.193$ & $\underline{0.224}$ \\
Heart & $0.043\,\pm\,0.003$ & $0.047\,\pm\,0.012$ & $0.040\,\pm\,0.007$ & $\mathbf{0.053}\,\pm\,0.001$ & $0.046\,\pm\,0.005$ & $0.033\,\pm\,0.007$ & $0.049$ & $\underline{0.053}$ \\
Kidney & $\underline{0.086}\,\pm\,0.004$ & $0.079\,\pm\,0.009$ & $0.080\,\pm\,0.003$ & $0.069\,\pm\,0.004$ & $0.084\,\pm\,0.001$ & $0.055\,\pm\,0.020$ & $0.078$ & $\mathbf{0.089}$ \\
Liver & $0.083\,\pm\,0.002$ & $0.081\,\pm\,0.009$ & $0.051\,\pm\,0.001$ & $\underline{0.098}\,\pm\,0.002$ & $0.095\,\pm\,0.005$ & $0.057\,\pm\,0.002$ & $0.091$ & $\mathbf{0.109}$ \\
Lung & $0.231\,\pm\,0.014$ & $0.228\,\pm\,0.019$ & $\mathbf{0.328}\,\pm\,0.005$ & $0.282\,\pm\,0.036$ & $0.232\,\pm\,0.062$ & $0.253\,\pm\,0.021$ & $0.271$ & $\underline{0.326}$ \\
Prostate & $0.032\,\pm\,0.005$ & $0.039\,\pm\,0.019$ & $\underline{0.061}\,\pm\,0.028$ & $0.032\,\pm\,0.011$ & $0.019\,\pm\,0.008$ & $\mathbf{0.064}\,\pm\,0.022$ & $-0.015$ & $0.052$ \\
Skin & $0.355\,\pm\,0.001$ & $0.364\,\pm\,0.000$ & $\underline{0.412}\,\pm\,0.007$ & $0.352\,\pm\,0.014$ & $0.340\,\pm\,0.021$ & $0.331\,\pm\,0.000$ & $0.377$ & $\mathbf{0.412}$ \\
Uterus & $0.090\,\pm\,0.002$ & $0.070\,\pm\,0.010$ & $\underline{0.125}\,\pm\,0.006$ & $0.093\,\pm\,0.010$ & $0.093\,\pm\,0.000$ & $0.074\,\pm\,0.001$ & $0.091$ & $\mathbf{0.127}$ \\
Breast & $0.093\,\pm\,0.028$ & $\underline{0.097}\,\pm\,0.013$ & $0.093\,\pm\,0.008$ & $0.062\,\pm\,0.005$ & $0.056\,\pm\,0.002$ & $\mathbf{0.142}\,\pm\,0.004$ & $0.067$ & $0.082$ \\
\cmidrule(lr){1-9}
\textit{Macro} & $0.126$ & $0.121$ & $\underline{0.148}$ & $0.123$ & $0.115$ & $0.119$ & $0.124$ & $\mathbf{0.152}$ \\
\midrule
\multicolumn{9}{l}{\(R^2_{\mathrm{local}}\;\uparrow\)} \\
Brain & $-0.059\,\pm\,0.011$ & $-0.026\,\pm\,0.010$ & $\mathbf{0.013}\,\pm\,0.002$ & $-0.194\,\pm\,0.202$ & $-0.416\,\pm\,0.067$ & $\underline{-0.008}\,\pm\,0.019$ & $-0.241$ & $-0.010$ \\
Colon & $\underline{0.193}\,\pm\,0.002$ & $0.157\,\pm\,0.005$ & $0.098\,\pm\,0.037$ & $0.030\,\pm\,0.061$ & $-0.049\,\pm\,0.004$ & $0.114\,\pm\,0.046$ & $-0.058$ & $\mathbf{0.197}$ \\
Heart & $-0.016\,\pm\,0.019$ & $\mathbf{0.034}\,\pm\,0.009$ & $\underline{0.004}\,\pm\,0.002$ & $-0.120\,\pm\,0.021$ & $-0.120\,\pm\,0.003$ & $-0.035\,\pm\,0.007$ & $-0.145$ & $-0.001$ \\
Kidney & $0.042\,\pm\,0.005$ & $\underline{0.052}\,\pm\,0.004$ & $0.035\,\pm\,0.000$ & $-0.101\,\pm\,0.044$ & $-0.080\,\pm\,0.006$ & $0.016\,\pm\,0.017$ & $-0.069$ & $\mathbf{0.069}$ \\
Liver & $0.016\,\pm\,0.017$ & $0.056\,\pm\,0.004$ & $0.020\,\pm\,0.002$ & $\underline{0.068}\,\pm\,0.017$ & $-0.070\,\pm\,0.024$ & $0.033\,\pm\,0.016$ & $-0.024$ & $\mathbf{0.068}$ \\
Lung & $0.188\,\pm\,0.033$ & $0.125\,\pm\,0.030$ & $0.084\,\pm\,0.004$ & $\underline{0.206}\,\pm\,0.039$ & $0.126\,\pm\,0.061$ & $0.163\,\pm\,0.038$ & $0.170$ & $\mathbf{0.225}$ \\
Prostate & $-0.008\,\pm\,0.002$ & $0.000\,\pm\,0.011$ & $0.007\,\pm\,0.006$ & $-0.170\,\pm\,0.116$ & $-0.005\,\pm\,0.013$ & $\mathbf{0.020}\,\pm\,0.022$ & $-0.132$ & $\underline{0.013}$ \\
Skin & $\mathbf{0.437}\,\pm\,0.007$ & $\underline{0.434}\,\pm\,0.001$ & $0.065\,\pm\,0.014$ & $0.337\,\pm\,0.004$ & $0.261\,\pm\,0.051$ & $0.360\,\pm\,0.008$ & $0.375$ & $0.425$ \\
Uterus & $0.050\,\pm\,0.010$ & $\underline{0.094}\,\pm\,0.024$ & $-0.332\,\pm\,0.040$ & $-0.041\,\pm\,0.044$ & $-0.052\,\pm\,0.013$ & $0.080\,\pm\,0.008$ & $0.002$ & $\mathbf{0.146}$ \\
Breast & $0.028\,\pm\,0.015$ & $\mathbf{0.053}\,\pm\,0.001$ & $0.012\,\pm\,0.000$ & $-0.371\,\pm\,0.065$ & $-0.062\,\pm\,0.006$ & $0.043\,\pm\,0.006$ & $-0.048$ & $\underline{0.047}$ \\
\cmidrule(lr){1-9}
\textit{Macro} & $0.087$ & $\underline{0.098}$ & $0.001$ & $-0.036$ & $-0.047$ & $0.079$ & $-0.017$ & $\mathbf{0.118}$ \\
\midrule
\multicolumn{9}{l}{\(R^2_{\mathrm{flat}}\;\uparrow\)} \\
Brain & $\underline{0.649}\,\pm\,0.013$ & $0.648\,\pm\,0.005$ & $0.639\,\pm\,0.001$ & $0.628\,\pm\,0.056$ & $0.552\,\pm\,0.007$ & $0.635\,\pm\,0.009$ & $0.590$ & $\mathbf{0.659}$ \\
Colon & $\underline{0.540}\,\pm\,0.028$ & $0.511\,\pm\,0.001$ & $0.505\,\pm\,0.014$ & $0.460\,\pm\,0.069$ & $0.398\,\pm\,0.012$ & $0.491\,\pm\,0.007$ & $0.505$ & $\mathbf{0.573}$ \\
Heart & $0.495\,\pm\,0.018$ & $0.539\,\pm\,0.040$ & $0.246\,\pm\,0.002$ & $\underline{0.552}\,\pm\,0.003$ & $\mathbf{0.583}\,\pm\,0.015$ & $0.552\,\pm\,0.016$ & $0.458$ & $0.444$ \\
Kidney & $0.507\,\pm\,0.023$ & $0.480\,\pm\,0.003$ & $0.501\,\pm\,0.003$ & $0.432\,\pm\,0.036$ & $0.502\,\pm\,0.003$ & $\underline{0.513}\,\pm\,0.033$ & $0.504$ & $\mathbf{0.537}$ \\
Liver & $0.739\,\pm\,0.001$ & $\mathbf{0.759}\,\pm\,0.008$ & $0.723\,\pm\,0.003$ & $0.752\,\pm\,0.004$ & $0.737\,\pm\,0.002$ & $0.733\,\pm\,0.008$ & $0.747$ & $\underline{0.756}$ \\
Lung & $0.359\,\pm\,0.034$ & $0.324\,\pm\,0.025$ & $0.251\,\pm\,0.042$ & $\underline{0.380}\,\pm\,0.036$ & $0.187\,\pm\,0.029$ & $0.281\,\pm\,0.055$ & $\mathbf{0.398}$ & $0.349$ \\
Prostate & $0.235\,\pm\,0.021$ & $0.123\,\pm\,0.002$ & $\mathbf{0.341}\,\pm\,0.001$ & $0.163\,\pm\,0.048$ & $-0.155\,\pm\,0.003$ & $0.241\,\pm\,0.055$ & $0.159$ & $\underline{0.245}$ \\
Skin & $0.461\,\pm\,0.025$ & $\underline{0.502}\,\pm\,0.012$ & $0.243\,\pm\,0.009$ & $0.414\,\pm\,0.003$ & $0.336\,\pm\,0.045$ & $0.439\,\pm\,0.014$ & $0.452$ & $\mathbf{0.506}$ \\
Uterus & $0.672\,\pm\,0.014$ & $0.664\,\pm\,0.025$ & $0.550\,\pm\,0.017$ & $0.651\,\pm\,0.012$ & $0.647\,\pm\,0.032$ & $\underline{0.708}\,\pm\,0.012$ & $0.674$ & $\mathbf{0.715}$ \\
Breast & $0.400\,\pm\,0.013$ & $\underline{0.411}\,\pm\,0.010$ & $0.399\,\pm\,0.000$ & $-0.069\,\pm\,0.093$ & $0.346\,\pm\,0.001$ & $\mathbf{0.443}\,\pm\,0.002$ & $0.369$ & $0.407$ \\
\cmidrule(lr){1-9}
\textit{Macro} & $\underline{0.506}$ & $0.496$ & $0.440$ & $0.436$ & $0.413$ & $0.504$ & $0.485$ & $\mathbf{0.519}$ \\
\bottomrule
\end{tabular}%
}
\end{table*}

\clearpage
\begin{table*}[!ht]
\centering
\caption{Cohort-level test results with UNI and \(G=1500\). Neural entries are mean \(\pm\) seed standard deviation over three runs. Linear and CGL-Linear are deterministic. Bold and underlining mark the best and second-best unrounded means in each row. Macro rows average cohort means without weighting by cohort size. Higher is better.}
\label{tab:cohort-uni-1500}
\scriptsize
\setlength{\tabcolsep}{2.0pt}
\renewcommand{\arraystretch}{1.08}
\resizebox{\textwidth}{!}{%
\begin{tabular}{lrrrrrrrr}
\toprule
Cohort & HiST & HistoPrism & HEXST & STFlow & BLEEP & SEPAL & Linear & CGL-Linear \\
\midrule
\multicolumn{9}{l}{\(\mathrm{PCC}_{\mathrm{flat}}\;\uparrow\)} \\
Brain & $0.814\,\pm\,0.005$ & $\underline{0.817}\,\pm\,0.003$ & $0.813\,\pm\,0.001$ & $0.806\,\pm\,0.015$ & $0.753\,\pm\,0.014$ & $0.817\,\pm\,0.001$ & $0.792$ & $\mathbf{0.823}$ \\
Colon & $\mathbf{0.768}\,\pm\,0.011$ & $0.747\,\pm\,0.008$ & $0.716\,\pm\,0.003$ & $0.739\,\pm\,0.014$ & $0.688\,\pm\,0.001$ & $0.726\,\pm\,0.016$ & $0.727$ & $\underline{0.767}$ \\
Heart & $0.783\,\pm\,0.003$ & $\mathbf{0.789}\,\pm\,0.008$ & $0.699\,\pm\,0.000$ & $\underline{0.786}\,\pm\,0.002$ & $0.783\,\pm\,0.000$ & $0.770\,\pm\,0.005$ & $0.772$ & $0.764$ \\
Kidney & $\underline{0.758}\,\pm\,0.009$ & $0.745\,\pm\,0.002$ & $0.747\,\pm\,0.001$ & $0.713\,\pm\,0.016$ & $0.750\,\pm\,0.000$ & $0.755\,\pm\,0.016$ & $0.753$ & $\mathbf{0.766}$ \\
Liver & $0.873\,\pm\,0.008$ & $0.875\,\pm\,0.003$ & $0.866\,\pm\,0.000$ & $\underline{0.876}\,\pm\,0.003$ & $0.872\,\pm\,0.002$ & $0.869\,\pm\,0.008$ & $0.870$ & $\mathbf{0.878}$ \\
Lung & $0.590\,\pm\,0.004$ & $0.551\,\pm\,0.002$ & $0.619\,\pm\,0.007$ & $0.634\,\pm\,0.010$ & $0.521\,\pm\,0.052$ & $0.629\,\pm\,0.026$ & $\underline{0.637}$ & $\mathbf{0.645}$ \\
Prostate & $0.656\,\pm\,0.007$ & $0.657\,\pm\,0.012$ & $\mathbf{0.703}\,\pm\,0.003$ & $0.616\,\pm\,0.010$ & $0.434\,\pm\,0.021$ & $0.649\,\pm\,0.031$ & $0.542$ & $\underline{0.684}$ \\
Skin & $0.693\,\pm\,0.000$ & $\underline{0.715}\,\pm\,0.004$ & $0.511\,\pm\,0.004$ & $0.696\,\pm\,0.032$ & $0.651\,\pm\,0.026$ & $0.686\,\pm\,0.013$ & $0.698$ & $\mathbf{0.724}$ \\
Uterus & $0.815\,\pm\,0.009$ & $\underline{0.820}\,\pm\,0.000$ & $0.723\,\pm\,0.006$ & $0.819\,\pm\,0.007$ & $0.788\,\pm\,0.021$ & $0.787\,\pm\,0.081$ & $0.818$ & $\mathbf{0.854}$ \\
Breast & $0.609\,\pm\,0.002$ & $\underline{0.623}\,\pm\,0.008$ & $0.609\,\pm\,0.000$ & $0.541\,\pm\,0.003$ & $0.596\,\pm\,0.007$ & $\mathbf{0.649}\,\pm\,0.009$ & $0.599$ & $0.618$ \\
\cmidrule(lr){1-9}
\textit{Macro} & $\underline{0.736}$ & $0.734$ & $0.700$ & $0.722$ & $0.684$ & $0.734$ & $0.721$ & $\mathbf{0.752}$ \\
\midrule
\multicolumn{9}{l}{\(\mathrm{PCC}_{\mathrm{gene}}\;\uparrow\)} \\
Brain & $\underline{0.037}\,\pm\,0.004$ & $0.013\,\pm\,0.000$ & $0.034\,\pm\,0.001$ & $0.032\,\pm\,0.005$ & $0.023\,\pm\,0.006$ & $0.012\,\pm\,0.000$ & $0.033$ & $\mathbf{0.045}$ \\
Colon & $0.218\,\pm\,0.004$ & $0.182\,\pm\,0.014$ & $\mathbf{0.249}\,\pm\,0.006$ & $0.193\,\pm\,0.004$ & $0.158\,\pm\,0.006$ & $0.177\,\pm\,0.020$ & $0.191$ & $\underline{0.221}$ \\
Heart & $0.041\,\pm\,0.001$ & $0.040\,\pm\,0.009$ & $0.032\,\pm\,0.010$ & $\mathbf{0.051}\,\pm\,0.001$ & $0.048\,\pm\,0.001$ & $0.029\,\pm\,0.004$ & $0.047$ & $\underline{0.051}$ \\
Kidney & $\mathbf{0.084}\,\pm\,0.009$ & $0.064\,\pm\,0.003$ & $0.072\,\pm\,0.013$ & $0.057\,\pm\,0.009$ & $0.077\,\pm\,0.005$ & $0.049\,\pm\,0.019$ & $0.073$ & $\underline{0.083}$ \\
Liver & $0.096\,\pm\,0.009$ & $0.077\,\pm\,0.003$ & $0.068\,\pm\,0.009$ & $0.093\,\pm\,0.001$ & $\underline{0.115}\,\pm\,0.005$ & $0.076\,\pm\,0.000$ & $0.098$ & $\mathbf{0.116}$ \\
Lung & $0.211\,\pm\,0.005$ & $0.227\,\pm\,0.006$ & $\underline{0.307}\,\pm\,0.019$ & $0.305\,\pm\,0.003$ & $0.182\,\pm\,0.005$ & $0.256\,\pm\,0.029$ & $0.266$ & $\mathbf{0.318}$ \\
Prostate & $0.029\,\pm\,0.006$ & $0.028\,\pm\,0.002$ & $\mathbf{0.047}\,\pm\,0.050$ & $0.036\,\pm\,0.036$ & $0.011\,\pm\,0.000$ & $0.037\,\pm\,0.022$ & $-0.023$ & $\underline{0.041}$ \\
Skin & $0.338\,\pm\,0.000$ & $0.348\,\pm\,0.001$ & $\mathbf{0.404}\,\pm\,0.008$ & $0.366\,\pm\,0.040$ & $0.352\,\pm\,0.024$ & $0.308\,\pm\,0.019$ & $0.366$ & $\underline{0.400}$ \\
Uterus & $0.077\,\pm\,0.005$ & $0.074\,\pm\,0.007$ & $\mathbf{0.117}\,\pm\,0.006$ & $0.096\,\pm\,0.005$ & $0.080\,\pm\,0.012$ & $0.046\,\pm\,0.042$ & $0.083$ & $\underline{0.115}$ \\
Breast & $0.074\,\pm\,0.013$ & $\underline{0.096}\,\pm\,0.013$ & $0.094\,\pm\,0.001$ & $0.046\,\pm\,0.007$ & $0.081\,\pm\,0.006$ & $\mathbf{0.162}\,\pm\,0.022$ & $0.068$ & $0.083$ \\
\cmidrule(lr){1-9}
\textit{Macro} & $0.121$ & $0.115$ & $\underline{0.142}$ & $0.127$ & $0.113$ & $0.115$ & $0.120$ & $\mathbf{0.147}$ \\
\midrule
\multicolumn{9}{l}{\(R^2_{\mathrm{local}}\;\uparrow\)} \\
Brain & $-0.081\,\pm\,0.024$ & $-0.020\,\pm\,0.005$ & $\mathbf{0.010}\,\pm\,0.001$ & $-0.204\,\pm\,0.104$ & $-0.423\,\pm\,0.044$ & $\underline{0.006}\,\pm\,0.001$ & $-0.223$ & $-0.009$ \\
Colon & $\mathbf{0.219}\,\pm\,0.022$ & $0.158\,\pm\,0.020$ & $0.064\,\pm\,0.012$ & $0.106\,\pm\,0.044$ & $-0.077\,\pm\,0.052$ & $0.096\,\pm\,0.057$ & $-0.026$ & $\underline{0.211}$ \\
Heart & $\underline{0.009}\,\pm\,0.007$ & $\mathbf{0.036}\,\pm\,0.001$ & $0.003\,\pm\,0.001$ & $-0.086\,\pm\,0.017$ & $-0.082\,\pm\,0.017$ & $-0.014\,\pm\,0.002$ & $-0.115$ & $0.008$ \\
Kidney & $\underline{0.046}\,\pm\,0.014$ & $0.043\,\pm\,0.004$ & $0.029\,\pm\,0.012$ & $-0.118\,\pm\,0.085$ & $-0.063\,\pm\,0.018$ & $0.021\,\pm\,0.015$ & $-0.057$ & $\mathbf{0.069}$ \\
Liver & $0.065\,\pm\,0.057$ & $0.028\,\pm\,0.039$ & $0.026\,\pm\,0.005$ & $\mathbf{0.083}\,\pm\,0.010$ & $0.010\,\pm\,0.012$ & $0.037\,\pm\,0.021$ & $-0.008$ & $\underline{0.077}$ \\
Lung & $0.170\,\pm\,0.003$ & $0.115\,\pm\,0.001$ & $0.072\,\pm\,0.009$ & $\underline{0.188}\,\pm\,0.020$ & $0.046\,\pm\,0.039$ & $0.181\,\pm\,0.044$ & $0.180$ & $\mathbf{0.221}$ \\
Prostate & $-0.005\,\pm\,0.004$ & $-0.006\,\pm\,0.002$ & $\underline{0.003}\,\pm\,0.011$ & $-0.119\,\pm\,0.167$ & $-0.004\,\pm\,0.008$ & $0.000\,\pm\,0.008$ & $-0.131$ & $\mathbf{0.008}$ \\
Skin & $\underline{0.427}\,\pm\,0.003$ & $\mathbf{0.436}\,\pm\,0.013$ & $0.058\,\pm\,0.005$ & $0.377\,\pm\,0.057$ & $0.302\,\pm\,0.042$ & $0.335\,\pm\,0.029$ & $0.376$ & $0.423$ \\
Uterus & $0.058\,\pm\,0.003$ & $\underline{0.094}\,\pm\,0.004$ & $-0.504\,\pm\,0.134$ & $0.015\,\pm\,0.042$ & $-0.027\,\pm\,0.013$ & $0.045\,\pm\,0.055$ & $0.005$ & $\mathbf{0.142}$ \\
Breast & $0.022\,\pm\,0.018$ & $\mathbf{0.048}\,\pm\,0.008$ & $0.008\,\pm\,0.000$ & $-0.244\,\pm\,0.002$ & $-0.043\,\pm\,0.011$ & $0.038\,\pm\,0.005$ & $-0.045$ & $\underline{0.042}$ \\
\cmidrule(lr){1-9}
\textit{Macro} & $0.093$ & $\underline{0.093}$ & $-0.023$ & $0.000$ & $-0.036$ & $0.074$ & $-0.005$ & $\mathbf{0.119}$ \\
\midrule
\multicolumn{9}{l}{\(R^2_{\mathrm{flat}}\;\uparrow\)} \\
Brain & $0.661\,\pm\,0.008$ & $\underline{0.663}\,\pm\,0.004$ & $0.659\,\pm\,0.001$ & $0.636\,\pm\,0.042$ & $0.558\,\pm\,0.023$ & $0.660\,\pm\,0.001$ & $0.614$ & $\mathbf{0.677}$ \\
Colon & $\mathbf{0.586}\,\pm\,0.013$ & $0.483\,\pm\,0.068$ & $0.508\,\pm\,0.004$ & $0.531\,\pm\,0.008$ & $0.461\,\pm\,0.008$ & $0.513\,\pm\,0.010$ & $0.521$ & $\underline{0.585}$ \\
Heart & $0.474\,\pm\,0.002$ & $0.533\,\pm\,0.029$ & $0.297\,\pm\,0.001$ & $\underline{0.563}\,\pm\,0.006$ & $\mathbf{0.596}\,\pm\,0.004$ & $0.551\,\pm\,0.021$ & $0.460$ & $0.464$ \\
Kidney & $0.563\,\pm\,0.019$ & $0.545\,\pm\,0.002$ & $0.556\,\pm\,0.001$ & $0.484\,\pm\,0.041$ & $0.560\,\pm\,0.001$ & $\underline{0.570}\,\pm\,0.024$ & $0.564$ & $\mathbf{0.587}$ \\
Liver & $0.759\,\pm\,0.016$ & $\underline{0.761}\,\pm\,0.004$ & $0.739\,\pm\,0.001$ & $\mathbf{0.766}\,\pm\,0.005$ & $0.759\,\pm\,0.004$ & $0.749\,\pm\,0.022$ & $0.756$ & $0.760$ \\
Lung & $0.340\,\pm\,0.004$ & $0.293\,\pm\,0.007$ & $0.207\,\pm\,0.010$ & $\underline{0.389}\,\pm\,0.008$ & $0.164\,\pm\,0.009$ & $0.298\,\pm\,0.020$ & $\mathbf{0.394}$ & $0.350$ \\
Prostate & $0.201\,\pm\,0.012$ & $0.121\,\pm\,0.021$ & $\mathbf{0.324}\,\pm\,0.002$ & $0.220\,\pm\,0.070$ & $-0.145\,\pm\,0.031$ & $0.149\,\pm\,0.086$ & $0.146$ & $\underline{0.232}$ \\
Skin & $0.469\,\pm\,0.003$ & $\underline{0.506}\,\pm\,0.007$ & $0.259\,\pm\,0.004$ & $0.471\,\pm\,0.039$ & $0.409\,\pm\,0.029$ & $0.447\,\pm\,0.021$ & $0.473$ & $\mathbf{0.523}$ \\
Uterus & $0.660\,\pm\,0.018$ & $0.654\,\pm\,0.005$ & $0.483\,\pm\,0.001$ & $\underline{0.668}\,\pm\,0.014$ & $0.610\,\pm\,0.035$ & $0.576\,\pm\,0.157$ & $0.667$ & $\mathbf{0.708}$ \\
Breast & $0.360\,\pm\,0.000$ & $\underline{0.383}\,\pm\,0.010$ & $0.369\,\pm\,0.000$ & $0.000\,\pm\,0.035$ & $0.345\,\pm\,0.008$ & $\mathbf{0.419}\,\pm\,0.011$ & $0.343$ & $0.378$ \\
\cmidrule(lr){1-9}
\textit{Macro} & $\underline{0.507}$ & $0.494$ & $0.440$ & $0.473$ & $0.432$ & $0.493$ & $0.494$ & $\mathbf{0.526}$ \\
\bottomrule
\end{tabular}%
}
\end{table*}

\clearpage
\begin{table*}[!ht]
\centering
\caption{Cohort-level test results with UNI and \(G=2000\). Neural entries are mean \(\pm\) seed standard deviation over three runs. Linear and CGL-Linear are deterministic. Bold and underlining mark the best and second-best unrounded means in each row. Macro rows average cohort means without weighting by cohort size. Higher is better.}
\label{tab:cohort-uni-2000}
\scriptsize
\setlength{\tabcolsep}{2.0pt}
\renewcommand{\arraystretch}{1.08}
\resizebox{\textwidth}{!}{%
\begin{tabular}{lrrrrrrrr}
\toprule
Cohort & HiST & HistoPrism & HEXST & STFlow & BLEEP & SEPAL & Linear & CGL-Linear \\
\midrule
\multicolumn{9}{l}{\(\mathrm{PCC}_{\mathrm{flat}}\;\uparrow\)} \\
Brain & $0.809\,\pm\,0.003$ & $\underline{0.810}\,\pm\,0.001$ & $0.807\,\pm\,0.001$ & $0.802\,\pm\,0.011$ & $0.767\,\pm\,0.019$ & $0.809\,\pm\,0.003$ & $0.788$ & $\mathbf{0.818}$ \\
Colon & $\underline{0.749}\,\pm\,0.035$ & $0.740\,\pm\,0.010$ & $0.705\,\pm\,0.002$ & $0.718\,\pm\,0.025$ & $0.678\,\pm\,0.007$ & $0.715\,\pm\,0.025$ & $0.722$ & $\mathbf{0.763}$ \\
Heart & $0.773\,\pm\,0.000$ & $\mathbf{0.786}\,\pm\,0.008$ & $0.709\,\pm\,0.000$ & $\underline{0.776}\,\pm\,0.002$ & $0.772\,\pm\,0.010$ & $0.749\,\pm\,0.006$ & $0.771$ & $0.766$ \\
Kidney & $\underline{0.753}\,\pm\,0.002$ & $0.747\,\pm\,0.002$ & $0.742\,\pm\,0.001$ & $0.726\,\pm\,0.017$ & $0.743\,\pm\,0.007$ & $0.750\,\pm\,0.010$ & $0.748$ & $\mathbf{0.762}$ \\
Liver & $\underline{0.876}\,\pm\,0.002$ & $0.872\,\pm\,0.004$ & $0.857\,\pm\,0.006$ & $\mathbf{0.878}\,\pm\,0.007$ & $0.870\,\pm\,0.000$ & $0.843\,\pm\,0.012$ & $0.868$ & $0.875$ \\
Lung & $0.585\,\pm\,0.005$ & $0.531\,\pm\,0.014$ & $0.604\,\pm\,0.005$ & $0.618\,\pm\,0.019$ & $0.460\,\pm\,0.020$ & $\mathbf{0.636}\,\pm\,0.000$ & $0.621$ & $\underline{0.631}$ \\
Prostate & $0.639\,\pm\,0.010$ & $0.653\,\pm\,0.002$ & $\mathbf{0.685}\,\pm\,0.005$ & $0.593\,\pm\,0.012$ & $0.390\,\pm\,0.091$ & $0.658\,\pm\,0.002$ & $0.519$ & $\underline{0.662}$ \\
Skin & $0.675\,\pm\,0.008$ & $\mathbf{0.731}\,\pm\,0.029$ & $0.507\,\pm\,0.004$ & $0.688\,\pm\,0.027$ & $0.614\,\pm\,0.065$ & $0.688\,\pm\,0.008$ & $0.700$ & $\underline{0.726}$ \\
Uterus & $0.796\,\pm\,0.015$ & $0.818\,\pm\,0.014$ & $0.768\,\pm\,0.080$ & $\underline{0.821}\,\pm\,0.003$ & $0.797\,\pm\,0.025$ & $0.783\,\pm\,0.086$ & $0.812$ & $\mathbf{0.846}$ \\
Breast & $\underline{0.607}\,\pm\,0.010$ & $0.606\,\pm\,0.014$ & $0.589\,\pm\,0.000$ & $0.522\,\pm\,0.017$ & $0.567\,\pm\,0.009$ & $\mathbf{0.627}\,\pm\,0.006$ & $0.577$ & $0.597$ \\
\cmidrule(lr){1-9}
\textit{Macro} & $0.726$ & $\underline{0.729}$ & $0.697$ & $0.714$ & $0.666$ & $0.726$ & $0.713$ & $\mathbf{0.744}$ \\
\midrule
\multicolumn{9}{l}{\(\mathrm{PCC}_{\mathrm{gene}}\;\uparrow\)} \\
Brain & $\underline{0.038}\,\pm\,0.003$ & $0.013\,\pm\,0.001$ & $0.033\,\pm\,0.003$ & $0.037\,\pm\,0.002$ & $0.038\,\pm\,0.006$ & $0.014\,\pm\,0.001$ & $0.032$ & $\mathbf{0.046}$ \\
Colon & $0.190\,\pm\,0.040$ & $0.183\,\pm\,0.013$ & $\mathbf{0.245}\,\pm\,0.003$ & $0.178\,\pm\,0.005$ & $0.142\,\pm\,0.012$ & $0.167\,\pm\,0.020$ & $0.188$ & $\underline{0.219}$ \\
Heart & $0.034\,\pm\,0.001$ & $0.038\,\pm\,0.010$ & $0.033\,\pm\,0.014$ & $\underline{0.044}\,\pm\,0.004$ & $0.037\,\pm\,0.009$ & $0.022\,\pm\,0.011$ & $0.043$ & $\mathbf{0.048}$ \\
Kidney & $\underline{0.072}\,\pm\,0.004$ & $0.065\,\pm\,0.007$ & $0.067\,\pm\,0.009$ & $0.064\,\pm\,0.011$ & $0.069\,\pm\,0.003$ & $0.044\,\pm\,0.015$ & $0.067$ & $\mathbf{0.077}$ \\
Liver & $0.110\,\pm\,0.000$ & $0.085\,\pm\,0.011$ & $0.073\,\pm\,0.011$ & $0.098\,\pm\,0.022$ & $\underline{0.122}\,\pm\,0.004$ & $0.061\,\pm\,0.022$ & $0.105$ & $\mathbf{0.123}$ \\
Lung & $0.203\,\pm\,0.001$ & $0.226\,\pm\,0.001$ & $\underline{0.304}\,\pm\,0.008$ & $0.298\,\pm\,0.020$ & $0.190\,\pm\,0.003$ & $0.248\,\pm\,0.006$ & $0.262$ & $\mathbf{0.313}$ \\
Prostate & $0.025\,\pm\,0.011$ & $0.032\,\pm\,0.008$ & $\mathbf{0.047}\,\pm\,0.029$ & $\underline{0.040}\,\pm\,0.005$ & $0.012\,\pm\,0.026$ & $0.021\,\pm\,0.023$ & $-0.028$ & $0.036$ \\
Skin & $0.305\,\pm\,0.006$ & $0.357\,\pm\,0.032$ & $\underline{0.382}\,\pm\,0.003$ & $0.341\,\pm\,0.039$ & $0.314\,\pm\,0.038$ & $0.296\,\pm\,0.015$ & $0.352$ & $\mathbf{0.386}$ \\
Uterus & $0.069\,\pm\,0.002$ & $0.078\,\pm\,0.008$ & $\underline{0.104}\,\pm\,0.003$ & $0.090\,\pm\,0.005$ & $0.071\,\pm\,0.021$ & $0.040\,\pm\,0.049$ & $0.078$ & $\mathbf{0.107}$ \\
Breast & $\underline{0.118}\,\pm\,0.013$ & $0.101\,\pm\,0.033$ & $0.103\,\pm\,0.010$ & $0.056\,\pm\,0.013$ & $0.056\,\pm\,0.005$ & $\mathbf{0.151}\,\pm\,0.020$ & $0.068$ & $0.083$ \\
\cmidrule(lr){1-9}
\textit{Macro} & $0.116$ & $0.118$ & $\underline{0.139}$ & $0.125$ & $0.105$ & $0.106$ & $0.117$ & $\mathbf{0.144}$ \\
\midrule
\multicolumn{9}{l}{\(R^2_{\mathrm{local}}\;\uparrow\)} \\
Brain & $-0.093\,\pm\,0.030$ & $-0.011\,\pm\,0.001$ & $\mathbf{0.011}\,\pm\,0.000$ & $-0.187\,\pm\,0.122$ & $-0.358\,\pm\,0.048$ & $\underline{0.004}\,\pm\,0.000$ & $-0.202$ & $-0.005$ \\
Colon & $\underline{0.199}\,\pm\,0.040$ & $0.164\,\pm\,0.009$ & $0.055\,\pm\,0.007$ & $0.036\,\pm\,0.105$ & $0.018\,\pm\,0.036$ & $0.108\,\pm\,0.070$ & $-0.009$ & $\mathbf{0.219}$ \\
Heart & $-0.010\,\pm\,0.034$ & $\mathbf{0.032}\,\pm\,0.002$ & $0.001\,\pm\,0.000$ & $-0.103\,\pm\,0.024$ & $-0.084\,\pm\,0.032$ & $0.000\,\pm\,0.007$ & $-0.101$ & $\underline{0.012}$ \\
Kidney & $0.044\,\pm\,0.009$ & $\underline{0.047}\,\pm\,0.006$ & $0.024\,\pm\,0.011$ & $-0.114\,\pm\,0.001$ & $-0.037\,\pm\,0.005$ & $0.018\,\pm\,0.016$ & $-0.053$ & $\mathbf{0.068}$ \\
Liver & $\mathbf{0.098}\,\pm\,0.005$ & $0.024\,\pm\,0.048$ & $0.017\,\pm\,0.002$ & $\underline{0.090}\,\pm\,0.009$ & $0.004\,\pm\,0.010$ & $0.030\,\pm\,0.016$ & $-0.002$ & $0.077$ \\
Lung & $0.169\,\pm\,0.002$ & $0.107\,\pm\,0.001$ & $0.066\,\pm\,0.007$ & $\underline{0.194}\,\pm\,0.029$ & $0.037\,\pm\,0.034$ & $0.167\,\pm\,0.005$ & $0.179$ & $\mathbf{0.211}$ \\
Prostate & $0.001\,\pm\,0.014$ & $0.002\,\pm\,0.006$ & $\underline{0.005}\,\pm\,0.004$ & $-0.029\,\pm\,0.018$ & $-0.012\,\pm\,0.030$ & $-0.004\,\pm\,0.015$ & $-0.129$ & $\mathbf{0.006}$ \\
Skin & $0.409\,\pm\,0.012$ & $\mathbf{0.448}\,\pm\,0.028$ & $0.049\,\pm\,0.005$ & $0.364\,\pm\,0.039$ & $0.267\,\pm\,0.097$ & $0.335\,\pm\,0.018$ & $0.376$ & $\underline{0.422}$ \\
Uterus & $0.040\,\pm\,0.023$ & $\underline{0.093}\,\pm\,0.017$ & $-0.272\,\pm\,0.445$ & $-0.006\,\pm\,0.051$ & $-0.054\,\pm\,0.064$ & $0.046\,\pm\,0.064$ & $0.003$ & $\mathbf{0.134}$ \\
Breast & $0.035\,\pm\,0.002$ & $\mathbf{0.045}\,\pm\,0.006$ & $0.006\,\pm\,0.000$ & $-0.197\,\pm\,0.022$ & $-0.056\,\pm\,0.001$ & $0.036\,\pm\,0.011$ & $-0.047$ & $\underline{0.039}$ \\
\cmidrule(lr){1-9}
\textit{Macro} & $0.089$ & $\underline{0.095}$ & $-0.004$ & $0.005$ & $-0.027$ & $0.074$ & $0.001$ & $\mathbf{0.118}$ \\
\midrule
\multicolumn{9}{l}{\(R^2_{\mathrm{flat}}\;\uparrow\)} \\
Brain & $0.652\,\pm\,0.004$ & $\underline{0.653}\,\pm\,0.002$ & $0.651\,\pm\,0.001$ & $0.633\,\pm\,0.030$ & $0.582\,\pm\,0.034$ & $0.647\,\pm\,0.004$ & $0.608$ & $\mathbf{0.668}$ \\
Colon & $\underline{0.538}\,\pm\,0.084$ & $0.517\,\pm\,0.002$ & $0.492\,\pm\,0.002$ & $0.505\,\pm\,0.034$ & $0.419\,\pm\,0.015$ & $0.488\,\pm\,0.006$ & $0.513$ & $\mathbf{0.578}$ \\
Heart & $0.391\,\pm\,0.049$ & $0.512\,\pm\,0.025$ & $0.325\,\pm\,0.001$ & $\mathbf{0.549}\,\pm\,0.022$ & $\underline{0.543}\,\pm\,0.044$ & $0.511\,\pm\,0.053$ & $0.451$ & $0.464$ \\
Kidney & $0.558\,\pm\,0.010$ & $0.553\,\pm\,0.004$ & $0.549\,\pm\,0.001$ & $0.502\,\pm\,0.036$ & $0.549\,\pm\,0.013$ & $\underline{0.561}\,\pm\,0.018$ & $0.557$ & $\mathbf{0.580}$ \\
Liver & $\underline{0.765}\,\pm\,0.007$ & $0.757\,\pm\,0.011$ & $0.725\,\pm\,0.005$ & $\mathbf{0.769}\,\pm\,0.013$ & $0.752\,\pm\,0.004$ & $0.668\,\pm\,0.056$ & $0.753$ & $0.754$ \\
Lung & $0.326\,\pm\,0.011$ & $0.262\,\pm\,0.015$ & $0.189\,\pm\,0.008$ & $\underline{0.369}\,\pm\,0.028$ & $0.144\,\pm\,0.009$ & $0.301\,\pm\,0.012$ & $\mathbf{0.374}$ & $0.334$ \\
Prostate & $0.172\,\pm\,0.015$ & $0.136\,\pm\,0.019$ & $\mathbf{0.296}\,\pm\,0.013$ & $0.141\,\pm\,0.064$ & $-0.154\,\pm\,0.020$ & $0.191\,\pm\,0.032$ & $0.124$ & $\underline{0.211}$ \\
Skin & $0.439\,\pm\,0.006$ & $\mathbf{0.528}\,\pm\,0.038$ & $0.256\,\pm\,0.003$ & $0.460\,\pm\,0.023$ & $0.363\,\pm\,0.088$ & $0.446\,\pm\,0.007$ & $0.475$ & $\underline{0.526}$ \\
Uterus & $0.625\,\pm\,0.035$ & $0.655\,\pm\,0.040$ & $0.558\,\pm\,0.154$ & $\underline{0.673}\,\pm\,0.005$ & $0.631\,\pm\,0.047$ & $0.512\,\pm\,0.243$ & $0.657$ & $\mathbf{0.698}$ \\
Breast & $0.363\,\pm\,0.014$ & $\underline{0.363}\,\pm\,0.019$ & $0.345\,\pm\,0.000$ & $0.002\,\pm\,0.037$ & $0.306\,\pm\,0.006$ & $\mathbf{0.390}\,\pm\,0.008$ & $0.316$ & $0.353$ \\
\cmidrule(lr){1-9}
\textit{Macro} & $0.483$ & $\underline{0.494}$ & $0.439$ & $0.460$ & $0.413$ & $0.472$ & $0.483$ & $\mathbf{0.517}$ \\
\bottomrule
\end{tabular}%
}
\end{table*}

\clearpage
\begin{table*}[!ht]
\centering
\caption{Cohort-level test results with DINOv2-Large and \(G=500\). Neural entries are mean \(\pm\) seed standard deviation over three runs. Linear and CGL-Linear are deterministic. Bold and underlining mark the best and second-best unrounded means in each row. Macro rows average cohort means without weighting by cohort size. Higher is better.}
\label{tab:cohort-dinov2-large-500}
\scriptsize
\setlength{\tabcolsep}{2.0pt}
\renewcommand{\arraystretch}{1.08}
\resizebox{\textwidth}{!}{%
\begin{tabular}{lrrrrrrrr}
\toprule
Cohort & HiST & HistoPrism & HEXST & STFlow & BLEEP & SEPAL & Linear & CGL-Linear \\
\midrule
\multicolumn{9}{l}{\(\mathrm{PCC}_{\mathrm{flat}}\;\uparrow\)} \\
Brain & $0.752\,\pm\,0.018$ & $\mathbf{0.777}\,\pm\,0.002$ & $0.765\,\pm\,0.001$ & $0.770\,\pm\,0.000$ & $0.706\,\pm\,0.010$ & $0.771\,\pm\,0.006$ & $0.738$ & $\underline{0.773}$ \\
Colon & $0.677\,\pm\,0.009$ & $0.699\,\pm\,0.023$ & $\underline{0.710}\,\pm\,0.002$ & $0.687\,\pm\,0.000$ & $0.625\,\pm\,0.021$ & $0.694\,\pm\,0.002$ & $0.664$ & $\mathbf{0.731}$ \\
Heart & $\mathbf{0.765}\,\pm\,0.005$ & $0.709\,\pm\,0.009$ & $0.646\,\pm\,0.000$ & $0.644\,\pm\,0.013$ & $\underline{0.748}\,\pm\,0.009$ & $0.738\,\pm\,0.039$ & $0.711$ & $0.684$ \\
Kidney & $0.719\,\pm\,0.020$ & $0.684\,\pm\,0.017$ & $0.717\,\pm\,0.000$ & $0.707\,\pm\,0.010$ & $0.686\,\pm\,0.010$ & $\mathbf{0.741}\,\pm\,0.000$ & $0.720$ & $\underline{0.730}$ \\
Liver & $0.863\,\pm\,0.005$ & $\mathbf{0.871}\,\pm\,0.000$ & $0.862\,\pm\,0.000$ & $0.859\,\pm\,0.005$ & $0.861\,\pm\,0.001$ & $0.857\,\pm\,0.005$ & $0.861$ & $\underline{0.869}$ \\
Lung & $\underline{0.651}\,\pm\,0.006$ & $0.603\,\pm\,0.020$ & $0.515\,\pm\,0.023$ & $\mathbf{0.669}\,\pm\,0.016$ & $0.522\,\pm\,0.021$ & $0.631\,\pm\,0.005$ & $0.591$ & $0.639$ \\
Prostate & $0.637\,\pm\,0.049$ & $0.653\,\pm\,0.004$ & $\mathbf{0.718}\,\pm\,0.004$ & $0.605\,\pm\,0.028$ & $0.615\,\pm\,0.001$ & $0.687\,\pm\,0.059$ & $0.622$ & $\underline{0.710}$ \\
Skin & $\mathbf{0.647}\,\pm\,0.011$ & $0.640\,\pm\,0.006$ & $0.498\,\pm\,0.009$ & $0.541\,\pm\,0.077$ & $\underline{0.643}\,\pm\,0.021$ & $0.598\,\pm\,0.001$ & $0.630$ & $0.643$ \\
Uterus & $0.822\,\pm\,0.005$ & $0.825\,\pm\,0.001$ & $0.803\,\pm\,0.054$ & $0.846\,\pm\,0.004$ & $0.785\,\pm\,0.019$ & $\underline{0.856}\,\pm\,0.004$ & $0.827$ & $\mathbf{0.865}$ \\
Breast & $0.678\,\pm\,0.017$ & $\mathbf{0.699}\,\pm\,0.005$ & $0.671\,\pm\,0.001$ & $0.646\,\pm\,0.008$ & $0.630\,\pm\,0.017$ & $\underline{0.697}\,\pm\,0.002$ & $0.661$ & $0.676$ \\
\cmidrule(lr){1-9}
\textit{Macro} & $0.721$ & $0.716$ & $0.691$ & $0.697$ & $0.682$ & $\underline{0.727}$ & $0.703$ & $\mathbf{0.732}$ \\
\midrule
\multicolumn{9}{l}{\(\mathrm{PCC}_{\mathrm{gene}}\;\uparrow\)} \\
Brain & $0.029\,\pm\,0.010$ & $0.014\,\pm\,0.003$ & $0.031\,\pm\,0.002$ & $0.027\,\pm\,0.003$ & $0.021\,\pm\,0.001$ & $0.006\,\pm\,0.000$ & $\mathbf{0.037}$ & $\underline{0.036}$ \\
Colon & $0.134\,\pm\,0.008$ & $0.118\,\pm\,0.046$ & $\mathbf{0.219}\,\pm\,0.001$ & $0.018\,\pm\,0.002$ & $0.135\,\pm\,0.037$ & $0.109\,\pm\,0.027$ & $0.130$ & $\underline{0.180}$ \\
Heart & $0.031\,\pm\,0.001$ & $0.015\,\pm\,0.003$ & $0.028\,\pm\,0.004$ & $0.034\,\pm\,0.002$ & $0.028\,\pm\,0.003$ & $\underline{0.035}\,\pm\,0.001$ & $0.028$ & $\mathbf{0.040}$ \\
Kidney & $\mathbf{0.091}\,\pm\,0.010$ & $0.047\,\pm\,0.002$ & $0.075\,\pm\,0.004$ & $0.054\,\pm\,0.008$ & $0.068\,\pm\,0.007$ & $0.071\,\pm\,0.001$ & $0.077$ & $\underline{0.078}$ \\
Liver & $\underline{0.088}\,\pm\,0.013$ & $0.078\,\pm\,0.001$ & $0.049\,\pm\,0.019$ & $0.087\,\pm\,0.013$ & $\mathbf{0.091}\,\pm\,0.000$ & $0.070\,\pm\,0.014$ & $0.080$ & $0.087$ \\
Lung & $0.239\,\pm\,0.011$ & $0.198\,\pm\,0.004$ & $\mathbf{0.344}\,\pm\,0.006$ & $0.230\,\pm\,0.009$ & $0.223\,\pm\,0.013$ & $0.204\,\pm\,0.003$ & $0.246$ & $\underline{0.285}$ \\
Prostate & $0.004\,\pm\,0.036$ & $0.019\,\pm\,0.026$ & $0.030\,\pm\,0.016$ & $-0.019\,\pm\,0.005$ & $\mathbf{0.053}\,\pm\,0.003$ & $0.018\,\pm\,0.033$ & $0.010$ & $\underline{0.049}$ \\
Skin & $0.354\,\pm\,0.000$ & $0.314\,\pm\,0.006$ & $\mathbf{0.419}\,\pm\,0.004$ & $0.233\,\pm\,0.089$ & $\underline{0.378}\,\pm\,0.013$ & $0.291\,\pm\,0.000$ & $0.347$ & $0.365$ \\
Uterus & $0.106\,\pm\,0.012$ & $0.090\,\pm\,0.005$ & $\underline{0.111}\,\pm\,0.003$ & $0.071\,\pm\,0.024$ & $0.079\,\pm\,0.002$ & $0.047\,\pm\,0.025$ & $0.089$ & $\mathbf{0.126}$ \\
Breast & $0.103\,\pm\,0.029$ & $\underline{0.134}\,\pm\,0.011$ & $0.049\,\pm\,0.011$ & $0.081\,\pm\,0.017$ & $0.055\,\pm\,0.013$ & $\mathbf{0.136}\,\pm\,0.009$ & $0.068$ & $0.074$ \\
\cmidrule(lr){1-9}
\textit{Macro} & $0.118$ & $0.103$ & $\mathbf{0.136}$ & $0.082$ & $0.113$ & $0.099$ & $0.111$ & $\underline{0.132}$ \\
\midrule
\multicolumn{9}{l}{\(R^2_{\mathrm{local}}\;\uparrow\)} \\
Brain & $-0.091\,\pm\,0.034$ & $-0.015\,\pm\,0.005$ & $\mathbf{0.008}\,\pm\,0.001$ & $-0.062\,\pm\,0.026$ & $-0.448\,\pm\,0.003$ & $\underline{-0.001}\,\pm\,0.007$ & $-0.303$ & $-0.042$ \\
Colon & $\underline{0.070}\,\pm\,0.005$ & $0.059\,\pm\,0.006$ & $0.060\,\pm\,0.005$ & $0.000\,\pm\,0.000$ & $-0.174\,\pm\,0.043$ & $0.036\,\pm\,0.018$ & $-0.142$ & $\mathbf{0.102}$ \\
Heart & $-0.090\,\pm\,0.006$ & $\underline{-0.046}\,\pm\,0.006$ & $\mathbf{0.003}\,\pm\,0.000$ & $-0.054\,\pm\,0.006$ & $-0.177\,\pm\,0.014$ & $-0.117\,\pm\,0.061$ & $-0.475$ & $-0.071$ \\
Kidney & $\underline{0.039}\,\pm\,0.003$ & $0.021\,\pm\,0.001$ & $0.030\,\pm\,0.007$ & $-0.010\,\pm\,0.001$ & $-0.254\,\pm\,0.028$ & $-0.018\,\pm\,0.001$ & $-0.127$ & $\mathbf{0.042}$ \\
Liver & $0.008\,\pm\,0.034$ & $-0.008\,\pm\,0.002$ & $\underline{0.013}\,\pm\,0.004$ & $0.001\,\pm\,0.005$ & $-0.078\,\pm\,0.060$ & $0.004\,\pm\,0.003$ & $-0.103$ & $\mathbf{0.018}$ \\
Lung & $\mathbf{0.218}\,\pm\,0.010$ & $0.087\,\pm\,0.002$ & $0.108\,\pm\,0.089$ & $0.141\,\pm\,0.001$ & $0.090\,\pm\,0.018$ & $0.069\,\pm\,0.014$ & $\underline{0.148}$ & $0.133$ \\
Prostate & $-0.132\,\pm\,0.070$ & $\underline{-0.015}\,\pm\,0.010$ & $\mathbf{-0.014}\,\pm\,0.001$ & $-0.258\,\pm\,0.098$ & $-0.232\,\pm\,0.136$ & $-0.063\,\pm\,0.080$ & $-0.177$ & $-0.052$ \\
Skin & $\mathbf{0.396}\,\pm\,0.002$ & $\underline{0.332}\,\pm\,0.003$ & $0.079\,\pm\,0.012$ & $0.191\,\pm\,0.090$ & $0.296\,\pm\,0.031$ & $0.254\,\pm\,0.004$ & $0.292$ & $0.323$ \\
Uterus & $0.069\,\pm\,0.000$ & $\underline{0.080}\,\pm\,0.002$ & $-0.143\,\pm\,0.282$ & $0.045\,\pm\,0.020$ & $-0.120\,\pm\,0.021$ & $0.044\,\pm\,0.036$ & $-0.087$ & $\mathbf{0.108}$ \\
Breast & $\mathbf{0.062}\,\pm\,0.016$ & $\underline{0.062}\,\pm\,0.007$ & $0.012\,\pm\,0.005$ & $-0.222\,\pm\,0.022$ & $-0.100\,\pm\,0.022$ & $0.009\,\pm\,0.002$ & $-0.049$ & $0.046$ \\
\cmidrule(lr){1-9}
\textit{Macro} & $0.055$ & $\underline{0.056}$ & $0.016$ & $-0.023$ & $-0.120$ & $0.022$ & $-0.102$ & $\mathbf{0.061}$ \\
\midrule
\multicolumn{9}{l}{\(R^2_{\mathrm{flat}}\;\uparrow\)} \\
Brain & $0.556\,\pm\,0.032$ & $\mathbf{0.599}\,\pm\,0.002$ & $0.584\,\pm\,0.001$ & $0.590\,\pm\,0.001$ & $0.481\,\pm\,0.005$ & $0.593\,\pm\,0.007$ & $0.525$ & $\underline{0.595}$ \\
Colon & $0.430\,\pm\,0.015$ & $0.470\,\pm\,0.055$ & $\underline{0.497}\,\pm\,0.002$ & $0.464\,\pm\,0.002$ & $0.345\,\pm\,0.060$ & $0.462\,\pm\,0.028$ & $0.440$ & $\mathbf{0.525}$ \\
Heart & $0.439\,\pm\,0.008$ & $0.342\,\pm\,0.047$ & $0.204\,\pm\,0.001$ & $0.300\,\pm\,0.022$ & $\mathbf{0.553}\,\pm\,0.017$ & $\underline{0.500}\,\pm\,0.091$ & $0.359$ & $0.303$ \\
Kidney & $0.479\,\pm\,0.049$ & $0.453\,\pm\,0.030$ & $0.514\,\pm\,0.000$ & $0.494\,\pm\,0.014$ & $0.457\,\pm\,0.004$ & $\mathbf{0.546}\,\pm\,0.001$ & $0.514$ & $\underline{0.533}$ \\
Liver & $0.741\,\pm\,0.008$ & $\mathbf{0.755}\,\pm\,0.002$ & $0.734\,\pm\,0.000$ & $0.731\,\pm\,0.009$ & $0.737\,\pm\,0.001$ & $0.733\,\pm\,0.008$ & $0.739$ & $\underline{0.751}$ \\
Lung & $\underline{0.361}\,\pm\,0.021$ & $0.295\,\pm\,0.030$ & $0.246\,\pm\,0.028$ & $\mathbf{0.437}\,\pm\,0.020$ & $0.155\,\pm\,0.018$ & $0.189\,\pm\,0.053$ & $0.279$ & $0.253$ \\
Prostate & $0.227\,\pm\,0.099$ & $0.077\,\pm\,0.036$ & $\mathbf{0.358}\,\pm\,0.002$ & $0.194\,\pm\,0.002$ & $0.247\,\pm\,0.037$ & $0.253\,\pm\,0.089$ & $0.219$ & $\underline{0.297}$ \\
Skin & $0.399\,\pm\,0.020$ & $\underline{0.401}\,\pm\,0.007$ & $0.244\,\pm\,0.008$ & $0.282\,\pm\,0.090$ & $0.381\,\pm\,0.024$ & $0.342\,\pm\,0.005$ & $0.391$ & $\mathbf{0.413}$ \\
Uterus & $0.661\,\pm\,0.011$ & $0.658\,\pm\,0.005$ & $0.634\,\pm\,0.078$ & $\underline{0.713}\,\pm\,0.005$ & $0.583\,\pm\,0.024$ & $0.704\,\pm\,0.010$ & $0.676$ & $\mathbf{0.724}$ \\
Breast & $0.454\,\pm\,0.025$ & $\underline{0.481}\,\pm\,0.004$ & $0.449\,\pm\,0.002$ & $0.172\,\pm\,0.008$ & $0.384\,\pm\,0.024$ & $\mathbf{0.481}\,\pm\,0.005$ & $0.429$ & $0.455$ \\
\cmidrule(lr){1-9}
\textit{Macro} & $0.475$ & $0.453$ & $0.446$ & $0.438$ & $0.432$ & $\underline{0.480}$ & $0.457$ & $\mathbf{0.485}$ \\
\bottomrule
\end{tabular}%
}
\end{table*}

\clearpage
\begin{table*}[!ht]
\centering
\caption{Cohort-level test results with DINOv2-Large and \(G=1000\). Neural entries are mean \(\pm\) seed standard deviation over three runs. Linear and CGL-Linear are deterministic. Bold and underlining mark the best and second-best unrounded means in each row. Macro rows average cohort means without weighting by cohort size. Higher is better.}
\label{tab:cohort-dinov2-large-1000}
\scriptsize
\setlength{\tabcolsep}{2.0pt}
\renewcommand{\arraystretch}{1.08}
\resizebox{\textwidth}{!}{%
\begin{tabular}{lrrrrrrrr}
\toprule
Cohort & HiST & HistoPrism & HEXST & STFlow & BLEEP & SEPAL & Linear & CGL-Linear \\
\midrule
\multicolumn{9}{l}{\(\mathrm{PCC}_{\mathrm{flat}}\;\uparrow\)} \\
Brain & $0.799\,\pm\,0.002$ & $\underline{0.810}\,\pm\,0.001$ & $0.801\,\pm\,0.001$ & $\mathbf{0.812}\,\pm\,0.000$ & $0.747\,\pm\,0.006$ & $0.805\,\pm\,0.003$ & $0.780$ & $0.808$ \\
Colon & $0.731\,\pm\,0.017$ & $\mathbf{0.737}\,\pm\,0.012$ & $0.703\,\pm\,0.005$ & $0.683\,\pm\,0.004$ & $0.540\,\pm\,0.066$ & $0.697\,\pm\,0.009$ & $0.675$ & $\underline{0.734}$ \\
Heart & $\mathbf{0.748}\,\pm\,0.001$ & $\underline{0.733}\,\pm\,0.002$ & $0.674\,\pm\,0.000$ & $0.686\,\pm\,0.008$ & $0.727\,\pm\,0.012$ & $0.722\,\pm\,0.000$ & $0.722$ & $0.710$ \\
Kidney & $0.705\,\pm\,0.013$ & $0.673\,\pm\,0.005$ & $0.707\,\pm\,0.001$ & $0.679\,\pm\,0.008$ & $0.659\,\pm\,0.006$ & $\underline{0.710}\,\pm\,0.021$ & $0.705$ & $\mathbf{0.719}$ \\
Liver & $0.863\,\pm\,0.001$ & $\underline{0.866}\,\pm\,0.001$ & $0.858\,\pm\,0.001$ & $0.865\,\pm\,0.009$ & $0.856\,\pm\,0.010$ & $0.863\,\pm\,0.002$ & $0.862$ & $\mathbf{0.867}$ \\
Lung & $\underline{0.619}\,\pm\,0.012$ & $0.543\,\pm\,0.006$ & $0.542\,\pm\,0.095$ & $0.516\,\pm\,0.156$ & $0.460\,\pm\,0.065$ & $0.619\,\pm\,0.023$ & $0.582$ & $\mathbf{0.621}$ \\
Prostate & $0.538\,\pm\,0.086$ & $0.628\,\pm\,0.011$ & $\underline{0.707}\,\pm\,0.004$ & $0.577\,\pm\,0.025$ & $0.610\,\pm\,0.027$ & $\mathbf{0.723}\,\pm\,0.010$ & $0.594$ & $0.688$ \\
Skin & $\underline{0.650}\,\pm\,0.013$ & $0.647\,\pm\,0.004$ & $0.489\,\pm\,0.006$ & $0.576\,\pm\,0.095$ & $0.640\,\pm\,0.043$ & $0.592\,\pm\,0.013$ & $0.642$ & $\mathbf{0.654}$ \\
Uterus & $0.821\,\pm\,0.002$ & $0.810\,\pm\,0.002$ & $0.831\,\pm\,0.002$ & $\underline{0.832}\,\pm\,0.004$ & $0.804\,\pm\,0.009$ & $0.819\,\pm\,0.043$ & $0.814$ & $\mathbf{0.852}$ \\
Breast & $\underline{0.651}\,\pm\,0.008$ & $0.649\,\pm\,0.008$ & $0.631\,\pm\,0.001$ & $0.616\,\pm\,0.007$ & $0.603\,\pm\,0.029$ & $\mathbf{0.658}\,\pm\,0.009$ & $0.621$ & $0.637$ \\
\cmidrule(lr){1-9}
\textit{Macro} & $0.713$ & $0.710$ & $0.694$ & $0.684$ & $0.665$ & $\underline{0.721}$ & $0.700$ & $\mathbf{0.729}$ \\
\midrule
\multicolumn{9}{l}{\(\mathrm{PCC}_{\mathrm{gene}}\;\uparrow\)} \\
Brain & $0.022\,\pm\,0.000$ & $0.010\,\pm\,0.000$ & $0.023\,\pm\,0.004$ & $0.027\,\pm\,0.006$ & $0.018\,\pm\,0.003$ & $0.003\,\pm\,0.002$ & $\underline{0.033}$ & $\mathbf{0.034}$ \\
Colon & $0.178\,\pm\,0.005$ & $0.176\,\pm\,0.011$ & $\mathbf{0.218}\,\pm\,0.004$ & $0.027\,\pm\,0.007$ & $0.091\,\pm\,0.052$ & $0.108\,\pm\,0.009$ & $0.131$ & $\underline{0.183}$ \\
Heart & $0.023\,\pm\,0.000$ & $0.015\,\pm\,0.003$ & $0.029\,\pm\,0.002$ & $\underline{0.033}\,\pm\,0.002$ & $0.025\,\pm\,0.001$ & $0.028\,\pm\,0.003$ & $0.024$ & $\mathbf{0.038}$ \\
Kidney & $0.066\,\pm\,0.001$ & $0.042\,\pm\,0.001$ & $\mathbf{0.068}\,\pm\,0.000$ & $0.038\,\pm\,0.010$ & $0.051\,\pm\,0.001$ & $0.041\,\pm\,0.028$ & $0.063$ & $\underline{0.068}$ \\
Liver & $\underline{0.085}\,\pm\,0.005$ & $0.065\,\pm\,0.002$ & $0.059\,\pm\,0.003$ & $0.076\,\pm\,0.010$ & $0.084\,\pm\,0.013$ & $0.066\,\pm\,0.007$ & $0.083$ & $\mathbf{0.088}$ \\
Lung & $0.217\,\pm\,0.007$ & $0.174\,\pm\,0.004$ & $\mathbf{0.302}\,\pm\,0.003$ & $0.175\,\pm\,0.045$ & $0.127\,\pm\,0.017$ & $0.194\,\pm\,0.062$ & $0.227$ & $\underline{0.264}$ \\
Prostate & $-0.018\,\pm\,0.019$ & $-0.002\,\pm\,0.012$ & $0.023\,\pm\,0.042$ & $0.001\,\pm\,0.014$ & $\underline{0.047}\,\pm\,0.007$ & $\mathbf{0.048}\,\pm\,0.004$ & $-0.011$ & $0.034$ \\
Skin & $0.303\,\pm\,0.006$ & $0.297\,\pm\,0.006$ & $\mathbf{0.386}\,\pm\,0.004$ & $0.242\,\pm\,0.106$ & $0.343\,\pm\,0.030$ & $0.260\,\pm\,0.000$ & $0.325$ & $\underline{0.343}$ \\
Uterus & $0.091\,\pm\,0.000$ & $0.083\,\pm\,0.000$ & $\underline{0.097}\,\pm\,0.003$ & $0.053\,\pm\,0.015$ & $0.078\,\pm\,0.008$ & $0.049\,\pm\,0.031$ & $0.074$ & $\mathbf{0.105}$ \\
Breast & $\underline{0.120}\,\pm\,0.024$ & $0.120\,\pm\,0.009$ & $0.051\,\pm\,0.015$ & $0.086\,\pm\,0.004$ & $0.070\,\pm\,0.033$ & $\mathbf{0.121}\,\pm\,0.031$ & $0.065$ & $0.072$ \\
\cmidrule(lr){1-9}
\textit{Macro} & $0.109$ & $0.098$ & $\mathbf{0.125}$ & $0.076$ & $0.093$ & $0.092$ & $0.101$ & $\underline{0.123}$ \\
\midrule
\multicolumn{9}{l}{\(R^2_{\mathrm{local}}\;\uparrow\)} \\
Brain & $-0.118\,\pm\,0.003$ & $-0.013\,\pm\,0.002$ & $\mathbf{0.007}\,\pm\,0.000$ & $-0.031\,\pm\,0.007$ & $-0.374\,\pm\,0.071$ & $\underline{0.003}\,\pm\,0.006$ & $-0.255$ & $-0.036$ \\
Colon & $\underline{0.135}\,\pm\,0.001$ & $0.099\,\pm\,0.002$ & $0.050\,\pm\,0.016$ & $0.002\,\pm\,0.003$ & $-0.081\,\pm\,0.058$ & $0.038\,\pm\,0.030$ & $-0.072$ & $\mathbf{0.143}$ \\
Heart & $-0.074\,\pm\,0.011$ & $\underline{-0.013}\,\pm\,0.003$ & $\mathbf{0.002}\,\pm\,0.000$ & $-0.054\,\pm\,0.018$ & $-0.235\,\pm\,0.126$ & $-0.052\,\pm\,0.046$ & $-0.340$ & $-0.046$ \\
Kidney & $0.013\,\pm\,0.000$ & $0.024\,\pm\,0.002$ & $\underline{0.027}\,\pm\,0.003$ & $-0.022\,\pm\,0.013$ & $-0.182\,\pm\,0.012$ & $0.002\,\pm\,0.002$ & $-0.107$ & $\mathbf{0.038}$ \\
Liver & $-0.009\,\pm\,0.001$ & $-0.008\,\pm\,0.004$ & $\mathbf{0.030}\,\pm\,0.008$ & $0.006\,\pm\,0.009$ & $-0.144\,\pm\,0.023$ & $0.010\,\pm\,0.005$ & $-0.077$ & $\underline{0.029}$ \\
Lung & $\mathbf{0.190}\,\pm\,0.005$ & $0.070\,\pm\,0.001$ & $0.086\,\pm\,0.028$ & $-0.089\,\pm\,0.296$ & $0.022\,\pm\,0.020$ & $0.081\,\pm\,0.034$ & $\underline{0.139}$ & $0.122$ \\
Prostate & $-0.246\,\pm\,0.084$ & $-0.076\,\pm\,0.028$ & $\underline{-0.003}\,\pm\,0.013$ & $-0.218\,\pm\,0.015$ & $-0.019\,\pm\,0.053$ & $\mathbf{0.011}\,\pm\,0.011$ & $-0.182$ & $-0.056$ \\
Skin & $\mathbf{0.375}\,\pm\,0.011$ & $\underline{0.336}\,\pm\,0.001$ & $0.057\,\pm\,0.007$ & $0.218\,\pm\,0.122$ & $0.263\,\pm\,0.052$ & $0.231\,\pm\,0.041$ & $0.292$ & $0.327$ \\
Uterus & $0.055\,\pm\,0.005$ & $\underline{0.085}\,\pm\,0.000$ & $0.053\,\pm\,0.005$ & $0.037\,\pm\,0.011$ & $-0.129\,\pm\,0.038$ & $0.036\,\pm\,0.039$ & $-0.069$ & $\mathbf{0.102}$ \\
Breast & $\underline{0.048}\,\pm\,0.004$ & $\mathbf{0.058}\,\pm\,0.001$ & $0.008\,\pm\,0.001$ & $-0.150\,\pm\,0.060$ & $-0.097\,\pm\,0.003$ & $0.013\,\pm\,0.002$ & $-0.041$ & $0.042$ \\
\cmidrule(lr){1-9}
\textit{Macro} & $0.037$ & $\underline{0.056}$ & $0.032$ & $-0.030$ & $-0.098$ & $0.037$ & $-0.071$ & $\mathbf{0.066}$ \\
\midrule
\multicolumn{9}{l}{\(R^2_{\mathrm{flat}}\;\uparrow\)} \\
Brain & $0.629\,\pm\,0.014$ & $\underline{0.652}\,\pm\,0.001$ & $0.641\,\pm\,0.001$ & $\mathbf{0.660}\,\pm\,0.001$ & $0.553\,\pm\,0.011$ & $0.644\,\pm\,0.004$ & $0.599$ & $0.652$ \\
Colon & $0.505\,\pm\,0.031$ & $\underline{0.523}\,\pm\,0.020$ & $0.488\,\pm\,0.007$ & $0.459\,\pm\,0.010$ & $0.173\,\pm\,0.161$ & $0.484\,\pm\,0.012$ & $0.453$ & $\mathbf{0.534}$ \\
Heart & $0.292\,\pm\,0.016$ & $0.341\,\pm\,0.012$ & $0.245\,\pm\,0.001$ & $0.373\,\pm\,0.012$ & $\mathbf{0.491}\,\pm\,0.057$ & $\underline{0.471}\,\pm\,0.023$ & $0.336$ & $0.330$ \\
Kidney & $0.473\,\pm\,0.010$ & $0.448\,\pm\,0.004$ & $\underline{0.499}\,\pm\,0.002$ & $0.443\,\pm\,0.021$ & $0.425\,\pm\,0.015$ & $0.499\,\pm\,0.034$ & $0.491$ & $\mathbf{0.517}$ \\
Liver & $0.738\,\pm\,0.005$ & $\underline{0.745}\,\pm\,0.003$ & $0.727\,\pm\,0.000$ & $0.745\,\pm\,0.019$ & $0.731\,\pm\,0.018$ & $0.744\,\pm\,0.003$ & $0.741$ & $\mathbf{0.748}$ \\
Lung & $\mathbf{0.323}\,\pm\,0.016$ & $0.257\,\pm\,0.011$ & $0.201\,\pm\,0.006$ & $0.115\,\pm\,0.382$ & $0.093\,\pm\,0.040$ & $0.178\,\pm\,0.023$ & $\underline{0.279}$ & $0.248$ \\
Prostate & $0.141\,\pm\,0.085$ & $0.135\,\pm\,0.043$ & $\underline{0.344}\,\pm\,0.007$ & $0.175\,\pm\,0.044$ & $0.134\,\pm\,0.086$ & $\mathbf{0.397}\,\pm\,0.081$ & $0.199$ & $0.281$ \\
Skin & $0.404\,\pm\,0.002$ & $\underline{0.418}\,\pm\,0.006$ & $0.238\,\pm\,0.005$ & $0.315\,\pm\,0.112$ & $0.377\,\pm\,0.060$ & $0.331\,\pm\,0.001$ & $0.403$ & $\mathbf{0.428}$ \\
Uterus & $0.668\,\pm\,0.008$ & $0.642\,\pm\,0.002$ & $0.677\,\pm\,0.003$ & $\underline{0.691}\,\pm\,0.006$ & $0.628\,\pm\,0.034$ & $0.642\,\pm\,0.065$ & $0.656$ & $\mathbf{0.704}$ \\
Breast & $\underline{0.420}\,\pm\,0.013$ & $0.418\,\pm\,0.011$ & $0.397\,\pm\,0.001$ & $0.086\,\pm\,0.099$ & $0.349\,\pm\,0.032$ & $\mathbf{0.424}\,\pm\,0.014$ & $0.378$ & $0.402$ \\
\cmidrule(lr){1-9}
\textit{Macro} & $0.459$ & $0.458$ & $0.446$ & $0.406$ & $0.395$ & $\underline{0.481}$ & $0.454$ & $\mathbf{0.484}$ \\
\bottomrule
\end{tabular}%
}
\end{table*}

\clearpage
\begin{table*}[!ht]
\centering
\caption{Cohort-level test results with DINOv2-Large and \(G=1500\). Neural entries are mean \(\pm\) seed standard deviation over three runs. Linear and CGL-Linear are deterministic. Bold and underlining mark the best and second-best unrounded means in each row. Macro rows average cohort means without weighting by cohort size. Higher is better.}
\label{tab:cohort-dinov2-large-1500}
\scriptsize
\setlength{\tabcolsep}{2.0pt}
\renewcommand{\arraystretch}{1.08}
\resizebox{\textwidth}{!}{%
\begin{tabular}{lrrrrrrrr}
\toprule
Cohort & HiST & HistoPrism & HEXST & STFlow & BLEEP & SEPAL & Linear & CGL-Linear \\
\midrule
\multicolumn{9}{l}{\(\mathrm{PCC}_{\mathrm{flat}}\;\uparrow\)} \\
Brain & $0.796\,\pm\,0.006$ & $\underline{0.819}\,\pm\,0.001$ & $0.814\,\pm\,0.001$ & $\mathbf{0.821}\,\pm\,0.001$ & $0.773\,\pm\,0.005$ & $0.813\,\pm\,0.002$ & $0.793$ & $0.818$ \\
Colon & $0.715\,\pm\,0.022$ & $\underline{0.740}\,\pm\,0.014$ & $0.713\,\pm\,0.000$ & $0.693\,\pm\,0.002$ & $0.619\,\pm\,0.019$ & $0.701\,\pm\,0.003$ & $0.690$ & $\mathbf{0.744}$ \\
Heart & $\underline{0.742}\,\pm\,0.003$ & $0.739\,\pm\,0.002$ & $0.658\,\pm\,0.058$ & $0.709\,\pm\,0.002$ & $\mathbf{0.750}\,\pm\,0.004$ & $0.732\,\pm\,0.009$ & $0.739$ & $0.734$ \\
Kidney & $0.738\,\pm\,0.009$ & $0.730\,\pm\,0.007$ & $\underline{0.745}\,\pm\,0.001$ & $0.714\,\pm\,0.002$ & $0.718\,\pm\,0.017$ & $0.744\,\pm\,0.003$ & $0.741$ & $\mathbf{0.756}$ \\
Liver & $0.863\,\pm\,0.005$ & $\underline{0.873}\,\pm\,0.001$ & $0.864\,\pm\,0.000$ & $\mathbf{0.875}\,\pm\,0.002$ & $0.827\,\pm\,0.032$ & $0.857\,\pm\,0.011$ & $0.867$ & $0.872$ \\
Lung & $\mathbf{0.632}\,\pm\,0.001$ & $0.555\,\pm\,0.044$ & $0.504\,\pm\,0.155$ & $0.452\,\pm\,0.094$ & $0.559\,\pm\,0.008$ & $0.608\,\pm\,0.001$ & $0.589$ & $\underline{0.623}$ \\
Prostate & $0.585\,\pm\,0.022$ & $0.615\,\pm\,0.007$ & $\underline{0.695}\,\pm\,0.001$ & $0.519\,\pm\,0.084$ & $0.508\,\pm\,0.110$ & $\mathbf{0.719}\,\pm\,0.008$ & $0.581$ & $0.679$ \\
Skin & $\mathbf{0.700}\,\pm\,0.007$ & $0.646\,\pm\,0.018$ & $0.508\,\pm\,0.009$ & $0.613\,\pm\,0.017$ & $\underline{0.686}\,\pm\,0.003$ & $0.620\,\pm\,0.025$ & $0.660$ & $0.671$ \\
Uterus & $0.820\,\pm\,0.015$ & $0.828\,\pm\,0.002$ & $0.753\,\pm\,0.102$ & $\underline{0.831}\,\pm\,0.003$ & $0.796\,\pm\,0.024$ & $0.784\,\pm\,0.086$ & $0.809$ & $\mathbf{0.847}$ \\
Breast & $0.606\,\pm\,0.007$ & $\mathbf{0.635}\,\pm\,0.007$ & $0.608\,\pm\,0.001$ & $0.589\,\pm\,0.001$ & $0.591\,\pm\,0.009$ & $\underline{0.620}\,\pm\,0.008$ & $0.599$ & $0.614$ \\
\cmidrule(lr){1-9}
\textit{Macro} & $0.720$ & $0.718$ & $0.686$ & $0.682$ & $0.683$ & $\underline{0.720}$ & $0.707$ & $\mathbf{0.736}$ \\
\midrule
\multicolumn{9}{l}{\(\mathrm{PCC}_{\mathrm{gene}}\;\uparrow\)} \\
Brain & $0.019\,\pm\,0.005$ & $0.008\,\pm\,0.002$ & $0.022\,\pm\,0.001$ & $0.023\,\pm\,0.001$ & $0.022\,\pm\,0.005$ & $0.003\,\pm\,0.002$ & $\underline{0.029}$ & $\mathbf{0.030}$ \\
Colon & $0.160\,\pm\,0.021$ & $0.172\,\pm\,0.031$ & $\mathbf{0.225}\,\pm\,0.009$ & $0.004\,\pm\,0.003$ & $0.129\,\pm\,0.011$ & $0.092\,\pm\,0.001$ & $0.132$ & $\underline{0.183}$ \\
Heart & $0.014\,\pm\,0.004$ & $0.016\,\pm\,0.004$ & $\underline{0.035}\,\pm\,0.007$ & $0.030\,\pm\,0.001$ & $0.027\,\pm\,0.004$ & $0.028\,\pm\,0.001$ & $0.024$ & $\mathbf{0.038}$ \\
Kidney & $\mathbf{0.067}\,\pm\,0.001$ & $0.043\,\pm\,0.002$ & $0.059\,\pm\,0.001$ & $0.036\,\pm\,0.004$ & $0.056\,\pm\,0.006$ & $0.029\,\pm\,0.009$ & $0.059$ & $\underline{0.065}$ \\
Liver & $0.078\,\pm\,0.012$ & $0.065\,\pm\,0.001$ & $0.062\,\pm\,0.013$ & $0.086\,\pm\,0.007$ & $0.049\,\pm\,0.040$ & $0.067\,\pm\,0.015$ & $\underline{0.089}$ & $\mathbf{0.092}$ \\
Lung & $0.222\,\pm\,0.034$ & $0.193\,\pm\,0.019$ & $\mathbf{0.297}\,\pm\,0.032$ & $0.152\,\pm\,0.040$ & $0.219\,\pm\,0.013$ & $0.159\,\pm\,0.013$ & $0.221$ & $\underline{0.259}$ \\
Prostate & $-0.016\,\pm\,0.006$ & $-0.006\,\pm\,0.010$ & $0.003\,\pm\,0.006$ & $-0.045\,\pm\,0.024$ & $0.016\,\pm\,0.023$ & $\mathbf{0.044}\,\pm\,0.007$ & $-0.020$ & $\underline{0.026}$ \\
Skin & $0.337\,\pm\,0.000$ & $0.281\,\pm\,0.013$ & $\mathbf{0.391}\,\pm\,0.002$ & $0.243\,\pm\,0.022$ & $\underline{0.362}\,\pm\,0.005$ & $0.268\,\pm\,0.015$ & $0.320$ & $0.338$ \\
Uterus & $0.084\,\pm\,0.004$ & $0.076\,\pm\,0.001$ & $\underline{0.094}\,\pm\,0.000$ & $0.057\,\pm\,0.006$ & $0.072\,\pm\,0.008$ & $0.040\,\pm\,0.029$ & $0.067$ & $\mathbf{0.095}$ \\
Breast & $\underline{0.105}\,\pm\,0.031$ & $\mathbf{0.125}\,\pm\,0.019$ & $0.059\,\pm\,0.024$ & $0.084\,\pm\,0.004$ & $0.078\,\pm\,0.007$ & $0.075\,\pm\,0.024$ & $0.066$ & $0.074$ \\
\cmidrule(lr){1-9}
\textit{Macro} & $0.107$ & $0.097$ & $\mathbf{0.125}$ & $0.067$ & $0.103$ & $0.080$ & $0.099$ & $\underline{0.120}$ \\
\midrule
\multicolumn{9}{l}{\(R^2_{\mathrm{local}}\;\uparrow\)} \\
Brain & $-0.139\,\pm\,0.015$ & $-0.011\,\pm\,0.002$ & $\mathbf{0.008}\,\pm\,0.000$ & $-0.028\,\pm\,0.005$ & $-0.305\,\pm\,0.016$ & $\underline{0.006}\,\pm\,0.001$ & $-0.234$ & $-0.035$ \\
Colon & $\underline{0.142}\,\pm\,0.014$ & $0.109\,\pm\,0.017$ & $0.058\,\pm\,0.002$ & $0.000\,\pm\,0.001$ & $-0.068\,\pm\,0.055$ & $0.014\,\pm\,0.002$ & $-0.040$ & $\mathbf{0.157}$ \\
Heart & $-0.107\,\pm\,0.012$ & $\mathbf{-0.002}\,\pm\,0.004$ & $-0.182\,\pm\,0.260$ & $-0.065\,\pm\,0.020$ & $-0.156\,\pm\,0.016$ & $-0.037\,\pm\,0.042$ & $-0.281$ & $\underline{-0.035}$ \\
Kidney & $\underline{0.029}\,\pm\,0.009$ & $0.024\,\pm\,0.004$ & $0.017\,\pm\,0.001$ & $-0.033\,\pm\,0.004$ & $-0.132\,\pm\,0.033$ & $-0.007\,\pm\,0.003$ & $-0.099$ & $\mathbf{0.039}$ \\
Liver & $-0.010\,\pm\,0.019$ & $-0.004\,\pm\,0.000$ & $\underline{0.025}\,\pm\,0.017$ & $0.011\,\pm\,0.024$ & $-0.229\,\pm\,0.012$ & $0.003\,\pm\,0.014$ & $-0.058$ & $\mathbf{0.035}$ \\
Lung & $\mathbf{0.191}\,\pm\,0.030$ & $0.090\,\pm\,0.017$ & $-0.159\,\pm\,0.329$ & $-0.234\,\pm\,0.281$ & $0.118\,\pm\,0.015$ & $0.068\,\pm\,0.001$ & $\underline{0.140}$ & $0.120$ \\
Prostate & $-0.211\,\pm\,0.006$ & $-0.078\,\pm\,0.006$ & $\underline{-0.010}\,\pm\,0.004$ & $-0.433\,\pm\,0.213$ & $-0.122\,\pm\,0.009$ & $\mathbf{0.015}\,\pm\,0.008$ & $-0.181$ & $-0.057$ \\
Skin & $\mathbf{0.415}\,\pm\,0.001$ & $0.325\,\pm\,0.014$ & $0.054\,\pm\,0.013$ & $0.228\,\pm\,0.023$ & $0.301\,\pm\,0.003$ & $0.267\,\pm\,0.007$ & $0.300$ & $\underline{0.332}$ \\
Uterus & $0.057\,\pm\,0.004$ & $\underline{0.079}\,\pm\,0.001$ & $-0.458\,\pm\,0.710$ & $0.044\,\pm\,0.003$ & $-0.152\,\pm\,0.039$ & $0.031\,\pm\,0.036$ & $-0.065$ & $\mathbf{0.099}$ \\
Breast & $\underline{0.042}\,\pm\,0.008$ & $\mathbf{0.052}\,\pm\,0.006$ & $0.006\,\pm\,0.002$ & $-0.106\,\pm\,0.024$ & $-0.073\,\pm\,0.003$ & $0.017\,\pm\,0.003$ & $-0.038$ & $0.037$ \\
\cmidrule(lr){1-9}
\textit{Macro} & $0.041$ & $\underline{0.058}$ & $-0.064$ & $-0.062$ & $-0.082$ & $0.038$ & $-0.056$ & $\mathbf{0.069}$ \\
\midrule
\multicolumn{9}{l}{\(R^2_{\mathrm{flat}}\;\uparrow\)} \\
Brain & $0.632\,\pm\,0.008$ & $0.666\,\pm\,0.001$ & $0.661\,\pm\,0.002$ & $\mathbf{0.674}\,\pm\,0.001$ & $0.596\,\pm\,0.007$ & $0.657\,\pm\,0.002$ & $0.622$ & $\underline{0.669}$ \\
Colon & $0.476\,\pm\,0.004$ & $\underline{0.529}\,\pm\,0.037$ & $0.503\,\pm\,0.002$ & $0.476\,\pm\,0.003$ & $0.332\,\pm\,0.007$ & $0.485\,\pm\,0.003$ & $0.475$ & $\mathbf{0.549}$ \\
Heart & $0.251\,\pm\,0.142$ & $0.337\,\pm\,0.020$ & $-0.002\,\pm\,0.424$ & $0.424\,\pm\,0.007$ & $\mathbf{0.552}\,\pm\,0.003$ & $\underline{0.473}\,\pm\,0.009$ & $0.355$ & $0.372$ \\
Kidney & $0.529\,\pm\,0.022$ & $0.528\,\pm\,0.008$ & $\underline{0.555}\,\pm\,0.001$ & $0.492\,\pm\,0.013$ & $0.511\,\pm\,0.025$ & $0.549\,\pm\,0.002$ & $0.544$ & $\mathbf{0.571}$ \\
Liver & $0.742\,\pm\,0.008$ & $0.755\,\pm\,0.002$ & $0.736\,\pm\,0.001$ & $\mathbf{0.764}\,\pm\,0.005$ & $0.664\,\pm\,0.059$ & $0.716\,\pm\,0.040$ & $0.751$ & $\underline{0.755}$ \\
Lung & $\mathbf{0.366}\,\pm\,0.033$ & $0.279\,\pm\,0.022$ & $-0.005\,\pm\,0.317$ & $0.005\,\pm\,0.282$ & $0.220\,\pm\,0.008$ & $0.211\,\pm\,0.024$ & $\underline{0.284}$ & $0.248$ \\
Prostate & $0.207\,\pm\,0.031$ & $0.124\,\pm\,0.011$ & $\underline{0.318}\,\pm\,0.002$ & $0.167\,\pm\,0.072$ & $0.071\,\pm\,0.152$ & $\mathbf{0.375}\,\pm\,0.054$ & $0.185$ & $0.269$ \\
Skin & $\mathbf{0.473}\,\pm\,0.006$ & $0.414\,\pm\,0.025$ & $0.256\,\pm\,0.009$ & $0.356\,\pm\,0.023$ & $0.442\,\pm\,0.001$ & $0.343\,\pm\,0.069$ & $0.429$ & $\underline{0.450}$ \\
Uterus & $0.658\,\pm\,0.035$ & $0.676\,\pm\,0.005$ & $0.525\,\pm\,0.204$ & $\underline{0.690}\,\pm\,0.004$ & $0.617\,\pm\,0.052$ & $0.583\,\pm\,0.165$ & $0.649$ & $\mathbf{0.698}$ \\
Breast & $0.355\,\pm\,0.007$ & $\mathbf{0.401}\,\pm\,0.009$ & $0.368\,\pm\,0.002$ & $0.134\,\pm\,0.038$ & $0.337\,\pm\,0.007$ & $\underline{0.382}\,\pm\,0.009$ & $0.351$ & $0.373$ \\
\cmidrule(lr){1-9}
\textit{Macro} & $0.469$ & $0.471$ & $0.392$ & $0.418$ & $0.434$ & $\underline{0.477}$ & $0.465$ & $\mathbf{0.496}$ \\
\bottomrule
\end{tabular}%
}
\end{table*}

\clearpage
\begin{table*}[!ht]
\centering
\caption{Cohort-level test results with DINOv2-Large and \(G=2000\). Neural entries are mean \(\pm\) seed standard deviation over three runs. Linear and CGL-Linear are deterministic. Bold and underlining mark the best and second-best unrounded means in each row. Macro rows average cohort means without weighting by cohort size. Higher is better.}
\label{tab:cohort-dinov2-large-2000}
\scriptsize
\setlength{\tabcolsep}{2.0pt}
\renewcommand{\arraystretch}{1.08}
\resizebox{\textwidth}{!}{%
\begin{tabular}{lrrrrrrrr}
\toprule
Cohort & HiST & HistoPrism & HEXST & STFlow & BLEEP & SEPAL & Linear & CGL-Linear \\
\midrule
\multicolumn{9}{l}{\(\mathrm{PCC}_{\mathrm{flat}}\;\uparrow\)} \\
Brain & $0.797\,\pm\,0.000$ & $0.810\,\pm\,0.000$ & $0.809\,\pm\,0.000$ & $\mathbf{0.815}\,\pm\,0.002$ & $0.771\,\pm\,0.005$ & $0.809\,\pm\,0.001$ & $0.788$ & $\underline{0.812}$ \\
Colon & $0.729\,\pm\,0.013$ & $\underline{0.731}\,\pm\,0.012$ & $0.702\,\pm\,0.003$ & $0.688\,\pm\,0.003$ & $0.659\,\pm\,0.013$ & $0.693\,\pm\,0.005$ & $0.686$ & $\mathbf{0.739}$ \\
Heart & $\mathbf{0.748}\,\pm\,0.002$ & $0.738\,\pm\,0.002$ & $0.668\,\pm\,0.058$ & $0.719\,\pm\,0.008$ & $0.739\,\pm\,0.002$ & $0.731\,\pm\,0.000$ & $\underline{0.742}$ & $0.740$ \\
Kidney & $0.736\,\pm\,0.014$ & $0.732\,\pm\,0.002$ & $\underline{0.741}\,\pm\,0.000$ & $0.706\,\pm\,0.023$ & $0.716\,\pm\,0.021$ & $0.740\,\pm\,0.002$ & $0.737$ & $\mathbf{0.752}$ \\
Liver & $0.867\,\pm\,0.004$ & $0.866\,\pm\,0.002$ & $0.859\,\pm\,0.000$ & $\mathbf{0.868}\,\pm\,0.004$ & $0.859\,\pm\,0.008$ & $0.851\,\pm\,0.018$ & $0.865$ & $\underline{0.868}$ \\
Lung & $\underline{0.605}\,\pm\,0.017$ & $0.563\,\pm\,0.029$ & $0.482\,\pm\,0.147$ & $0.586\,\pm\,0.060$ & $0.522\,\pm\,0.043$ & $0.596\,\pm\,0.018$ & $0.577$ & $\mathbf{0.608}$ \\
Prostate & $0.569\,\pm\,0.027$ & $0.595\,\pm\,0.004$ & $\underline{0.676}\,\pm\,0.000$ & $0.551\,\pm\,0.016$ & $0.575\,\pm\,0.031$ & $\mathbf{0.695}\,\pm\,0.002$ & $0.558$ & $0.658$ \\
Skin & $\mathbf{0.679}\,\pm\,0.017$ & $0.658\,\pm\,0.022$ & $0.501\,\pm\,0.016$ & $0.641\,\pm\,0.036$ & $0.647\,\pm\,0.060$ & $0.650\,\pm\,0.020$ & $0.664$ & $\underline{0.675}$ \\
Uterus & $0.802\,\pm\,0.007$ & $0.821\,\pm\,0.007$ & $\underline{0.821}\,\pm\,0.000$ & $0.814\,\pm\,0.017$ & $0.783\,\pm\,0.007$ & $0.762\,\pm\,0.092$ & $0.802$ & $\mathbf{0.840}$ \\
Breast & $\underline{0.611}\,\pm\,0.003$ & $\mathbf{0.621}\,\pm\,0.005$ & $0.588\,\pm\,0.000$ & $0.567\,\pm\,0.014$ & $0.546\,\pm\,0.009$ & $0.610\,\pm\,0.012$ & $0.576$ & $0.594$ \\
\cmidrule(lr){1-9}
\textit{Macro} & $\underline{0.714}$ & $0.714$ & $0.685$ & $0.696$ & $0.682$ & $0.714$ & $0.700$ & $\mathbf{0.729}$ \\
\midrule
\multicolumn{9}{l}{\(\mathrm{PCC}_{\mathrm{gene}}\;\uparrow\)} \\
Brain & $0.021\,\pm\,0.003$ & $0.006\,\pm\,0.002$ & $0.023\,\pm\,0.004$ & $0.027\,\pm\,0.002$ & $0.022\,\pm\,0.003$ & $0.004\,\pm\,0.001$ & $\underline{0.029}$ & $\mathbf{0.031}$ \\
Colon & $0.149\,\pm\,0.008$ & $0.157\,\pm\,0.027$ & $\mathbf{0.222}\,\pm\,0.016$ & $0.020\,\pm\,0.034$ & $0.146\,\pm\,0.010$ & $0.081\,\pm\,0.014$ & $0.132$ & $\underline{0.182}$ \\
Heart & $0.015\,\pm\,0.000$ & $0.015\,\pm\,0.001$ & $0.029\,\pm\,0.010$ & $\underline{0.032}\,\pm\,0.001$ & $0.023\,\pm\,0.002$ & $0.027\,\pm\,0.003$ & $0.022$ & $\mathbf{0.035}$ \\
Kidney & $\mathbf{0.062}\,\pm\,0.003$ & $0.038\,\pm\,0.004$ & $0.052\,\pm\,0.001$ & $0.027\,\pm\,0.003$ & $0.052\,\pm\,0.010$ & $0.025\,\pm\,0.002$ & $0.054$ & $\underline{0.060}$ \\
Liver & $0.095\,\pm\,0.001$ & $0.069\,\pm\,0.003$ & $0.063\,\pm\,0.003$ & $0.086\,\pm\,0.010$ & $\mathbf{0.100}\,\pm\,0.016$ & $0.063\,\pm\,0.017$ & $0.097$ & $\underline{0.098}$ \\
Lung & $0.225\,\pm\,0.010$ & $0.212\,\pm\,0.000$ & $\mathbf{0.277}\,\pm\,0.019$ & $0.208\,\pm\,0.040$ & $0.198\,\pm\,0.037$ & $0.166\,\pm\,0.050$ & $0.216$ & $\underline{0.255}$ \\
Prostate & $-0.021\,\pm\,0.007$ & $-0.017\,\pm\,0.006$ & $-0.002\,\pm\,0.013$ & $-0.016\,\pm\,0.007$ & $\mathbf{0.036}\,\pm\,0.034$ & $\underline{0.035}\,\pm\,0.003$ & $-0.028$ & $0.020$ \\
Skin & $0.299\,\pm\,0.002$ & $0.279\,\pm\,0.016$ & $\mathbf{0.362}\,\pm\,0.008$ & $0.257\,\pm\,0.025$ & $0.308\,\pm\,0.058$ & $0.273\,\pm\,0.004$ & $0.310$ & $\underline{0.328}$ \\
Uterus & $0.076\,\pm\,0.013$ & $0.073\,\pm\,0.003$ & $\underline{0.084}\,\pm\,0.004$ & $0.061\,\pm\,0.003$ & $0.067\,\pm\,0.009$ & $0.032\,\pm\,0.033$ & $0.063$ & $\mathbf{0.089}$ \\
Breast & $\mathbf{0.133}\,\pm\,0.000$ & $\underline{0.132}\,\pm\,0.014$ & $0.066\,\pm\,0.010$ & $0.068\,\pm\,0.012$ & $0.054\,\pm\,0.010$ & $0.104\,\pm\,0.043$ & $0.066$ & $0.074$ \\
\cmidrule(lr){1-9}
\textit{Macro} & $0.105$ & $0.096$ & $\mathbf{0.118}$ & $0.077$ & $0.101$ & $0.081$ & $0.096$ & $\underline{0.117}$ \\
\midrule
\multicolumn{9}{l}{\(R^2_{\mathrm{local}}\;\uparrow\)} \\
Brain & $-0.108\,\pm\,0.001$ & $-0.009\,\pm\,0.001$ & $\mathbf{0.009}\,\pm\,0.001$ & $-0.027\,\pm\,0.000$ & $-0.285\,\pm\,0.010$ & $\underline{0.007}\,\pm\,0.000$ & $-0.212$ & $-0.030$ \\
Colon & $\underline{0.148}\,\pm\,0.007$ & $0.106\,\pm\,0.007$ & $0.040\,\pm\,0.013$ & $0.013\,\pm\,0.017$ & $0.011\,\pm\,0.023$ & $0.024\,\pm\,0.022$ & $-0.022$ & $\mathbf{0.164}$ \\
Heart & $-0.090\,\pm\,0.043$ & $\mathbf{-0.009}\,\pm\,0.006$ & $-0.212\,\pm\,0.301$ & $-0.067\,\pm\,0.032$ & $-0.189\,\pm\,0.005$ & $\underline{-0.010}\,\pm\,0.005$ & $-0.253$ & $-0.030$ \\
Kidney & $\underline{0.024}\,\pm\,0.000$ & $0.023\,\pm\,0.001$ & $0.012\,\pm\,0.001$ & $-0.038\,\pm\,0.026$ & $-0.118\,\pm\,0.011$ & $-0.004\,\pm\,0.009$ & $-0.094$ & $\mathbf{0.039}$ \\
Liver & $\underline{0.033}\,\pm\,0.004$ & $-0.006\,\pm\,0.002$ & $0.013\,\pm\,0.001$ & $0.000\,\pm\,0.006$ & $-0.094\,\pm\,0.034$ & $0.010\,\pm\,0.001$ & $-0.050$ & $\mathbf{0.037}$ \\
Lung & $\mathbf{0.189}\,\pm\,0.013$ & $0.113\,\pm\,0.006$ & $-0.042\,\pm\,0.130$ & $0.081\,\pm\,0.087$ & $0.094\,\pm\,0.036$ & $0.069\,\pm\,0.027$ & $\underline{0.134}$ & $0.113$ \\
Prostate & $-0.200\,\pm\,0.033$ & $-0.076\,\pm\,0.023$ & $\underline{-0.005}\,\pm\,0.003$ & $-0.209\,\pm\,0.028$ & $-0.017\,\pm\,0.067$ & $\mathbf{0.004}\,\pm\,0.001$ & $-0.176$ & $-0.056$ \\
Skin & $\mathbf{0.387}\,\pm\,0.002$ & $0.334\,\pm\,0.009$ & $0.040\,\pm\,0.021$ & $0.253\,\pm\,0.054$ & $0.272\,\pm\,0.074$ & $0.284\,\pm\,0.018$ & $0.303$ & $\underline{0.334}$ \\
Uterus & $0.052\,\pm\,0.017$ & $\underline{0.081}\,\pm\,0.005$ & $0.036\,\pm\,0.003$ & $0.004\,\pm\,0.064$ & $-0.058\,\pm\,0.005$ & $0.024\,\pm\,0.034$ & $-0.062$ & $\mathbf{0.093}$ \\
Breast & $\underline{0.047}\,\pm\,0.011$ & $\mathbf{0.054}\,\pm\,0.002$ & $0.006\,\pm\,0.001$ & $-0.085\,\pm\,0.045$ & $-0.163\,\pm\,0.157$ & $0.013\,\pm\,0.006$ & $-0.040$ & $0.036$ \\
\cmidrule(lr){1-9}
\textit{Macro} & $0.048$ & $\underline{0.061}$ & $-0.010$ & $-0.007$ & $-0.055$ & $0.042$ & $-0.047$ & $\mathbf{0.070}$ \\
\midrule
\multicolumn{9}{l}{\(R^2_{\mathrm{flat}}\;\uparrow\)} \\
Brain & $0.632\,\pm\,0.000$ & $0.654\,\pm\,0.000$ & $0.653\,\pm\,0.001$ & $\mathbf{0.663}\,\pm\,0.004$ & $0.591\,\pm\,0.009$ & $0.649\,\pm\,0.004$ & $0.615$ & $\underline{0.660}$ \\
Colon & $\underline{0.525}\,\pm\,0.012$ & $0.499\,\pm\,0.045$ & $0.487\,\pm\,0.004$ & $0.468\,\pm\,0.002$ & $0.375\,\pm\,0.016$ & $0.475\,\pm\,0.002$ & $0.470$ & $\mathbf{0.543}$ \\
Heart & $0.086\,\pm\,0.170$ & $0.327\,\pm\,0.001$ & $-0.009\,\pm\,0.472$ & $0.455\,\pm\,0.015$ & $\mathbf{0.510}\,\pm\,0.016$ & $\underline{0.482}\,\pm\,0.009$ & $0.358$ & $0.386$ \\
Kidney & $0.533\,\pm\,0.031$ & $0.530\,\pm\,0.009$ & $\underline{0.548}\,\pm\,0.000$ & $0.481\,\pm\,0.042$ & $0.503\,\pm\,0.036$ & $0.547\,\pm\,0.004$ & $0.537$ & $\mathbf{0.566}$ \\
Liver & $\underline{0.749}\,\pm\,0.006$ & $0.741\,\pm\,0.002$ & $0.726\,\pm\,0.001$ & $\mathbf{0.751}\,\pm\,0.009$ & $0.731\,\pm\,0.022$ & $0.688\,\pm\,0.080$ & $0.748$ & $0.748$ \\
Lung & $\mathbf{0.338}\,\pm\,0.053$ & $0.282\,\pm\,0.013$ & $0.032\,\pm\,0.212$ & $\underline{0.329}\,\pm\,0.085$ & $0.145\,\pm\,0.080$ & $0.193\,\pm\,0.053$ & $0.270$ & $0.232$ \\
Prostate & $0.194\,\pm\,0.059$ & $0.101\,\pm\,0.018$ & $\underline{0.290}\,\pm\,0.000$ & $0.164\,\pm\,0.037$ & $0.097\,\pm\,0.006$ & $\mathbf{0.352}\,\pm\,0.028$ & $0.162$ & $0.248$ \\
Skin & $\underline{0.437}\,\pm\,0.044$ & $0.430\,\pm\,0.029$ & $0.250\,\pm\,0.015$ & $0.397\,\pm\,0.060$ & $0.394\,\pm\,0.066$ & $0.401\,\pm\,0.028$ & $0.434$ & $\mathbf{0.455}$ \\
Uterus & $0.628\,\pm\,0.004$ & $\underline{0.665}\,\pm\,0.013$ & $0.663\,\pm\,0.002$ & $0.644\,\pm\,0.054$ & $0.592\,\pm\,0.012$ & $0.487\,\pm\,0.240$ & $0.639$ & $\mathbf{0.688}$ \\
Breast & $0.361\,\pm\,0.015$ & $\mathbf{0.384}\,\pm\,0.006$ & $0.344\,\pm\,0.000$ & $0.185\,\pm\,0.087$ & $0.203\,\pm\,0.127$ & $\underline{0.368}\,\pm\,0.011$ & $0.322$ & $0.348$ \\
\cmidrule(lr){1-9}
\textit{Macro} & $0.448$ & $0.461$ & $0.399$ & $0.453$ & $0.414$ & $\underline{0.464}$ & $0.455$ & $\mathbf{0.487}$ \\
\bottomrule
\end{tabular}%
}
\end{table*}

\clearpage
\begin{table*}[!ht]
\centering
\caption{Cohort-level test results with ResNet18 and \(G=500\). Neural entries are mean \(\pm\) seed standard deviation over three runs. Linear and CGL-Linear are deterministic. Bold and underlining mark the best and second-best unrounded means in each row. Macro rows average cohort means without weighting by cohort size. Higher is better.}
\label{tab:cohort-resnet18-500}
\scriptsize
\setlength{\tabcolsep}{2.0pt}
\renewcommand{\arraystretch}{1.08}
\resizebox{\textwidth}{!}{%
\begin{tabular}{lrrrrrrrr}
\toprule
Cohort & HiST & HistoPrism & HEXST & STFlow & BLEEP & SEPAL & Linear & CGL-Linear \\
\midrule
\multicolumn{9}{l}{\(\mathrm{PCC}_{\mathrm{flat}}\;\uparrow\)} \\
Brain & $0.761\,\pm\,0.004$ & $0.772\,\pm\,0.001$ & $0.763\,\pm\,0.002$ & $\mathbf{0.778}\,\pm\,0.002$ & $0.704\,\pm\,0.045$ & $0.774\,\pm\,0.002$ & $0.755$ & $\underline{0.774}$ \\
Colon & $0.706\,\pm\,0.024$ & $\underline{0.717}\,\pm\,0.005$ & $0.709\,\pm\,0.008$ & $0.688\,\pm\,0.002$ & $0.511\,\pm\,0.037$ & $0.686\,\pm\,0.014$ & $0.693$ & $\mathbf{0.720}$ \\
Heart & $0.710\,\pm\,0.014$ & $\underline{0.719}\,\pm\,0.003$ & $0.647\,\pm\,0.000$ & $0.654\,\pm\,0.000$ & $\mathbf{0.727}\,\pm\,0.005$ & $0.644\,\pm\,0.006$ & $0.642$ & $0.661$ \\
Kidney & $0.671\,\pm\,0.013$ & $0.679\,\pm\,0.007$ & $0.711\,\pm\,0.006$ & $0.708\,\pm\,0.001$ & $0.687\,\pm\,0.001$ & $0.710\,\pm\,0.030$ & $\underline{0.716}$ & $\mathbf{0.725}$ \\
Liver & $0.869\,\pm\,0.002$ & $\underline{0.875}\,\pm\,0.000$ & $0.856\,\pm\,0.006$ & $0.860\,\pm\,0.003$ & $0.852\,\pm\,0.009$ & $0.860\,\pm\,0.001$ & $0.853$ & $\mathbf{0.877}$ \\
Lung & $0.598\,\pm\,0.013$ & $\underline{0.638}\,\pm\,0.010$ & $0.586\,\pm\,0.035$ & $0.556\,\pm\,0.040$ & $0.473\,\pm\,0.002$ & $\mathbf{0.649}\,\pm\,0.008$ & $0.578$ & $0.619$ \\
Prostate & $0.600\,\pm\,0.014$ & $0.658\,\pm\,0.035$ & $\mathbf{0.726}\,\pm\,0.006$ & $0.656\,\pm\,0.032$ & $0.555\,\pm\,0.044$ & $\underline{0.690}\,\pm\,0.027$ & $0.540$ & $0.678$ \\
Skin & $0.611\,\pm\,0.029$ & $\underline{0.612}\,\pm\,0.030$ & $0.485\,\pm\,0.009$ & $0.583\,\pm\,0.017$ & $0.610\,\pm\,0.024$ & $0.553\,\pm\,0.095$ & $\mathbf{0.620}$ & $0.608$ \\
Uterus & $0.790\,\pm\,0.015$ & $0.792\,\pm\,0.009$ & $0.776\,\pm\,0.015$ & $0.844\,\pm\,0.001$ & $0.798\,\pm\,0.003$ & $\mathbf{0.864}\,\pm\,0.001$ & $0.836$ & $\underline{0.861}$ \\
Breast & $0.661\,\pm\,0.012$ & $\underline{0.692}\,\pm\,0.004$ & $0.672\,\pm\,0.001$ & $0.644\,\pm\,0.002$ & $0.614\,\pm\,0.009$ & $\mathbf{0.704}\,\pm\,0.009$ & $0.665$ & $0.678$ \\
\cmidrule(lr){1-9}
\textit{Macro} & $0.698$ & $\underline{0.715}$ & $0.693$ & $0.697$ & $0.653$ & $0.713$ & $0.690$ & $\mathbf{0.720}$ \\
\midrule
\multicolumn{9}{l}{\(\mathrm{PCC}_{\mathrm{gene}}\;\uparrow\)} \\
Brain & $0.024\,\pm\,0.007$ & $0.016\,\pm\,0.004$ & $0.031\,\pm\,0.004$ & $0.024\,\pm\,0.002$ & $0.026\,\pm\,0.003$ & $0.018\,\pm\,0.003$ & $\underline{0.036}$ & $\mathbf{0.038}$ \\
Colon & $\underline{0.155}\,\pm\,0.014$ & $0.121\,\pm\,0.019$ & $\mathbf{0.207}\,\pm\,0.010$ & $-0.023\,\pm\,0.011$ & $0.073\,\pm\,0.029$ & $0.094\,\pm\,0.022$ & $0.129$ & $0.152$ \\
Heart & $0.017\,\pm\,0.005$ & $\underline{0.026}\,\pm\,0.001$ & $0.025\,\pm\,0.009$ & $-0.002\,\pm\,0.002$ & $0.022\,\pm\,0.005$ & $0.018\,\pm\,0.001$ & $0.022$ & $\mathbf{0.041}$ \\
Kidney & $0.059\,\pm\,0.001$ & $0.042\,\pm\,0.008$ & $0.060\,\pm\,0.000$ & $0.005\,\pm\,0.006$ & $\underline{0.066}\,\pm\,0.011$ & $0.040\,\pm\,0.019$ & $0.064$ & $\mathbf{0.068}$ \\
Liver & $\underline{0.087}\,\pm\,0.000$ & $0.076\,\pm\,0.004$ & $0.024\,\pm\,0.013$ & $0.085\,\pm\,0.010$ & $0.080\,\pm\,0.009$ & $0.061\,\pm\,0.005$ & $0.068$ & $\mathbf{0.104}$ \\
Lung & $0.151\,\pm\,0.004$ & $0.173\,\pm\,0.014$ & $0.197\,\pm\,0.039$ & $0.138\,\pm\,0.043$ & $0.136\,\pm\,0.003$ & $\mathbf{0.233}\,\pm\,0.007$ & $0.167$ & $\underline{0.206}$ \\
Prostate & $0.033\,\pm\,0.001$ & $0.043\,\pm\,0.008$ & $\mathbf{0.106}\,\pm\,0.009$ & $0.055\,\pm\,0.018$ & $0.069\,\pm\,0.020$ & $0.057\,\pm\,0.016$ & $0.008$ & $\underline{0.074}$ \\
Skin & $0.322\,\pm\,0.028$ & $0.292\,\pm\,0.024$ & $\mathbf{0.363}\,\pm\,0.009$ & $0.278\,\pm\,0.012$ & $\underline{0.335}\,\pm\,0.023$ & $0.267\,\pm\,0.052$ & $0.334$ & $0.322$ \\
Uterus & $0.062\,\pm\,0.004$ & $0.048\,\pm\,0.000$ & $\mathbf{0.125}\,\pm\,0.003$ & $0.029\,\pm\,0.014$ & $0.062\,\pm\,0.000$ & $0.091\,\pm\,0.007$ & $0.088$ & $\underline{0.114}$ \\
Breast & $0.071\,\pm\,0.008$ & $\underline{0.121}\,\pm\,0.008$ & $0.078\,\pm\,0.012$ & $0.072\,\pm\,0.001$ & $0.035\,\pm\,0.012$ & $\mathbf{0.152}\,\pm\,0.018$ & $0.072$ & $0.086$ \\
\cmidrule(lr){1-9}
\textit{Macro} & $0.098$ & $0.096$ & $\mathbf{0.122}$ & $0.066$ & $0.090$ & $0.103$ & $0.099$ & $\underline{0.121}$ \\
\midrule
\multicolumn{9}{l}{\(R^2_{\mathrm{local}}\;\uparrow\)} \\
Brain & $-0.052\,\pm\,0.011$ & $-0.007\,\pm\,0.002$ & $\mathbf{0.004}\,\pm\,0.002$ & $-0.015\,\pm\,0.002$ & $-0.551\,\pm\,0.350$ & $\underline{0.004}\,\pm\,0.008$ & $-0.224$ & $-0.010$ \\
Colon & $\mathbf{0.090}\,\pm\,0.002$ & $0.029\,\pm\,0.009$ & $0.059\,\pm\,0.032$ & $0.000\,\pm\,0.000$ & $-0.220\,\pm\,0.134$ & $0.008\,\pm\,0.011$ & $-0.105$ & $\underline{0.079}$ \\
Heart & $-0.169\,\pm\,0.045$ & $-0.051\,\pm\,0.020$ & $\mathbf{0.005}\,\pm\,0.001$ & $\underline{0.000}\,\pm\,0.000$ & $-0.155\,\pm\,0.019$ & $-0.178\,\pm\,0.067$ & $-0.568$ & $-0.092$ \\
Kidney & $0.005\,\pm\,0.004$ & $0.010\,\pm\,0.005$ & $\underline{0.021}\,\pm\,0.002$ & $0.000\,\pm\,0.000$ & $-0.192\,\pm\,0.075$ & $-0.033\,\pm\,0.002$ & $-0.102$ & $\mathbf{0.028}$ \\
Liver & $-0.029\,\pm\,0.032$ & $0.001\,\pm\,0.005$ & $-0.013\,\pm\,0.021$ & $-0.045\,\pm\,0.044$ & $-0.257\,\pm\,0.184$ & $\underline{0.014}\,\pm\,0.002$ & $-0.164$ & $\mathbf{0.019}$ \\
Lung & $\mathbf{0.141}\,\pm\,0.014$ & $0.098\,\pm\,0.001$ & $0.074\,\pm\,0.042$ & $0.025\,\pm\,0.064$ & $0.038\,\pm\,0.001$ & $\underline{0.127}\,\pm\,0.001$ & $0.097$ & $0.106$ \\
Prostate & $0.002\,\pm\,0.006$ & $0.013\,\pm\,0.003$ & $\mathbf{0.030}\,\pm\,0.014$ & $0.006\,\pm\,0.031$ & $-0.027\,\pm\,0.009$ & $0.001\,\pm\,0.036$ & $-0.110$ & $\underline{0.029}$ \\
Skin & $\mathbf{0.342}\,\pm\,0.030$ & $\underline{0.288}\,\pm\,0.027$ & $0.065\,\pm\,0.012$ & $0.227\,\pm\,0.021$ & $0.224\,\pm\,0.043$ & $0.235\,\pm\,0.059$ & $0.257$ & $0.257$ \\
Uterus & $\underline{0.081}\,\pm\,0.004$ & $0.065\,\pm\,0.000$ & $-0.360\,\pm\,0.082$ & $0.008\,\pm\,0.004$ & $-0.195\,\pm\,0.082$ & $0.064\,\pm\,0.019$ & $-0.060$ & $\mathbf{0.095}$ \\
Breast & $\underline{0.048}\,\pm\,0.012$ & $\mathbf{0.052}\,\pm\,0.003$ & $0.014\,\pm\,0.000$ & $-0.173\,\pm\,0.001$ & $-0.217\,\pm\,0.032$ & $0.031\,\pm\,0.006$ & $-0.025$ & $0.044$ \\
\cmidrule(lr){1-9}
\textit{Macro} & $0.046$ & $\underline{0.050}$ & $-0.010$ & $0.003$ & $-0.155$ & $0.027$ & $-0.100$ & $\mathbf{0.055}$ \\
\midrule
\multicolumn{9}{l}{\(R^2_{\mathrm{flat}}\;\uparrow\)} \\
Brain & $0.576\,\pm\,0.005$ & $0.595\,\pm\,0.003$ & $0.581\,\pm\,0.003$ & $\mathbf{0.605}\,\pm\,0.004$ & $0.455\,\pm\,0.114$ & $\underline{0.598}\,\pm\,0.002$ & $0.554$ & $0.593$ \\
Colon & $0.432\,\pm\,0.058$ & $0.431\,\pm\,0.032$ & $\underline{0.495}\,\pm\,0.012$ & $0.466\,\pm\,0.003$ & $0.174\,\pm\,0.094$ & $0.465\,\pm\,0.020$ & $0.480$ & $\mathbf{0.512}$ \\
Heart & $0.230\,\pm\,0.122$ & $\underline{0.422}\,\pm\,0.004$ & $0.208\,\pm\,0.000$ & $0.250\,\pm\,0.001$ & $\mathbf{0.523}\,\pm\,0.010$ & $0.319\,\pm\,0.019$ & $0.183$ & $0.236$ \\
Kidney & $0.420\,\pm\,0.001$ & $0.426\,\pm\,0.023$ & $0.505\,\pm\,0.007$ & $0.500\,\pm\,0.002$ & $0.462\,\pm\,0.004$ & $0.501\,\pm\,0.046$ & $\underline{0.509}$ & $\mathbf{0.525}$ \\
Liver & $0.754\,\pm\,0.003$ & $\mathbf{0.763}\,\pm\,0.000$ & $0.721\,\pm\,0.012$ & $0.737\,\pm\,0.004$ & $0.722\,\pm\,0.014$ & $0.738\,\pm\,0.002$ & $0.725$ & $\underline{0.763}$ \\
Lung & $0.210\,\pm\,0.041$ & $0.210\,\pm\,0.008$ & $0.202\,\pm\,0.027$ & $\mathbf{0.284}\,\pm\,0.034$ & $0.036\,\pm\,0.008$ & $\underline{0.253}\,\pm\,0.028$ & $0.234$ & $0.211$ \\
Prostate & $0.036\,\pm\,0.016$ & $0.120\,\pm\,0.036$ & $\mathbf{0.353}\,\pm\,0.025$ & $0.181\,\pm\,0.042$ & $0.120\,\pm\,0.063$ & $\underline{0.262}\,\pm\,0.050$ & $0.088$ & $0.207$ \\
Skin & $0.367\,\pm\,0.035$ & $\underline{0.371}\,\pm\,0.041$ & $0.231\,\pm\,0.008$ & $0.322\,\pm\,0.026$ & $0.348\,\pm\,0.009$ & $0.249\,\pm\,0.155$ & $\mathbf{0.380}$ & $0.366$ \\
Uterus & $0.612\,\pm\,0.026$ & $0.603\,\pm\,0.018$ & $0.584\,\pm\,0.040$ & $0.706\,\pm\,0.001$ & $0.625\,\pm\,0.011$ & $\mathbf{0.729}\,\pm\,0.009$ & $0.690$ & $\underline{0.713}$ \\
Breast & $0.426\,\pm\,0.019$ & $\underline{0.474}\,\pm\,0.010$ & $0.451\,\pm\,0.002$ & $0.202\,\pm\,0.045$ & $0.326\,\pm\,0.038$ & $\mathbf{0.492}\,\pm\,0.011$ & $0.436$ & $0.457$ \\
\cmidrule(lr){1-9}
\textit{Macro} & $0.406$ & $0.441$ & $0.433$ & $0.425$ & $0.379$ & $\mathbf{0.461}$ & $0.428$ & $\underline{0.458}$ \\
\bottomrule
\end{tabular}%
}
\end{table*}

\clearpage
\begin{table*}[!ht]
\centering
\caption{Cohort-level test results with ResNet18 and \(G=1000\). Neural entries are mean \(\pm\) seed standard deviation over three runs. Linear and CGL-Linear are deterministic. Bold and underlining mark the best and second-best unrounded means in each row. Macro rows average cohort means without weighting by cohort size. Higher is better.}
\label{tab:cohort-resnet18-1000}
\scriptsize
\setlength{\tabcolsep}{2.0pt}
\renewcommand{\arraystretch}{1.08}
\resizebox{\textwidth}{!}{%
\begin{tabular}{lrrrrrrrr}
\toprule
Cohort & HiST & HistoPrism & HEXST & STFlow & BLEEP & SEPAL & Linear & CGL-Linear \\
\midrule
\multicolumn{9}{l}{\(\mathrm{PCC}_{\mathrm{flat}}\;\uparrow\)} \\
Brain & $0.801\,\pm\,0.006$ & $0.808\,\pm\,0.002$ & $0.796\,\pm\,0.003$ & $\mathbf{0.818}\,\pm\,0.003$ & $0.752\,\pm\,0.036$ & $0.807\,\pm\,0.001$ & $0.793$ & $\underline{0.808}$ \\
Colon & $\mathbf{0.736}\,\pm\,0.012$ & $0.710\,\pm\,0.001$ & $0.705\,\pm\,0.005$ & $0.685\,\pm\,0.001$ & $0.516\,\pm\,0.096$ & $0.682\,\pm\,0.020$ & $0.694$ & $\underline{0.718}$ \\
Heart & $\underline{0.721}\,\pm\,0.021$ & $\mathbf{0.724}\,\pm\,0.031$ & $0.674\,\pm\,0.001$ & $0.675\,\pm\,0.011$ & $0.713\,\pm\,0.008$ & $0.685\,\pm\,0.007$ & $0.679$ & $0.695$ \\
Kidney & $0.701\,\pm\,0.004$ & $0.690\,\pm\,0.001$ & $0.704\,\pm\,0.000$ & $0.699\,\pm\,0.001$ & $0.689\,\pm\,0.008$ & $\mathbf{0.718}\,\pm\,0.014$ & $0.706$ & $\underline{0.716}$ \\
Liver & $0.865\,\pm\,0.003$ & $\underline{0.871}\,\pm\,0.004$ & $0.852\,\pm\,0.006$ & $0.862\,\pm\,0.006$ & $0.841\,\pm\,0.022$ & $0.863\,\pm\,0.007$ & $0.854$ & $\mathbf{0.874}$ \\
Lung & $0.495\,\pm\,0.043$ & $\underline{0.613}\,\pm\,0.006$ & $0.582\,\pm\,0.002$ & $0.526\,\pm\,0.004$ & $0.449\,\pm\,0.044$ & $\mathbf{0.626}\,\pm\,0.004$ & $0.559$ & $0.597$ \\
Prostate & $0.571\,\pm\,0.037$ & $\underline{0.687}\,\pm\,0.014$ & $\mathbf{0.705}\,\pm\,0.004$ & $0.612\,\pm\,0.077$ & $0.550\,\pm\,0.076$ & $0.683\,\pm\,0.052$ & $0.522$ & $0.662$ \\
Skin & $\mathbf{0.654}\,\pm\,0.024$ & $0.603\,\pm\,0.001$ & $0.475\,\pm\,0.000$ & $\underline{0.651}\,\pm\,0.014$ & $0.562\,\pm\,0.056$ & $0.539\,\pm\,0.040$ & $0.632$ & $0.621$ \\
Uterus & $0.767\,\pm\,0.006$ & $0.796\,\pm\,0.003$ & $0.739\,\pm\,0.011$ & $0.817\,\pm\,0.024$ & $0.790\,\pm\,0.003$ & $0.807\,\pm\,0.059$ & $\underline{0.823}$ & $\mathbf{0.848}$ \\
Breast & $0.627\,\pm\,0.009$ & $\underline{0.658}\,\pm\,0.002$ & $0.631\,\pm\,0.000$ & $0.557\,\pm\,0.002$ & $0.565\,\pm\,0.017$ & $\mathbf{0.665}\,\pm\,0.004$ & $0.625$ & $0.638$ \\
\cmidrule(lr){1-9}
\textit{Macro} & $0.694$ & $\underline{0.716}$ & $0.686$ & $0.690$ & $0.643$ & $0.708$ & $0.689$ & $\mathbf{0.718}$ \\
\midrule
\multicolumn{9}{l}{\(\mathrm{PCC}_{\mathrm{gene}}\;\uparrow\)} \\
Brain & $0.021\,\pm\,0.004$ & $0.017\,\pm\,0.001$ & $0.023\,\pm\,0.011$ & $0.031\,\pm\,0.007$ & $0.014\,\pm\,0.008$ & $0.020\,\pm\,0.004$ & $\underline{0.033}$ & $\mathbf{0.035}$ \\
Colon & $\underline{0.169}\,\pm\,0.010$ & $0.133\,\pm\,0.004$ & $\mathbf{0.208}\,\pm\,0.015$ & $0.008\,\pm\,0.017$ & $0.076\,\pm\,0.047$ & $0.089\,\pm\,0.040$ & $0.130$ & $0.149$ \\
Heart & $0.018\,\pm\,0.006$ & $0.026\,\pm\,0.007$ & $\underline{0.038}\,\pm\,0.002$ & $0.016\,\pm\,0.022$ & $0.018\,\pm\,0.004$ & $0.011\,\pm\,0.015$ & $0.022$ & $\mathbf{0.040}$ \\
Kidney & $\mathbf{0.065}\,\pm\,0.001$ & $0.038\,\pm\,0.002$ & $0.047\,\pm\,0.005$ & $0.003\,\pm\,0.001$ & $0.058\,\pm\,0.002$ & $0.043\,\pm\,0.010$ & $0.058$ & $\underline{0.061}$ \\
Liver & $0.078\,\pm\,0.000$ & $0.077\,\pm\,0.002$ & $0.040\,\pm\,0.019$ & $\underline{0.083}\,\pm\,0.019$ & $0.068\,\pm\,0.025$ & $0.061\,\pm\,0.001$ & $0.068$ & $\mathbf{0.104}$ \\
Lung & $0.112\,\pm\,0.024$ & $0.157\,\pm\,0.003$ & $0.167\,\pm\,0.021$ & $0.045\,\pm\,0.078$ & $0.122\,\pm\,0.005$ & $\mathbf{0.209}\,\pm\,0.020$ & $0.150$ & $\underline{0.188}$ \\
Prostate & $0.047\,\pm\,0.002$ & $\underline{0.072}\,\pm\,0.001$ & $\mathbf{0.094}\,\pm\,0.008$ & $0.045\,\pm\,0.029$ & $0.051\,\pm\,0.038$ & $0.070\,\pm\,0.004$ & $-0.005$ & $0.070$ \\
Skin & $0.309\,\pm\,0.016$ & $0.254\,\pm\,0.003$ & $\mathbf{0.331}\,\pm\,0.004$ & $0.309\,\pm\,0.025$ & $0.289\,\pm\,0.008$ & $0.206\,\pm\,0.012$ & $\underline{0.315}$ & $0.305$ \\
Uterus & $0.058\,\pm\,0.005$ & $0.048\,\pm\,0.001$ & $\mathbf{0.111}\,\pm\,0.003$ & $0.056\,\pm\,0.015$ & $0.064\,\pm\,0.008$ & $0.050\,\pm\,0.034$ & $0.075$ & $\underline{0.095}$ \\
Breast & $0.065\,\pm\,0.014$ & $\underline{0.122}\,\pm\,0.001$ & $0.074\,\pm\,0.007$ & $0.069\,\pm\,0.012$ & $0.031\,\pm\,0.014$ & $\mathbf{0.141}\,\pm\,0.019$ & $0.070$ & $0.084$ \\
\cmidrule(lr){1-9}
\textit{Macro} & $0.094$ & $0.094$ & $\mathbf{0.113}$ & $0.066$ & $0.079$ & $0.090$ & $0.091$ & $\underline{0.113}$ \\
\midrule
\multicolumn{9}{l}{\(R^2_{\mathrm{local}}\;\uparrow\)} \\
Brain & $-0.051\,\pm\,0.008$ & $-0.001\,\pm\,0.001$ & $\underline{0.008}\,\pm\,0.004$ & $0.004\,\pm\,0.008$ & $-0.332\,\pm\,0.063$ & $\mathbf{0.014}\,\pm\,0.001$ & $-0.187$ & $-0.004$ \\
Colon & $\mathbf{0.125}\,\pm\,0.014$ & $0.060\,\pm\,0.001$ & $0.045\,\pm\,0.015$ & $0.000\,\pm\,0.000$ & $-0.118\,\pm\,0.136$ & $0.038\,\pm\,0.032$ & $-0.044$ & $\underline{0.110}$ \\
Heart & $-0.095\,\pm\,0.037$ & $\underline{-0.031}\,\pm\,0.036$ & $\mathbf{0.004}\,\pm\,0.002$ & $-0.039\,\pm\,0.055$ & $-0.224\,\pm\,0.044$ & $-0.078\,\pm\,0.054$ & $-0.403$ & $-0.061$ \\
Kidney & $\underline{0.019}\,\pm\,0.010$ & $0.010\,\pm\,0.002$ & $0.016\,\pm\,0.005$ & $0.000\,\pm\,0.000$ & $-0.114\,\pm\,0.011$ & $-0.013\,\pm\,0.026$ & $-0.084$ & $\mathbf{0.027}$ \\
Liver & $-0.084\,\pm\,0.078$ & $0.007\,\pm\,0.001$ & $-0.014\,\pm\,0.029$ & $\underline{0.013}\,\pm\,0.005$ & $-0.218\,\pm\,0.158$ & $0.012\,\pm\,0.009$ & $-0.150$ & $\mathbf{0.026}$ \\
Lung & $0.086\,\pm\,0.017$ & $0.079\,\pm\,0.008$ & $0.033\,\pm\,0.002$ & $0.009\,\pm\,0.013$ & $0.041\,\pm\,0.002$ & $\mathbf{0.118}\,\pm\,0.024$ & $0.088$ & $\underline{0.093}$ \\
Prostate & $0.018\,\pm\,0.004$ & $\mathbf{0.040}\,\pm\,0.003$ & $0.022\,\pm\,0.009$ & $-0.026\,\pm\,0.056$ & $-0.034\,\pm\,0.032$ & $\underline{0.034}\,\pm\,0.002$ & $-0.106$ & $0.027$ \\
Skin & $\mathbf{0.361}\,\pm\,0.013$ & $0.266\,\pm\,0.004$ & $0.040\,\pm\,0.000$ & $\underline{0.296}\,\pm\,0.017$ & $0.194\,\pm\,0.006$ & $0.190\,\pm\,0.000$ & $0.258$ & $0.262$ \\
Uterus & $0.063\,\pm\,0.004$ & $\underline{0.065}\,\pm\,0.005$ & $-0.419\,\pm\,0.046$ & $-0.052\,\pm\,0.104$ & $-0.137\,\pm\,0.034$ & $0.025\,\pm\,0.027$ & $-0.041$ & $\mathbf{0.090}$ \\
Breast & $\mathbf{0.049}\,\pm\,0.005$ & $\underline{0.047}\,\pm\,0.002$ & $0.008\,\pm\,0.002$ & $-0.277\,\pm\,0.012$ & $-0.126\,\pm\,0.053$ & $0.031\,\pm\,0.007$ & $-0.019$ & $0.040$ \\
\cmidrule(lr){1-9}
\textit{Macro} & $0.049$ & $\underline{0.054}$ & $-0.026$ & $-0.007$ & $-0.107$ & $0.037$ & $-0.069$ & $\mathbf{0.061}$ \\
\midrule
\multicolumn{9}{l}{\(R^2_{\mathrm{flat}}\;\uparrow\)} \\
Brain & $0.636\,\pm\,0.014$ & $0.649\,\pm\,0.003$ & $0.633\,\pm\,0.004$ & $\mathbf{0.669}\,\pm\,0.004$ & $0.558\,\pm\,0.060$ & $0.649\,\pm\,0.000$ & $0.622$ & $\underline{0.652}$ \\
Colon & $\mathbf{0.523}\,\pm\,0.027$ & $0.462\,\pm\,0.011$ & $0.491\,\pm\,0.007$ & $0.466\,\pm\,0.002$ & $0.139\,\pm\,0.230$ & $0.424\,\pm\,0.084$ & $0.481$ & $\underline{0.513}$ \\
Heart & $0.220\,\pm\,0.084$ & $0.380\,\pm\,0.086$ & $0.248\,\pm\,0.001$ & $0.303\,\pm\,0.002$ & $\mathbf{0.500}\,\pm\,0.017$ & $\underline{0.394}\,\pm\,0.035$ & $0.233$ & $0.288$ \\
Kidney & $0.479\,\pm\,0.011$ & $0.460\,\pm\,0.003$ & $0.495\,\pm\,0.000$ & $0.488\,\pm\,0.001$ & $0.464\,\pm\,0.015$ & $\mathbf{0.515}\,\pm\,0.020$ & $0.495$ & $\underline{0.513}$ \\
Liver & $0.747\,\pm\,0.004$ & $\underline{0.755}\,\pm\,0.009$ & $0.719\,\pm\,0.007$ & $0.741\,\pm\,0.010$ & $0.695\,\pm\,0.039$ & $0.744\,\pm\,0.012$ & $0.727$ & $\mathbf{0.759}$ \\
Lung & $0.081\,\pm\,0.077$ & $0.171\,\pm\,0.023$ & $0.170\,\pm\,0.006$ & $0.208\,\pm\,0.078$ & $0.046\,\pm\,0.057$ & $\mathbf{0.271}\,\pm\,0.018$ & $\underline{0.230}$ & $0.206$ \\
Prostate & $0.060\,\pm\,0.043$ & $0.101\,\pm\,0.027$ & $\mathbf{0.329}\,\pm\,0.009$ & $0.140\,\pm\,0.118$ & $0.116\,\pm\,0.122$ & $\underline{0.272}\,\pm\,0.117$ & $0.087$ & $0.203$ \\
Skin & $\mathbf{0.421}\,\pm\,0.032$ & $0.358\,\pm\,0.003$ & $0.225\,\pm\,0.000$ & $\underline{0.412}\,\pm\,0.012$ & $0.288\,\pm\,0.062$ & $0.229\,\pm\,0.092$ & $0.396$ & $0.384$ \\
Uterus & $0.578\,\pm\,0.008$ & $0.609\,\pm\,0.011$ & $0.519\,\pm\,0.024$ & $0.656\,\pm\,0.049$ & $0.604\,\pm\,0.012$ & $0.615\,\pm\,0.107$ & $\underline{0.670}$ & $\mathbf{0.693}$ \\
Breast & $0.383\,\pm\,0.015$ & $\underline{0.431}\,\pm\,0.002$ & $0.398\,\pm\,0.001$ & $-0.020\,\pm\,0.001$ & $0.303\,\pm\,0.023$ & $\mathbf{0.440}\,\pm\,0.006$ & $0.384$ & $0.404$ \\
\cmidrule(lr){1-9}
\textit{Macro} & $0.413$ & $0.438$ & $0.423$ & $0.406$ & $0.371$ & $\underline{0.455}$ & $0.433$ & $\mathbf{0.461}$ \\
\bottomrule
\end{tabular}%
}
\end{table*}

\clearpage
\begin{table*}[!ht]
\centering
\caption{Cohort-level test results with ResNet18 and \(G=1500\). Neural entries are mean \(\pm\) seed standard deviation over three runs. Linear and CGL-Linear are deterministic. Bold and underlining mark the best and second-best unrounded means in each row. Macro rows average cohort means without weighting by cohort size. Higher is better.}
\label{tab:cohort-resnet18-1500}
\scriptsize
\setlength{\tabcolsep}{2.0pt}
\renewcommand{\arraystretch}{1.08}
\resizebox{\textwidth}{!}{%
\begin{tabular}{lrrrrrrrr}
\toprule
Cohort & HiST & HistoPrism & HEXST & STFlow & BLEEP & SEPAL & Linear & CGL-Linear \\
\midrule
\multicolumn{9}{l}{\(\mathrm{PCC}_{\mathrm{flat}}\;\uparrow\)} \\
Brain & $0.813\,\pm\,0.001$ & $\underline{0.818}\,\pm\,0.001$ & $0.809\,\pm\,0.002$ & $0.817\,\pm\,0.007$ & $0.766\,\pm\,0.007$ & $0.817\,\pm\,0.000$ & $0.805$ & $\mathbf{0.819}$ \\
Colon & $0.721\,\pm\,0.001$ & $\underline{0.724}\,\pm\,0.004$ & $0.711\,\pm\,0.005$ & $0.694\,\pm\,0.001$ & $0.558\,\pm\,0.008$ & $0.706\,\pm\,0.016$ & $0.705$ & $\mathbf{0.727}$ \\
Heart & $\mathbf{0.748}\,\pm\,0.012$ & $0.740\,\pm\,0.021$ & $0.699\,\pm\,0.000$ & $0.707\,\pm\,0.001$ & $\underline{0.741}\,\pm\,0.001$ & $0.711\,\pm\,0.001$ & $0.707$ & $0.722$ \\
Kidney & $0.741\,\pm\,0.019$ & $0.738\,\pm\,0.005$ & $0.744\,\pm\,0.000$ & $0.739\,\pm\,0.001$ & $0.723\,\pm\,0.018$ & $\underline{0.745}\,\pm\,0.011$ & $0.745$ & $\mathbf{0.755}$ \\
Liver & $0.869\,\pm\,0.006$ & $\mathbf{0.880}\,\pm\,0.001$ & $0.846\,\pm\,0.025$ & $0.867\,\pm\,0.007$ & $0.857\,\pm\,0.013$ & $0.855\,\pm\,0.008$ & $0.857$ & $\underline{0.880}$ \\
Lung & $0.559\,\pm\,0.001$ & $\underline{0.599}\,\pm\,0.004$ & $0.558\,\pm\,0.028$ & $0.535\,\pm\,0.016$ & $0.478\,\pm\,0.031$ & $\mathbf{0.615}\,\pm\,0.007$ & $0.562$ & $0.597$ \\
Prostate & $0.573\,\pm\,0.046$ & $0.648\,\pm\,0.046$ & $\mathbf{0.698}\,\pm\,0.001$ & $0.577\,\pm\,0.124$ & $0.436\,\pm\,0.007$ & $\underline{0.686}\,\pm\,0.036$ & $0.516$ & $0.658$ \\
Skin & $\underline{0.679}\,\pm\,0.015$ & $0.603\,\pm\,0.019$ & $0.487\,\pm\,0.005$ & $\mathbf{0.687}\,\pm\,0.001$ & $0.585\,\pm\,0.012$ & $0.552\,\pm\,0.013$ & $0.649$ & $0.638$ \\
Uterus & $0.772\,\pm\,0.002$ & $0.810\,\pm\,0.000$ & $0.704\,\pm\,0.003$ & $0.811\,\pm\,0.027$ & $0.810\,\pm\,0.020$ & $0.779\,\pm\,0.092$ & $\underline{0.820}$ & $\mathbf{0.844}$ \\
Breast & $0.612\,\pm\,0.019$ & $\underline{0.640}\,\pm\,0.001$ & $0.609\,\pm\,0.000$ & $0.583\,\pm\,0.017$ & $0.556\,\pm\,0.006$ & $\mathbf{0.645}\,\pm\,0.012$ & $0.603$ & $0.616$ \\
\cmidrule(lr){1-9}
\textit{Macro} & $0.709$ & $\underline{0.720}$ & $0.687$ & $0.702$ & $0.651$ & $0.711$ & $0.697$ & $\mathbf{0.726}$ \\
\midrule
\multicolumn{9}{l}{\(\mathrm{PCC}_{\mathrm{gene}}\;\uparrow\)} \\
Brain & $0.022\,\pm\,0.004$ & $0.017\,\pm\,0.000$ & $0.021\,\pm\,0.003$ & $0.022\,\pm\,0.004$ & $0.009\,\pm\,0.006$ & $0.018\,\pm\,0.000$ & $\underline{0.030}$ & $\mathbf{0.031}$ \\
Colon & $0.148\,\pm\,0.002$ & $0.130\,\pm\,0.004$ & $\mathbf{0.220}\,\pm\,0.005$ & $0.020\,\pm\,0.033$ & $0.078\,\pm\,0.025$ & $0.091\,\pm\,0.021$ & $0.129$ & $\underline{0.148}$ \\
Heart & $0.021\,\pm\,0.006$ & $0.024\,\pm\,0.009$ & $\underline{0.036}\,\pm\,0.002$ & $0.001\,\pm\,0.001$ & $0.030\,\pm\,0.001$ & $0.031\,\pm\,0.008$ & $0.022$ & $\mathbf{0.040}$ \\
Kidney & $\mathbf{0.061}\,\pm\,0.018$ & $0.041\,\pm\,0.006$ & $0.046\,\pm\,0.002$ & $0.001\,\pm\,0.000$ & $0.057\,\pm\,0.005$ & $0.037\,\pm\,0.013$ & $0.056$ & $\underline{0.059}$ \\
Liver & $0.077\,\pm\,0.012$ & $\underline{0.086}\,\pm\,0.000$ & $0.026\,\pm\,0.021$ & $0.085\,\pm\,0.028$ & $0.082\,\pm\,0.018$ & $0.049\,\pm\,0.034$ & $0.070$ & $\mathbf{0.111}$ \\
Lung & $0.126\,\pm\,0.014$ & $0.169\,\pm\,0.000$ & $\mathbf{0.185}\,\pm\,0.056$ & $0.071\,\pm\,0.103$ & $0.144\,\pm\,0.022$ & $0.183\,\pm\,0.020$ & $0.146$ & $\underline{0.184}$ \\
Prostate & $0.034\,\pm\,0.014$ & $0.045\,\pm\,0.011$ & $\mathbf{0.090}\,\pm\,0.010$ & $0.031\,\pm\,0.049$ & $0.012\,\pm\,0.017$ & $\underline{0.067}\,\pm\,0.009$ & $-0.011$ & $0.066$ \\
Skin & $0.310\,\pm\,0.022$ & $0.233\,\pm\,0.009$ & $\mathbf{0.325}\,\pm\,0.012$ & $\underline{0.323}\,\pm\,0.004$ & $0.305\,\pm\,0.001$ & $0.218\,\pm\,0.037$ & $0.309$ & $0.301$ \\
Uterus & $0.053\,\pm\,0.005$ & $0.055\,\pm\,0.000$ & $\mathbf{0.102}\,\pm\,0.007$ & $0.049\,\pm\,0.010$ & $0.065\,\pm\,0.004$ & $0.040\,\pm\,0.032$ & $0.068$ & $\underline{0.087}$ \\
Breast & $0.075\,\pm\,0.038$ & $\underline{0.130}\,\pm\,0.006$ & $0.087\,\pm\,0.006$ & $0.079\,\pm\,0.020$ & $0.039\,\pm\,0.011$ & $\mathbf{0.147}\,\pm\,0.028$ & $0.071$ & $0.085$ \\
\cmidrule(lr){1-9}
\textit{Macro} & $0.093$ & $0.093$ & $\mathbf{0.114}$ & $0.068$ & $0.082$ & $0.088$ & $0.089$ & $\underline{0.111}$ \\
\midrule
\multicolumn{9}{l}{\(R^2_{\mathrm{local}}\;\uparrow\)} \\
Brain & $-0.071\,\pm\,0.003$ & $0.001\,\pm\,0.002$ & $\underline{0.006}\,\pm\,0.002$ & $0.001\,\pm\,0.019$ & $-0.355\,\pm\,0.118$ & $\mathbf{0.011}\,\pm\,0.000$ & $-0.170$ & $-0.003$ \\
Colon & $\mathbf{0.132}\,\pm\,0.003$ & $0.055\,\pm\,0.004$ & $0.042\,\pm\,0.016$ & $0.000\,\pm\,0.000$ & $-0.126\,\pm\,0.033$ & $0.041\,\pm\,0.020$ & $-0.020$ & $\underline{0.122}$ \\
Heart & $-0.090\,\pm\,0.028$ & $-0.036\,\pm\,0.053$ & $\mathbf{0.002}\,\pm\,0.001$ & $\underline{0.000}\,\pm\,0.000$ & $-0.206\,\pm\,0.015$ & $-0.018\,\pm\,0.008$ & $-0.332$ & $-0.048$ \\
Kidney & $0.013\,\pm\,0.005$ & $\underline{0.015}\,\pm\,0.004$ & $0.015\,\pm\,0.001$ & $0.000\,\pm\,0.000$ & $-0.145\,\pm\,0.018$ & $0.005\,\pm\,0.004$ & $-0.080$ & $\mathbf{0.029}$ \\
Liver & $-0.033\,\pm\,0.008$ & $0.017\,\pm\,0.004$ & $-0.195\,\pm\,0.296$ & $0.014\,\pm\,0.015$ & $-0.203\,\pm\,0.170$ & $\underline{0.022}\,\pm\,0.007$ & $-0.136$ & $\mathbf{0.034}$ \\
Lung & $\mathbf{0.104}\,\pm\,0.007$ & $0.086\,\pm\,0.001$ & $0.061\,\pm\,0.043$ & $0.019\,\pm\,0.027$ & $0.055\,\pm\,0.017$ & $0.086\,\pm\,0.012$ & $0.087$ & $\underline{0.090}$ \\
Prostate & $0.009\,\pm\,0.004$ & $0.019\,\pm\,0.009$ & $0.016\,\pm\,0.002$ & $-0.017\,\pm\,0.070$ & $-0.017\,\pm\,0.005$ & $\mathbf{0.036}\,\pm\,0.013$ & $-0.101$ & $\underline{0.025}$ \\
Skin & $\mathbf{0.359}\,\pm\,0.020$ & $0.248\,\pm\,0.011$ & $0.028\,\pm\,0.006$ & $\underline{0.323}\,\pm\,0.004$ & $0.227\,\pm\,0.002$ & $0.212\,\pm\,0.066$ & $0.263$ & $0.267$ \\
Uterus & $0.057\,\pm\,0.002$ & $\underline{0.068}\,\pm\,0.001$ & $-0.632\,\pm\,0.104$ & $-0.037\,\pm\,0.086$ & $-0.080\,\pm\,0.000$ & $0.029\,\pm\,0.038$ & $-0.039$ & $\mathbf{0.087}$ \\
Breast & $\underline{0.038}\,\pm\,0.008$ & $\mathbf{0.045}\,\pm\,0.001$ & $0.007\,\pm\,0.000$ & $-0.143\,\pm\,0.077$ & $-0.093\,\pm\,0.007$ & $0.025\,\pm\,0.005$ & $-0.017$ & $0.036$ \\
\cmidrule(lr){1-9}
\textit{Macro} & $0.052$ & $\underline{0.052}$ & $-0.065$ & $0.016$ & $-0.094$ & $0.045$ & $-0.055$ & $\mathbf{0.064}$ \\
\midrule
\multicolumn{9}{l}{\(R^2_{\mathrm{flat}}\;\uparrow\)} \\
Brain & $0.657\,\pm\,0.002$ & $0.666\,\pm\,0.001$ & $0.654\,\pm\,0.003$ & $\underline{0.667}\,\pm\,0.013$ & $0.579\,\pm\,0.019$ & $0.664\,\pm\,0.000$ & $0.643$ & $\mathbf{0.670}$ \\
Colon & $0.478\,\pm\,0.015$ & $0.450\,\pm\,0.017$ & $\underline{0.501}\,\pm\,0.006$ & $0.477\,\pm\,0.002$ & $0.225\,\pm\,0.039$ & $0.483\,\pm\,0.009$ & $0.497$ & $\mathbf{0.526}$ \\
Heart & $0.328\,\pm\,0.073$ & $0.380\,\pm\,0.087$ & $0.298\,\pm\,0.000$ & $0.339\,\pm\,0.003$ & $\mathbf{0.534}\,\pm\,0.003$ & $\underline{0.463}\,\pm\,0.012$ & $0.290$ & $0.342$ \\
Kidney & $0.533\,\pm\,0.038$ & $0.535\,\pm\,0.009$ & $0.553\,\pm\,0.000$ & $0.545\,\pm\,0.001$ & $0.518\,\pm\,0.029$ & $\underline{0.554}\,\pm\,0.017$ & $0.553$ & $\mathbf{0.569}$ \\
Liver & $0.752\,\pm\,0.009$ & $\mathbf{0.770}\,\pm\,0.004$ & $0.710\,\pm\,0.035$ & $0.747\,\pm\,0.010$ & $0.727\,\pm\,0.017$ & $0.708\,\pm\,0.028$ & $0.733$ & $\underline{0.767}$ \\
Lung & $0.197\,\pm\,0.000$ & $\underline{0.226}\,\pm\,0.049$ & $0.193\,\pm\,0.036$ & $0.165\,\pm\,0.022$ & $0.088\,\pm\,0.007$ & $0.217\,\pm\,0.044$ & $\mathbf{0.230}$ & $0.202$ \\
Prostate & $0.055\,\pm\,0.043$ & $0.096\,\pm\,0.023$ & $\mathbf{0.316}\,\pm\,0.001$ & $0.068\,\pm\,0.132$ & $-0.057\,\pm\,0.023$ & $\underline{0.283}\,\pm\,0.087$ & $0.084$ & $0.197$ \\
Skin & $\underline{0.441}\,\pm\,0.006$ & $0.354\,\pm\,0.023$ & $0.237\,\pm\,0.004$ & $\mathbf{0.454}\,\pm\,0.001$ & $0.322\,\pm\,0.020$ & $0.243\,\pm\,0.051$ & $0.418$ & $0.406$ \\
Uterus & $0.589\,\pm\,0.004$ & $0.631\,\pm\,0.000$ & $0.446\,\pm\,0.002$ & $0.646\,\pm\,0.056$ & $0.645\,\pm\,0.044$ & $0.541\,\pm\,0.184$ & $\underline{0.665}$ & $\mathbf{0.689}$ \\
Breast & $0.362\,\pm\,0.029$ & $\underline{0.408}\,\pm\,0.003$ & $0.369\,\pm\,0.000$ & $0.138\,\pm\,0.056$ & $0.294\,\pm\,0.007$ & $\mathbf{0.411}\,\pm\,0.012$ & $0.357$ & $0.375$ \\
\cmidrule(lr){1-9}
\textit{Macro} & $0.439$ & $0.452$ & $0.428$ & $0.425$ & $0.387$ & $\underline{0.457}$ & $0.447$ & $\mathbf{0.474}$ \\
\bottomrule
\end{tabular}%
}
\end{table*}

\clearpage
\begin{table*}[!ht]
\centering
\caption{Cohort-level test results with ResNet18 and \(G=2000\). Neural entries are mean \(\pm\) seed standard deviation over three runs. Linear and CGL-Linear are deterministic. Bold and underlining mark the best and second-best unrounded means in each row. Macro rows average cohort means without weighting by cohort size. Higher is better.}
\label{tab:cohort-resnet18-2000}
\scriptsize
\setlength{\tabcolsep}{2.0pt}
\renewcommand{\arraystretch}{1.08}
\resizebox{\textwidth}{!}{%
\begin{tabular}{lrrrrrrrr}
\toprule
Cohort & HiST & HistoPrism & HEXST & STFlow & BLEEP & SEPAL & Linear & CGL-Linear \\
\midrule
\multicolumn{9}{l}{\(\mathrm{PCC}_{\mathrm{flat}}\;\uparrow\)} \\
Brain & $0.810\,\pm\,0.000$ & $\underline{0.812}\,\pm\,0.000$ & $0.806\,\pm\,0.002$ & $0.812\,\pm\,0.000$ & $0.763\,\pm\,0.010$ & $0.809\,\pm\,0.001$ & $0.801$ & $\mathbf{0.813}$ \\
Colon & $\mathbf{0.730}\,\pm\,0.004$ & $0.719\,\pm\,0.003$ & $0.699\,\pm\,0.001$ & $0.686\,\pm\,0.000$ & $0.579\,\pm\,0.029$ & $0.700\,\pm\,0.012$ & $0.699$ & $\underline{0.721}$ \\
Heart & $\mathbf{0.753}\,\pm\,0.006$ & $\underline{0.748}\,\pm\,0.022$ & $0.709\,\pm\,0.000$ & $0.705\,\pm\,0.017$ & $0.741\,\pm\,0.004$ & $0.718\,\pm\,0.019$ & $0.718$ & $0.732$ \\
Kidney & $0.740\,\pm\,0.001$ & $0.731\,\pm\,0.001$ & $0.739\,\pm\,0.000$ & $0.731\,\pm\,0.000$ & $0.723\,\pm\,0.005$ & $0.739\,\pm\,0.004$ & $\underline{0.741}$ & $\mathbf{0.751}$ \\
Liver & $0.866\,\pm\,0.003$ & $\underline{0.873}\,\pm\,0.002$ & $0.833\,\pm\,0.035$ & $0.865\,\pm\,0.012$ & $0.854\,\pm\,0.011$ & $0.850\,\pm\,0.024$ & $0.855$ & $\mathbf{0.877}$ \\
Lung & $0.546\,\pm\,0.007$ & $0.578\,\pm\,0.009$ & $0.510\,\pm\,0.067$ & $0.510\,\pm\,0.001$ & $0.447\,\pm\,0.004$ & $\mathbf{0.600}\,\pm\,0.017$ & $0.547$ & $\underline{0.581}$ \\
Prostate & $0.606\,\pm\,0.034$ & $0.614\,\pm\,0.041$ & $\mathbf{0.681}\,\pm\,0.001$ & $\underline{0.667}\,\pm\,0.022$ & $0.493\,\pm\,0.052$ & $0.664\,\pm\,0.016$ & $0.500$ & $0.642$ \\
Skin & $\mathbf{0.679}\,\pm\,0.001$ & $0.635\,\pm\,0.001$ & $0.493\,\pm\,0.003$ & $0.648\,\pm\,0.004$ & $0.517\,\pm\,0.211$ & $0.598\,\pm\,0.026$ & $\underline{0.653}$ & $0.642$ \\
Uterus & $0.785\,\pm\,0.019$ & $0.805\,\pm\,0.004$ & $0.754\,\pm\,0.093$ & $\underline{0.815}\,\pm\,0.005$ & $0.803\,\pm\,0.004$ & $0.765\,\pm\,0.096$ & $0.813$ & $\mathbf{0.837}$ \\
Breast & $0.583\,\pm\,0.011$ & $\underline{0.615}\,\pm\,0.005$ & $0.589\,\pm\,0.000$ & $0.536\,\pm\,0.010$ & $0.525\,\pm\,0.013$ & $\mathbf{0.621}\,\pm\,0.013$ & $0.581$ & $0.595$ \\
\cmidrule(lr){1-9}
\textit{Macro} & $0.710$ & $\underline{0.713}$ & $0.681$ & $0.698$ & $0.645$ & $0.706$ & $0.691$ & $\mathbf{0.719}$ \\
\midrule
\multicolumn{9}{l}{\(\mathrm{PCC}_{\mathrm{gene}}\;\uparrow\)} \\
Brain & $0.027\,\pm\,0.003$ & $0.018\,\pm\,0.000$ & $0.026\,\pm\,0.002$ & $0.012\,\pm\,0.018$ & $0.018\,\pm\,0.010$ & $0.019\,\pm\,0.001$ & $\underline{0.031}$ & $\mathbf{0.032}$ \\
Colon & $\underline{0.155}\,\pm\,0.007$ & $0.134\,\pm\,0.013$ & $\mathbf{0.211}\,\pm\,0.000$ & $-0.004\,\pm\,0.014$ & $0.101\,\pm\,0.037$ & $0.075\,\pm\,0.006$ & $0.129$ & $0.147$ \\
Heart & $0.019\,\pm\,0.000$ & $0.026\,\pm\,0.009$ & $\underline{0.035}\,\pm\,0.003$ & $0.017\,\pm\,0.024$ & $0.028\,\pm\,0.002$ & $0.029\,\pm\,0.008$ & $0.021$ & $\mathbf{0.039}$ \\
Kidney & $\mathbf{0.060}\,\pm\,0.002$ & $0.035\,\pm\,0.006$ & $0.040\,\pm\,0.003$ & $0.000\,\pm\,0.000$ & $0.051\,\pm\,0.004$ & $0.025\,\pm\,0.007$ & $0.052$ & $\underline{0.056}$ \\
Liver & $0.084\,\pm\,0.001$ & $0.091\,\pm\,0.001$ & $0.014\,\pm\,0.016$ & $\underline{0.098}\,\pm\,0.021$ & $0.093\,\pm\,0.020$ & $0.050\,\pm\,0.044$ & $0.076$ & $\mathbf{0.119}$ \\
Lung & $0.122\,\pm\,0.008$ & $0.160\,\pm\,0.007$ & $0.171\,\pm\,0.051$ & $-0.001\,\pm\,0.000$ & $0.100\,\pm\,0.005$ & $\mathbf{0.185}\,\pm\,0.043$ & $0.142$ & $\underline{0.180}$ \\
Prostate & $0.047\,\pm\,0.043$ & $0.039\,\pm\,0.004$ & $\mathbf{0.087}\,\pm\,0.023$ & $\underline{0.067}\,\pm\,0.038$ & $0.027\,\pm\,0.022$ & $0.054\,\pm\,0.025$ & $-0.014$ & $0.067$ \\
Skin & $0.292\,\pm\,0.005$ & $0.248\,\pm\,0.003$ & $\mathbf{0.322}\,\pm\,0.017$ & $0.267\,\pm\,0.008$ & $0.240\,\pm\,0.108$ & $0.227\,\pm\,0.018$ & $\underline{0.300}$ & $0.293$ \\
Uterus & $0.058\,\pm\,0.006$ & $0.055\,\pm\,0.002$ & $\mathbf{0.083}\,\pm\,0.015$ & $0.040\,\pm\,0.032$ & $0.064\,\pm\,0.006$ & $0.039\,\pm\,0.037$ & $0.064$ & $\underline{0.082}$ \\
Breast & $0.075\,\pm\,0.011$ & $\underline{0.127}\,\pm\,0.010$ & $0.098\,\pm\,0.008$ & $0.070\,\pm\,0.010$ & $0.033\,\pm\,0.009$ & $\mathbf{0.133}\,\pm\,0.040$ & $0.071$ & $0.085$ \\
\cmidrule(lr){1-9}
\textit{Macro} & $0.094$ & $0.093$ & $\underline{0.109}$ & $0.057$ & $0.076$ & $0.084$ & $0.087$ & $\mathbf{0.110}$ \\
\midrule
\multicolumn{9}{l}{\(R^2_{\mathrm{local}}\;\uparrow\)} \\
Brain & $-0.059\,\pm\,0.014$ & $0.003\,\pm\,0.002$ & $\underline{0.007}\,\pm\,0.002$ & $\mathbf{0.012}\,\pm\,0.017$ & $-0.311\,\pm\,0.011$ & $0.005\,\pm\,0.006$ & $-0.151$ & $0.000$ \\
Colon & $\mathbf{0.143}\,\pm\,0.016$ & $0.063\,\pm\,0.004$ & $0.028\,\pm\,0.003$ & $0.000\,\pm\,0.000$ & $-0.149\,\pm\,0.085$ & $0.030\,\pm\,0.013$ & $-0.008$ & $\underline{0.128}$ \\
Heart & $-0.071\,\pm\,0.005$ & $-0.028\,\pm\,0.040$ & $\mathbf{0.001}\,\pm\,0.000$ & $-0.059\,\pm\,0.083$ & $-0.162\,\pm\,0.005$ & $\underline{-0.015}\,\pm\,0.005$ & $-0.295$ & $-0.042$ \\
Kidney & $\underline{0.026}\,\pm\,0.008$ & $0.014\,\pm\,0.002$ & $0.010\,\pm\,0.002$ & $0.000\,\pm\,0.000$ & $-0.108\,\pm\,0.022$ & $0.003\,\pm\,0.001$ & $-0.075$ & $\mathbf{0.029}$ \\
Liver & $-0.049\,\pm\,0.034$ & $0.008\,\pm\,0.002$ & $-0.305\,\pm\,0.440$ & $-0.003\,\pm\,0.001$ & $-0.186\,\pm\,0.124$ & $\underline{0.017}\,\pm\,0.004$ & $-0.127$ & $\mathbf{0.034}$ \\
Lung & $\mathbf{0.101}\,\pm\,0.003$ & $0.075\,\pm\,0.000$ & $0.054\,\pm\,0.049$ & $0.000\,\pm\,0.000$ & $0.023\,\pm\,0.008$ & $\underline{0.091}\,\pm\,0.039$ & $0.082$ & $0.083$ \\
Prostate & $0.024\,\pm\,0.049$ & $0.016\,\pm\,0.001$ & $0.011\,\pm\,0.005$ & $\mathbf{0.041}\,\pm\,0.039$ & $-0.017\,\pm\,0.001$ & $0.018\,\pm\,0.030$ & $-0.094$ & $\underline{0.024}$ \\
Skin & $\mathbf{0.371}\,\pm\,0.009$ & $\underline{0.276}\,\pm\,0.011$ & $0.029\,\pm\,0.003$ & $0.256\,\pm\,0.023$ & $0.170\,\pm\,0.142$ & $0.218\,\pm\,0.023$ & $0.265$ & $0.270$ \\
Uterus & $0.060\,\pm\,0.011$ & $\underline{0.066}\,\pm\,0.002$ & $-0.350\,\pm\,0.533$ & $0.017\,\pm\,0.010$ & $-0.045\,\pm\,0.018$ & $0.025\,\pm\,0.036$ & $-0.035$ & $\mathbf{0.082}$ \\
Breast & $\underline{0.036}\,\pm\,0.008$ & $\mathbf{0.043}\,\pm\,0.004$ & $0.006\,\pm\,0.000$ & $-0.218\,\pm\,0.019$ & $-0.110\,\pm\,0.047$ & $0.023\,\pm\,0.006$ & $-0.020$ & $0.034$ \\
\cmidrule(lr){1-9}
\textit{Macro} & $\underline{0.058}$ & $0.054$ & $-0.051$ & $0.005$ & $-0.090$ & $0.041$ & $-0.046$ & $\mathbf{0.064}$ \\
\midrule
\multicolumn{9}{l}{\(R^2_{\mathrm{flat}}\;\uparrow\)} \\
Brain & $0.652\,\pm\,0.004$ & $0.656\,\pm\,0.000$ & $0.649\,\pm\,0.003$ & $\underline{0.657}\,\pm\,0.001$ & $0.574\,\pm\,0.017$ & $0.651\,\pm\,0.003$ & $0.636$ & $\mathbf{0.661}$ \\
Colon & $\underline{0.501}\,\pm\,0.019$ & $0.445\,\pm\,0.001$ & $0.484\,\pm\,0.001$ & $0.468\,\pm\,0.000$ & $0.273\,\pm\,0.083$ & $0.459\,\pm\,0.024$ & $0.488$ & $\mathbf{0.517}$ \\
Heart & $0.356\,\pm\,0.085$ & $0.422\,\pm\,0.070$ & $0.326\,\pm\,0.000$ & $0.399\,\pm\,0.060$ & $\mathbf{0.534}\,\pm\,0.010$ & $\underline{0.482}\,\pm\,0.040$ & $0.311$ & $0.365$ \\
Kidney & $0.538\,\pm\,0.004$ & $0.524\,\pm\,0.005$ & $0.546\,\pm\,0.001$ & $0.534\,\pm\,0.001$ & $0.519\,\pm\,0.010$ & $0.546\,\pm\,0.007$ & $\underline{0.547}$ & $\mathbf{0.564}$ \\
Liver & $0.749\,\pm\,0.005$ & $\underline{0.754}\,\pm\,0.005$ & $0.684\,\pm\,0.055$ & $0.748\,\pm\,0.021$ & $0.726\,\pm\,0.017$ & $0.685\,\pm\,0.093$ & $0.728$ & $\mathbf{0.760}$ \\
Lung & $0.175\,\pm\,0.013$ & $0.182\,\pm\,0.010$ & $0.163\,\pm\,0.036$ & $0.141\,\pm\,0.002$ & $0.014\,\pm\,0.002$ & $\mathbf{0.221}\,\pm\,0.053$ & $\underline{0.214}$ & $0.187$ \\
Prostate & $0.104\,\pm\,0.024$ & $0.056\,\pm\,0.001$ & $\mathbf{0.293}\,\pm\,0.003$ & $0.150\,\pm\,0.091$ & $0.022\,\pm\,0.081$ & $\underline{0.266}\,\pm\,0.063$ & $0.072$ & $0.182$ \\
Skin & $\mathbf{0.451}\,\pm\,0.007$ & $0.399\,\pm\,0.005$ & $0.242\,\pm\,0.002$ & $0.397\,\pm\,0.029$ & $0.259\,\pm\,0.243$ & $0.332\,\pm\,0.040$ & $\underline{0.423}$ & $0.411$ \\
Uterus & $0.605\,\pm\,0.033$ & $0.624\,\pm\,0.005$ & $0.532\,\pm\,0.185$ & $0.638\,\pm\,0.042$ & $0.630\,\pm\,0.000$ & $0.495\,\pm\,0.235$ & $\underline{0.654}$ & $\mathbf{0.678}$ \\
Breast & $0.330\,\pm\,0.017$ & $\underline{0.372}\,\pm\,0.003$ & $0.345\,\pm\,0.000$ & $0.000\,\pm\,0.036$ & $0.251\,\pm\,0.028$ & $\mathbf{0.384}\,\pm\,0.017$ & $0.329$ & $0.350$ \\
\cmidrule(lr){1-9}
\textit{Macro} & $0.446$ & $0.443$ & $0.426$ & $0.413$ & $0.380$ & $\underline{0.452}$ & $0.440$ & $\mathbf{0.468}$ \\
\bottomrule
\end{tabular}%
}
\end{table*}

\clearpage

\end{document}